# Sparse Input View Synthesis:
# 3D Representations and Reliable Priors

A Thesis

Submitted for the Degree of

## 𝔇𝔬𝔠𝔱𝔬𝔯 𝔬𝔣 𝔓𝔥𝔦𝔩𝔬𝔰𝔬𝔭𝔥𝔶

in the Faculty of Engineering

Submitted by

## Nagabhushan S N

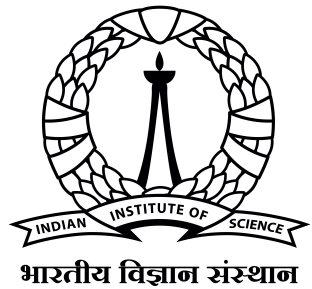

भारतीय विज्ञान संस्थान

Electrical Communication Engineering

Indian Institute of Science

Bangalore – 560 012

May 2024

Dedicated To

*My Mother*

# Acknowledgements

कर्मण्येवाधिकारस्ते मा फलेषु कदाचन। This verse, spoken by Shree Krishna Paramatma to Arjuna in the Bhagavad Gita, emphasizes that we have control only over our actions and not over the results. In other words, we should focus on giving our best efforts without worrying about the outcomes. My mother instilled this principle in me from a young age, and it has guided me throughout my PhD journey. Indeed, I believe it holds particular relevance in research, where despite our best efforts, positive results are not guaranteed. I am eternally grateful to my parents for nurturing me with such principles, which have empowered me to persevere through my doctoral studies.

This journey would not have been possible without the guidance and support of my advisor, Dr. Rajiv Soundararajan. His insistence on never settling for a solution that merely works, but rather striving for one that is not only effective but also scientifically comprehensible, has been a profound source of inspiration. His remarkable attention to detail has consistently impressed me, and his exemplary presentation skills have served as a model for my own development as a presenter. I have learned invaluable techniques from him that undoubtedly have enhanced my ability to conduct research, present my findings effectively, and teach a subject. I am thankful for his patience, mentorship, and unwavering support throughout my PhD.

I extend my gratitude to the Indian Institute of Science (IISc) for providing a picturesque campus in the heart of Bengaluru. My heartfelt thanks to the Department of Electrical Communication Engineering for furnishing me with the essential resources and infrastructure needed for my research. Special appreciation goes to the department's administrative staff for managing all logistical aspects, allowing me to focus entirely on my research. I am indebted to the mess workers for providing healthy and tasty food, which ensured that I could concentrate on my research without feeling hungry. I am grateful to the Ministry of Education (MoE), Government of India, for their financial support through the Prime Minister's Research Fellowship (PMRF). Additionally, I express my appreciation to Qualcomm for partially funding my research endeavors.

I seize this opportunity to express my gratitude to all my teachers, teaching assistants, advisors, and mentors who have played pivotal roles in shaping my academic journey. I am thankful to my labmates, collaborators, and colleagues for engaging in valuable technical discussions and brainstorming sessions, which have enabled me to approach problems from various perspectives. Engaging in sports, particularly badminton, has been a significant stress reliever for me, and I am thankful to my friends for providing unwavering support throughout this journey. Additionally, I extend my heartfelt appreciation to my relatives, friends, and all acquaintances who have supported me in numerous ways at various stages of my life. I am grateful to all my F.R.I.E.N.D.S. who have created unforgettable memories with me, memories that I will cherish for a lifetime. Finally, I would like to express my gratitude to everyone who supported me directly or indirectly.

# Abstract

Novel view synthesis refers to the problem of synthesizing novel viewpoints of a scene given the images from a few viewpoints. This is a fundamental problem in computer vision and graphics, and enables a vast variety of applications such as meta-verse, free-view watching of events, video gaming, video stabilization and video compression. Recent 3D representations such as radiance fields and multi-plane images significantly improve the quality of images rendered from novel viewpoints. However, these models require a dense sampling of input views for high quality renders. Their performance goes down significantly when only a few input views are available. In this thesis, we focus on the sparse input novel view synthesis problem for both static and dynamic scenes.

In the first part of this work, we mainly focus on sparse input novel view synthesis of static scenes using neural radiance fields (NeRF). We study the design of reliable and dense priors to better regularize the NeRF in such situations. In particular, we propose a prior on the visibility of the pixels in a pair of input views. We show that this visibility prior, which is related to the relative depth of objects, is dense and more reliable than existing priors on absolute depth. We compute the visibility prior using plane sweep volumes without the need to train a neural network on large datasets. We evaluate our approach on multiple datasets and show that our model outperforms existing approaches for sparse input novel view synthesis.

In the second part, we aim to further improve the regularization by learning a scene-specific prior that does not suffer from generalization issues. We achieve this by learning the prior on the given scene alone without pre-training on large datasets. In particular, we design augmented NeRFs to obtain better depth supervision in certain regions of the scene for the main NeRF. Further, we extend this framework to also apply to newer and faster radiance field models such as TensoRF and ZipNeRF. Through extensive experiments on multiple datasets, we show the superiority of our approach in sparse input novel view synthesis.

The design of sparse input fast dynamic radiance fields is severely constrained by the lack of suitable representations and reliable priors for motion. We address the first challenge by designing an explicit motion model based on factorized volumes that is compact and optimizes quickly. We also introduce reliable sparse flow priors to constrain the motion field, since we find that the popularly employed dense optical flow priors are unreliable. We show the benefits of our motion representation and reliable priors on multiple datasets.

In the final part of this thesis, we study the application of view synthesis for frame rate upsampling in video gaming. Specifically, we consider the problem of temporal view synthesis, where the goal is to predict the future frames given the past frames and the camera motion. The key challenge here is in predicting the future motion of the objects by estimating their past motion and extrapolating it. We explore the use of multi-plane image representations and scene depth to reliably estimate the object motion, particularly in the occluded regions. We design a new database to effectively evaluate our approach for temporal view synthesis of dynamic scenes and show that we achieve state-of-the-art performance.

# List of publications from the thesis

## Journal papers

1. N. Somraj, A. Karanayil, S. H. Mupparaju, R. Soundararajan, "Simple-RF: Regularizing Sparse Input Radiance Fields with Simpler Solutions", under review at Transactions on Graphics (TOG).

## Conference papers

1. N. Somraj, R. Soundararajan, "ViP-NeRF: Visibility Prior for Sparse Input Neural Radiance Fields", in Proc. ACM Special Interest Group on Computer Graphics and Interactive Techniques (SIGGRAPH), Los Angeles, CA, USA, Aug 2023.

2. N. Somraj, A. Karanayil, R. Soundararajan, "SimpleNeRF: Regularizing Sparse Input Neural Radiance Fields with Simpler Solutions", in Proc. ACM SIGGRAPH Asia, Sydney, Australia, Dec 2023.

3. N. Somraj, K. Choudhary, S. H. Mupparaju, R. Soundararajan, "Factorized Motion Fields for Fast Sparse Input Dynamic View Synthesis", in Proc. ACM Special Interest Group on Computer Graphics and Interactive Techniques (SIGGRAPH), Denver, CO, USA, Jul 2024.

4. N. Somraj, P. Sancheti, R. Soundararajan, "Temporal View Synthesis of Dynamic Scenes through 3D Object Motion Estimation with Multi-Plane Images", in Proc.

IEEE International Symposium on Mixed and Augmented Reality (ISMAR), Singapore, Oct 2022.

## Other publications (not part of this thesis)

1. V. Kanchana, N. Somraj, S. Y., R. Soundararajan, "Revealing Disocclusions in Temporal View Synthesis through Infilling Vector Prediction", in Proc. IEEE Winter Conference on Applications of Computer Vision (WACV), Hawaii, USA, Jan 2022.

2. N. Somraj, M. S. Kashi, S. P. Arun, R. Soundararajan, "Understanding the Perceived Quality of Video Predictions", in Signal Processing: Image Communication, vol 102, Mar 2022.

# Contents

# List of Tables

x

# List of Figures

# Chapter 1

# Introduction

The ability to synthesize novel views of a scene enables diverse applications such as 3D TV, extended reality, video gaming, video compression and video stabilization. Novel view synthesis has been a long-standing problem of interest in computer vision and graphics. Image based Rendering (IBR) approaches synthesize the novel views by copying the RGB color from the source views to the target pixel. The use of depth to determine the source and target pixels can help improve the performance. However, IBR based approaches struggle in handling specular objects, soft edges, and thin structures [134]. This motivated researchers to explore volumetric representations for novel view synthesis. Such an approach is also known as inverse rendering, since it involves learning a 3D representation of the scene from 2D images. Researchers explored multiple techniques such as plane sweep volume based representations [52], multi-plane images [216] and layered depth images [153], culminating in the neural radiance field (NeRF) representation [123]. NeRFs achieve remarkable performance in novel view synthesis by employing a continuous depth representation using neural networks.

NeRFs require a dense sampling of the input views for photo-realistic rendering of novel views. However, in multiple applications such as virtual or augmented reality, telepresence, robotics, and autonomous driving, obtaining a dense set of views is expensive, and only a few input images may be available for training [128]. In such settings however, external sensors or a pre-calibrated fixed camera array may be employed to obtain accurate camera poses. Thus, there is a need to train NeRFs with few input views referred to as the sparse input NeRF problem.

Popular approaches to train NeRF with sparse input views include conditioning the NeRF on a latent scene representation [210] or regularizing with scene priors [46, 142]. However, the existing approaches [142, 210] suffer from issues related to the generalization of the scene latent or learned depth priors. Other depth priors [46] are not dense enough to sufficiently constrain the NeRF. This motivates the exploration of dense and reliable priors to constrain the sparse input NeRF. Firstly, we attempt to obtain a reliable and dense prior in terms of visibility of pixels in a pair of input views. We estimate this visibility prior without the need to train a neural network on large datasets. Although this approach outperforms existing priors, we then explore if scene-specific priors can be learned in-situ. Specifically, we design augmented models that provide better depth supervision in certain regions of the scene. We outperform existing approaches on multiple popular datasets used to evaluate sparse input NeRFs. We also extend this framework to explicit radiance fields for fast optimization and rendering times. We show that similar depth supervision can enable effective learning of sparse input explicit radiance fields.

While our focus so far has been on the novel view synthesis of static scenes, there exist several applications where the scene contains dynamic objects. It is of considerable interest to synthesize novel views of such scenes, where the movement of the objects in the scene adds an additional challenge. This motivates the study of motion priors for sparse input dynamic view synthesis. Further, we wish to study this in the context of explicit radiance fields for fast optimization and rendering. However, existing explicit dynamic radiance fields [28, 53] employ 4D volumetric representations without a motion model. Hence, they do not allow the motion implicitly learned by the model to be regularized using motion priors. Thus, we design a dynamic radiance field with an explicit motion field that lends itself to be constrained with motion supervision. Further, the motion priors used in the literature such as dense optical flow maps [169] are unreliable, and hence we design reliable motion priors for constraining the motion model.

Novel view synthesis also finds applications in synthetically rendered videos such as video games and graphical rendering of high resolution content on handheld mobile devices. Specifically, novel view synthesis can be employed to improve the user experience by upsampling the frame-rate of graphically rendered videos [7]. This is also known as interleaved reprojection [4, 19, 20, 93], where the frames are rendered graphically periodically, and the intermediate frames are predicted based on the past frames. The prediction of these frames needs to adjust for the camera motion or change in view point of the user as well as the movement of objects in the scene. This leads us to the problem of temporal view synthesis of dynamic scenes (TVS-DS), where the goal is to generate the future video frame given its camera pose. In such applications, the content creators make the depth of the scene available, while the motion vectors for moving objects may not typically be available. We leverage the Multiplane Image (MPI) 3D representation for effective learning of the object motion for TVS-DS.

## 1.1 Contributions

The main contributions of this thesis are the design of reliable priors to better constrain the learning process and the design of volumetric representations that are better suited for sparse input view synthesis. In the following sections, we provide a brief overview of the contributions of this thesis for various problems in novel view synthesis as discussed above.

### 1.1.1 Visibility Prior for Sparse Input Neural Radiance Fields

Our goal is to design a dense and reliable prior for constraining the sparse input NeRF for static scenes. In this context, we explore the use of regularization in terms of the visibility of a pixel from a pair of viewpoints. Here, the visibility of a pixel refers to whether the corresponding object is seen in both viewpoints. For example, foreground objects are typically visible in multiple views whereas the background objects may be partially occluded. The visibility of a pixel in different views relies more on the relative depth of the scene objects than the absolute depth. We hypothesize that, given sparse input views, it may be easier to estimate the relative depth and visibility instead of the absolute depth. Thus, the key idea is to regularize the NeRF with a dense visibility prior estimated using the given sparse input views. This allows the NeRF to learn better scene representation. We refer to our Visibility Prior regularized NeRF model as ViP-NeRF.

To obtain the visibility prior, we employ the plane sweep volumes (PSV) [42] that have successfully been used in depth estimation [55, 65, 74, 206] and view synthesis models [216]. We create the PSV by warping one of the images to the view of the other at different depths (or planes) and compare them to obtain error maps. We determine a binary visibility map for each pixel based on the corresponding errors in the PSV. We regularize the NeRF training by using such a map as supervision for every pair of input views. We use the visibility prior in conjunction with the depth prior from DS-NeRF [46], where the former provides a dense prior on relative depth while the latter provides a sparse prior on absolute depth. Note that the estimation of our visibility prior does not require any pre-training on a large dataset.

Regularizing the NeRF with a dense visibility prior is computationally intensive and can lead to impractical training times. We reformulate the NeRF to directly and additionally output visibility to impose the regularization in a computationally efficient manner. We conduct experiments on two popular datasets to demonstrate the efficacy of the visibility prior for sparse input NeRF.

The main contributions of our work are as follows.

- We introduce visibility regularization to train the NeRF with sparse input views and refer to our model as ViP-NeRF.

- We estimate the dense visibility prior reliably using plane sweep volumes.

- We reformulate the NeRF MLP to output visibility thereby significantly reducing the training time.

- We outperform prior approaches on sparse input NeRFs on multiple datasets.

## 1.1.2  Regularizing Sparse Input Radiance Fields with Simpler Solutions

Although the visibility prior is dense and reliable, it is not as rich as a dense depth prior. We aim to obtain dense depth priors that do not suffer from generalization issues by learning the dense depth supervision in-situ without employing any pre-training. Further, we desire to design a single framework of regularization that is applicable to both implicit and explicit radiance fields. While there exists a plethora of implicit and explicit radiance field models, we consider the NeRF as the representative model for implicit radiance fields and consider two explicit radiance fields, namely, TensoRF [33] and ZipNeRF [18].

We first observe that the radiance field models often exploit their high capability to learn unnecessary complex solutions when training with sparse input views. While these solutions perfectly explain the observed images, they can cause severe distortions in novel views. For example, some of the key components of the radiance fields, such as positional encoding in the NeRF or vector-matrix decomposition employed in TensoRF, provide powerful capabilities to the radiance field and are designed for training the model with dense input views. Existing implementations of these components may be sub-optimal with fewer input views due to the highly under-constrained system, causing several distortions. We follow the popular Occam's razor principle and regularize the radiance fields to choose simpler solutions over complex ones, wherever possible. In particular, we design augmented models by reducing the capabilities of the radiance fields and use the depth estimated by these models to supervise the main radiance field.

We design different augmentations for NeRF, TensoRF and ZipNeRF based on different shortcomings and architectures of these models. The high positional encoding degree used in the NeRF leads to undesired depth discontinuities, creating floaters. Further, the view-dependent radiance feature leads to shape-radiance ambiguity, creating duplication artifacts. We design augmentations for the NeRF by reducing the positional encoding degree and disabling the view-dependent radiance feature. On the other hand, the large number of high-resolution factorized components in TensoRF and the large hash table in ZipNeRF cause floaters in these models in the few-shot setting. Thus, we design augmentations to constrain the model with respect to such components to learn simpler solutions.

We use the simplified models as augmentations for depth supervision and not as the main NeRF model since naïvely reducing the capacity of the radiance fields may lead to sub-optimal solutions in certain regions [75]. For example, the model that can learn only smooth depth transitions may fail to learn sharp depth discontinuities at object boundaries. Further, the augmented models need to be used for supervision only if they explain the observed images accurately. We gauge the reliability of the depths by reprojecting pixels using the estimated depths onto a different nearest train view and comparing them with the corresponding images.

We refer to our family of regularized models as Simple Radiance Fields (Simple-RF) since we regularize the models to choose simple solutions over complex ones, wherever feasible. We refer to the individual models as Simple-NeRF, Simple-TensoRF and Simple-ZipNeRF respectively. We evaluate our models on four popular datasets that include forward-facing scenes (NeRF-LLFF), unbounded forward-facing scenes (RealEstate-10K), unbounded 360° scenes (MipNeRF360) and bounded 360° scenes (NeRF-Synthetic) and show that our models achieve significant improvement in performance on all the datasets. Further, we show that our model learns geometry significantly better than prior art. We will release the source code for all our models.

We list the main contributions of our work in the following.

- We find that the high positional encoding degree and view-dependent radiance of the NeRF cause floater and duplication artifacts when training with sparse inputs. We design augmented models on both these fronts to supervise the main NeRF and mitigate both artifacts.

- We observe that the large number of high-resolution decomposed components in TensoRF leads to floater artifacts with sparse inputs. Thus, the augmented model is obtained by reducing the number and resolutions of the decomposed components.

- We find that the large hash table in ZipNeRF causes floaters when training with sparse inputs. The augmented model is designed by reducing the size of the hash table.

- We design a mechanism to determine whether the depths estimated by the augmented models are accurate and utilize only the accurate estimates to supervise the main radiance field.

- We show that our regularization achieves substantial improvements on different radiance fields and on four different datasets.

### 1.1.3 Factorized Motion Fields for Fast Sparse Input Dynamic View Synthesis

To achieve novel view synthesis of sparse input dynamic scenes, we seek to design a fast and compact dynamic radiance field model that is amenable to supervision with motion priors. We design a dynamic radiance field consisting of two models, a 5D radiance field that learns the 3D scene at a canonical time instant and a 4D motion or deformation field that learns the motion from any time instant $t$ to the canonical time instant $t'$. Since the motion model is a one-directional mapping from $t$ to $t'$, it is not obvious how to impose flow priors across two arbitrary time instants. We achieve this by constraining our motion model to map a pair of matched points, obtained using the motion prior, to the same 3D point in the canonical volume. This allows us to impose motion priors across any two time instants and across any cameras.

Prior and concurrent works [49, 150] employ deep neural networks (DNN) to learn the motion field. However, the use of the DNN to learn the motion makes the model computationally expensive. The challenge here is to design a motion model that can learn the motion efficiently while yielding fast training and rendering. While explicit models such as voxel grids [54, 165], hash-grids [125] and 3D Gaussians (3DGS) [81] are shown to be effective in learning static scenes, naively extending these techniques to model 4D motion may not be efficient. For example, extending 3D voxel grids to 4D scales the memory requirement to the fourth power of grid resolution. Further, for a given object, since the motion exists at every time instant, models that exploit scene sparsity to reduce the memory requirement may not extend effectively to 4D. Since the motion is dense and has a high spatio-temporal correlation, we employ factorized volumes to exploit the correlation. Specifically, we employ a 4D factorized volume to learn the scene flow from any time $t$ to the canonical time $t'$.

We find that dense optical flow priors obtained using deep flow estimation networks suffer from generalization issues. Regularizing the motion model with such noisy flow priors may lead to sub-optimal performance. We address this challenge by employing a reliable sparse flow prior obtaining by matching SIFT keypoints.

We evaluate our model on two popular multi-view dynamic scene datasets and find that our model outperforms the state-of-the-art dynamic view synthesis models with fewer input viewpoints. We refer to our model as SF-DeRF, since we employ Sparse Flow priors for Deformable Radiance Fields. We summarize the main contributions of our work in the following:

- We design a fast and compact dynamic radiance field for sparse input dynamic view synthesis by employing an explicit motion model that can be easily regularized using motion priors. We employ a 4D factorized volume to exploit the spatio-temporal correlation of the motion field.

- We propose reliable flow priors based on matching sparse SIFT keypoints across cameras and time instants.

- We achieve very good dynamic view synthesis performance on two popular multi-view datasets with very few views.

### 1.1.4   Temporal View Synthesis of Dynamic Scenes through 3D Object Motion Estimation with Multi-Plane Images

The key challenges in TVS-DS involve leveraging the user or camera motion to extrapolate the past motion of moving objects, combining these to predict the next frame, and infilling any disocclusions arising out of the combined motion. We design a framework to decouple camera and object motion, which allows us to effectively use the available camera motion and only predict the object motion. To predict future object motion, we estimate the object motion in the past frames and then extrapolate it. However both the camera motion and object motion are intertwined in the past frames. To estimate the object motion alone, we first nullify the camera motion between the past frames by warping them to the same view using projective geometry. Decoupling camera and object motion makes the predicted object motion independent of the past or future camera motion, and thus we can synthesize future frames even when there is a change in the camera trajectory.

The depth of moving objects in a scene is usually different from that of their neighboring pixels. This difference can be exploited to better estimate the object motion by matching the points in 3D instead of 2D. Driven by this observation, we propose a method to estimate object motion in 3D, which we show to be more accurate than 2D motion estimation. It is also beneficial to use 3D motion estimation in occluded/disoccluded regions since such regions do not have matching points, and the motion estimation is guided by the neighborhood motion only. Occluded regions typically belong to the relative background, and hence motion in such regions is similar to that of the neighborhood background. Estimating motion in 3D can utilize this correlation to estimate better object motion.

We employ multi-plane images (MPI) as a 3D representation of the scenes, which represents the objects in the scene using multiple images placed at different depths. We choose the MPI representation since it can be directly processed by convolutional neural networks (CNN) and the frames can be reconstructed from MPI via differentiable alpha-compositing [161]. We estimate 3D motion as displacement vectors between the corresponding points on the MPIs by training a CNN in an unsupervised fashion. Since MPI representations are inherently sparse, we process the MPIs using partial convolution layers and employ masked correlations to compute the 3D cost volumes. We feed the 3D cost volumes to the subsequent partial convolution layers, which estimate the displacement or flow vectors. Since the depth dimension in MPIs is discrete, we predict the motion in the depth dimension as a probability distribution over the depth planes. The expected value of this predicted distribution gives the displacement in the depth dimension.

We then incorporate the available camera motion to determine all locations in the predicted frame that can be reconstructed from the past frame. Employing a 3D infilling network similar to that of Srinivasan et al. [161], we synthesize the regions which are newly uncovered in the predicted frames. We dub our model as DeCOMPnet since we explicitly decompose the motion into camera and object motion for predicting the next frame.

Since most view synthesis and video prediction datasets do not satisfy the problem assumptions for TVS-DS, we develop a new challenging dataset named the Indian Institute of Science Virtual Environment Exploration Dataset - Dynamic Scenes (IISc VEED-Dynamic). Our dataset contains 800 videos with 12 frames per video with a wide variety of camera and object motion. We render the videos using Blender at full HD resolution and a frame rate of 30fps. We evaluate our model and benchmark other video prediction and view synthesis models on our dataset and the MPI-Sintel [26] dataset for frame-rate upsampling. We show that our model achieves state-of-the-art performance in terms of the quality of the predicted frames. We further upper bound the performance of our model components using an oracle that has knowledge of the future frames.

We summarize our main contributions as follows:

- We formulate a framework for temporal view synthesis of dynamic scenes that uses the available user or camera motion and only predicts the object motion.

- We design a 3D motion estimation model using an MPI representation of past frames after nullifying the camera motion between them. We introduce masked correlation and partial convolution layers to handle sparsity in the MPI representation.

- We develop a challenging dataset, IISc VEED-Dynamic, consisting of 800 videos at full HD resolution to evaluate our algorithm. We show that our model outperforms other competing models on both MPI-Sintel and our datasets.

## 1.2 Organization

The rest of this thesis is organized as follows. In Chapter 2, we provide a discussion of the related work in the areas of novel view synthesis, dynamic view synthesis and motion estimation. Chapter 3 and Chapter 4 describe our approach to regularizing sparse input NeRFs with visibility priors and simpler solutions respectively. We study the design of fast and compact motion field and reliable motion priors for dynamic radiance fields in Chapter 5. We present our approach to temporal view synthesis of dynamic scenes in Chapter 6. Finally, we discuss future directions and conclude the thesis in Chapter 7. The codes for all our models and the databases we developed are available in the respective project pages that can be accessed from `https://nagabhushansn95.github.io/publications.html`.

# Chapter 2

# Related Work

The prior work related to this thesis can be broadly classified into three parts. In the first part, we review the prior work on novel view synthesis of static scenes, followed up by dynamic scenes in the second part. Finally, we review the techniques related to temporal view synthesis in the third part.

## 2.1   Novel View Synthesis of Static Scenes

Chen and Williams [36] introduce the problem of novel view synthesis and propose an image-based rendering (IBR) approach to synthesize novel views. The follow-up approaches introduce the geometry of the scene for synthesizing novel views through approximate representations such as light fields [94], lumigraphs [60], plenoptic functions [118] and layered depth images [149]. Chai et al. [30] study the minimum sampling needed for light field rendering and also show that depth information enables better view synthesis with sparse viewpoints. McMillan Jr [119] and Mark [115] introduce depth image based rendering (DIBR) to synthesize new views. Multiple variants of DIBR [32, 79, 167, 193] find use in various applications such as 3D-TV [50] and free-viewpoint video [29, 41, 154]. Ramamoorthi [139] conducts a detailed survey on classical work for novel view synthesis.

With the advent of deep learning, volumetric models utilize the power of learning by training the model on a large dataset of multi-view images. While the early approaches predict volumetric representations in each of the target views [52, 78], latter approaches predict a single volumetric representation and warp the representation to the target view while rendering [122, 134, 153, 161, 216]. Prominent among these approaches is the Multiplane Images (MPI) representation that represents the scene as a set of planar images at different depths [67, 122, 161, 172, 216]. However, these approaches employ discrete depth planes and hence suffer from discretization artifacts. The seminal work by Mildenhall et al. [123] employ a continuous representation using multi-layer perceptrons (MLP). This started a new pathway in neural view synthesis. However, these models suffer from two major limitations, namely, the need for the dense sampling of input views and the large time required to render novel views from the given input views. The prior work that address these limitations can be broadly classified into three categories. In Sec. 2.1.1, we review various approaches in the literature to regularize the NeRF when training with sparse input views. We review the explicit radiance fields that aim at fast optimization and rendering in Sec. 2.1.2, and also review the recent work on regularizing explicit models for the few-shot setting. Finally, in Sec. 2.1.3, we review the generalized NeRFs that address both issues jointly.

## 2.1.1 Implicit Radiance Fields

There exists extensive literature on regularizing scene-specific NeRFs when training with sparse inputs. Hence, we further group these models based on their approaches.

### 2.1.1.1 Hand-Crafted Depth Priors:

The prior work on sparse input NeRFs explore a plethora of hand-crafted priors on the NeRF rendered depth. RegNeRF [128] imposes a smoothness constraint on the rendered depth maps. DS-NeRF [46] uses sparse depth provided by a Structure from Motion (SfM) module to supervise the NeRF estimated depth at sparse keypoints. HG3-NeRF [59] uses sparse depth given by colmap to guide the sampling 3D points instead of supervising the NeRF rendered depth. While these priors are more robust across different scenes, they do not exploit the power of learning.

### 2.1.1.2 Deep Learning Based Depth Priors:

There exist multiple models that utilize the advances in dense depth estimation using deep neural networks. DDP-NeRF [142] extends DS-NeRF by employing a CNN to complete the sparse depth into dense depth for more supervision. SCADE [176] and SparseNeRF [185] use the depth map output by single image depth models to constrain the absolute and the relative order of pixel depths, respectively. DiffusioNeRF [198] learns the joint distribution of RGBD patches using denoising diffusion models (DDM) and utilizes the gradient of the distribution provided by the DDM to regularize NeRF rendered RGBD patches. However, the deep-learning based priors require pre-training on a large dataset and may suffer from generalization issues when obtaining the prior on unseen test scenes.

### 2.1.1.3 View Hallucination based Methods:

Another line of regularization based approaches simulate dense sampling by hallucinating new viewpoints and regularizing the NeRF on different aspects such as semantic consistency [75], depth smoothness [128], sparsity of mass [83] and depth based reprojection consistency [22, 35, 87, 201]. Instead of sampling new viewpoints randomly, Flip-NeRF [147] utilizes ray reflections to determine new viewpoints. Deceptive-NeRF [107] and ReconFusion [196] employ a diffusion model to generate images in hallucinated views and use the generated views in addition to the input views to train the NeRF. However, supervision with generative models could lead to content hallucinations, leading to poor fidelity [90].

### 2.1.1.4 Other regularizations:

A few models also explore regularizations other than depth supervision and view hallucinations. FreeNeRF [205] and MI-MLP-NeRF [220] regularize the NeRF by modifying the inputs. Specifically, FreeNeRF anneals the frequency range of positional encoded NeRF inputs as the training progresses, and MI-MLP-NeRF adds the 5D inputs to every layer of the NeRF MLP. MixNeRF [148] models the volume density along a ray as a mixture of Laplacian distributions. Philip and Deschaintre [135] scale the gradients corresponding to 3D points close to the camera when sampling the 3D points in inverse depth to reduce floaters close to the camera. VDN-NeRF [219] on the other hand, aims to resolve shape-radiance ambiguity in the case of dense input views. However, these approaches are designed for specific cases and are either sub-optimal or do not extend to more recent radiance field models.

## 2.1.2 Explicit Radiance Fields

The NeRF takes a long time to optimize and render novel views due to the need to query the NeRF MLP hundreds of times to render a single pixel. Hence, a common approach to fast optimization and rendering is to reduce the time taken per query. Early works such as PlenOctress [209] and KiloNeRF [140] focus on improving only the rendering time by baking the trained NeRF into an explicit structure such as Octrees or thousands of tiny MLPs. PlenOxels [54] and DVGO [165] reduce the optimization time by directly optimizing voxel grids, but at the cost of large memory requirements to store the voxel grids. TensoRF [33] and K-Planes [53] reduce the memory consumption using factorized tensors that exploit the spatial correlation of the radiance field. Alternately, iNGP [125] and ZipNeRF [18] employ multi-resolution hash-grids to reduce the memory consumption. Recently, 3DGS [81] propose an alternative volumetric model for real-time rendering of novel views. Specifically, 3DGS employs 3D Gaussians to represent the scene and renders a view by splatting the Gaussians onto the corresponding image plane. While the above methods enable fast optimization and rendering, their performance still reduces significantly with fewer input views.

### 2.1.2.1  Sparse Input Explicit Radiance Fields:

Recently, there is increasing interest in regularizing explicit models to learn with sparse inputs [97, 203]. However, the regularizations designed in these models are limited to a specific explicit radiance field and do not generalize to other explicit models. For example, ZeroRF [151] imposes a deep image prior [175] on the components of the TensoRF [33] model. FSGS [221] and SparseGS [200] improve the performance of 3DGS [81] in the sparse input case by improving the initialization of the 3D Gaussian point cloud and pruning Gaussians responsible for floaters respectively.

## 2.1.3  Generalized Sparse Input NeRF

Obtaining a volumetric model of a scene by optimizing the NeRF is a time-consuming process. In order to reduce the time required to obtain a volumetric model of a scene and learn with fewer input views, generalized NeRF models train a neural network on a large dataset of multi-view scenes that can be directly applied to a test scene without any optimization [34, 89, 168]. Early pieces of work such as PixelNeRF [210], GRF [171], and IBRNet [189] obtain convolutional features of the input images and additionally condition the NeRF by projecting the 3D points onto the feature grids. MVSNeRF [34] incorporates cross-view knowledge into the features by constructing a 3D cost volume. However, the resolution of the 3D cost volume is limited by the available memory size, which limits the performance of MVS-NeRF [100]. On the other hand, SRF [39] processes individual frame features in a pair-wise manner, and GNT [187] employs a transformer to efficiently incorporate cross-view knowledge.

NeuRay [108] and GeoNeRF [76] further improve the performance by employing visibility priors and a transformer respectively to effectively reason about the occlusions in the scene. More recent work such as GARF [152], DINER [137] and MatchNeRF [37] try to provide explicit knowledge about the scene geometry through depth maps and similarity of the projected features. Different from the above, MetaNeRF [168] learns the latent information as initial weights of the NeRF MLPs by employing meta-learning.

However, the need for pre-training on a large dataset of scenes with multi-view images and generalization issues due to domain shift have motivated researchers to continue to be interested in regularizing scene-specific radiance fields.

#### 2.1.3.1 Single Image NeRF

This approach of conditioning the NeRF on learned features is also popular among single image NeRF models [100, 201], which can be considered as an extreme case of the sparse input NeRF. A common thread in single image NeRF models is to use an encoder to obtain a latent representation of the input image. A NeRF based decoder conditioned on the representation, outputs volume density and color at given 3D points. For example, pix2NeRF [27] combines $\pi$-GAN [31] with NeRF to render photo-realistic images of objects or human faces. Gao et al. [57] focus on human faces alone and use a more structured approach by exploiting facial geometry. MINE [96] combines NeRF with MPI by replacing the MLP based implicit representation with an MPI based explicit representation in the decoder. Lin et al. [100] obtain a richer latent representation by fusing global and local features obtained using a vision transformer and CNN respectively. Different from the above models, Wimbauer et al. [194] use the MLP decoder to predict volume density alone and obtain the color by directly sampling from the given images. However, a common drawback of these models is the need for pre-training. Thus, the performance may be inferior when testing on a generic scene.

## 2.2 Novel View Synthesis of Dynamic Scenes

### 2.2.1 Classical Work on Deformation Models

The modeling of dynamic radiance fields through a static radiance field and a motion field is similar to the use of a deformation model that deforms a canonical representation of an entity. This is a popular approach to model deformable solids [146], human motion [109], facial expressions [23], deformable garments [121, 136], fluids [3, 61, 69], gases [8], smoke [69], flames [72] as well as CT (Computed Tomography) and MRI (Magnetic Resonance Imaging) scans in the medical field [129]. Learning the motion or deformation field is also found to improve the reconstruction of 3D scenes [178] and free-viewpoint rendering of dynamic scenes [29].

The canonical space in such deformable models is popularly represented as triangular meshes estimated through a multi-view stereo algorithm [5, 23, 24, 85], or specialized parametric meshes for humans and faces [6, 109, 181], or sums of Gaussians [163]. In contrast, we employ a radiance field to learn the canonical space owing to its advantages over meshes [123]. Different deformation models include free-form deformation (FFD) fields or optical flows [48, 186], bijective mappings [84], morphable models [21], splines [182], volumetric Laplacian deformations [45] and linear systems with basis functions [72]. Our motion field is closer to the free-form deformation among the above deformation approaches, but differs in the representation used for the motion field. Finally, the sparse flow priors we employ can be thought of as similar to tracking the markers [64, 68] or keypoints [24, 77] while estimating the classical deformation models.

## 2.2.2  Dynamic View Synthesis

Different from the use of deformable models, Zitnick et al. [222] learn the 3D scene dynamics using a layered depth representation with motion compensation. More recent volumetric representations such as Multiplane Images (MPI) [216] are extended to handle dynamic scenes by employing an MPI per frame [101] or temporal basis functions [199]. To handle 360° scenes, Broxton et al. [25] replace MPI with multi-sphere images (MSI). However, such approaches suffer from depth discretization artifacts [123]. Instead of volumetric models, Yoon et al. [208] employ depth image based warping, where the depth is obtained by combining monodepth with multi-view depth. Recent work on dynamic view synthesis with sparse input viewpoints require depth information obtained through the use of RGB-D cameras [95] or multi-view stereo [11]. Further, such approaches struggle to handle soft edges and translucent objects [134].

### 2.2.3 Dynamic Radiance Fields

In contrast to the classical approaches, dynamic view synthesis can be solved by learning a 6D radiance field that maps the position, time and viewing direction to the radiance. However, the performance of such approaches degrades when the input viewpoints are sparse. We provide a quick overview of prior work that share a few attributes of our model in Tab. 5.1. To the best of our knowledge, ours is the first work to address dynamic radiance fields with few input viewpoints while achieving fast training and rendering.

Dynamic radiance fields can be broadly classified into two categories based on how the temporal modeling is handled, which is crucial in the sparse input setting. A simple approach is to model the dynamic radiance field as a 6D function of position, time, and viewing direction [56]. To utilize flow priors, NSFF [99] predicts flow as an auxiliary task and constrains the volume density of the mapped 3D points provided by the flow. K-Planes [53] and HexPlane [28] extend TensoRF [33] to a 4D model that maps the position and time to a latent feature, which is then decoded by a tiny multi-layer perceptron (MLP). Instead of conditioning the radiance field directly on time, DyNeRF [98] conditions the radiance field on a per-time-instant feature vector which is jointly optimized with the radiance field. The lack of a motion model in these approaches makes it incompatible to impose motion priors when learning with sparse input viewpoints.

The second set of models employs a motion or deformation field that maps the 3D points from a given time instant to a canonical time instant [138, 184, 188]. TiNeuVox [49], SWAGS [150] and CoGS [211] replace the scene representation MLP in D-NeRF [138] with a TensoRF or 3DGS model, but use MLPs to model the motion field, leading to expensive training and rendering time. To achieve fast optimization and rendering, DeVRF [104] employs a 4D voxel grid to learn the deformation field, but its memory requirements scale with the fourth power of the grid resolution. Instead of employing a free-form deformation model, Wang et al. [183] models the motion using discrete cosine transform (DCT) basis functions. Different from the above, prior [63] and concurrent work [195] employ a motion model that maps the 3D points from the canonical time instant to a given time instant. However, the forward warping of points can lead to holes in novel views, which may need to be infilled using a separate inpainting network [63, 124, 158].

## 2.3 Temporal View Synthesis

Recall that in TVS-DS, we aim to predict the future frames given the past frames and the camera poses of the past and future frames. We now review the prior work on techniques required to solve the temporal view synthesis of dynamic scenes. Specifically, we review the prior art on Video Prediction and motion estimation. Further, TVS-DS also involves infilling disocclusions in the scene caused by the motion of the camera and objects. Thus, we also review existing inpainting techniques.

**Video Prediction:** Deep video prediction was initially proposed as a self-supervised approach for representation learning of videos [162]. Video prediction has also found diverse applications such as robotic path planning [51], anomaly detection [106], video compression [102] and autonomous driving [110]. Various video prediction approaches include multiscale prediction [117], predictive coding [110], decomposing video into motion and content [174, 180], decoupling motion of background and foreground objects [197], decomposing motion into velocity and acceleration maps [144], action conditioned prediction [88] and so on. DPG [58], which disentangles motion propagation and content generation, is closely related to our work. However, our approach differs in decomposing the motion into camera motion and object motion and estimating object motion in 3D using MPIs.

To account for the uncertainty of the future in long term prediction, stochastic video prediction models [9, 47, 179] aim to predict multiple future motion-trajectories for a given past. A detailed review of video prediction models can be found in [130]. However, video prediction models, in general, do not use camera motion and depth available in temporal view synthesis. In contrast, temporal view synthesis deals with the question of how to use camera motion and only predict the local motion of objects.

**Motion Estimation:** One of the most popular techniques for motion estimation is optical flow. Optical flow estimation is a classical problem [70, 112] which has found renewed interest due to the success of deep neural networks [105, 166]. In contrast to 2D based optical flow, scene flow estimates the motion in 3D. Recently, Yang et al. [204] estimate scene flow by expanding 2D optical flow to 3D using camera geometry. Our 3D motion estimation differs from the above through the use of the 3D MPI representation.

**Inpainting:** Depth image based rendering (DIBR) models employ the popular warp-and-infill approach and focus on infilling the disocclusions [40, 114]. Luo et al.[113] detect and remove foreground objects, reconstruct the background to infill the disocclusions and then apply motion compensation. Srinivasan et al. [161] propose to infill in the 3D MPI representation by copying the radiance from the background planes. Several image and video inpainting algorithms exist in the literature including classical [14, 44, 192] and deep learning [73, 82, 91, 126, 133, 202, 212] based models. Recently, Kanchana et al. [79] consider the problem of temporal view synthesis for static scenes. Specifically, they follow the DIBR approach to synthesize the next frame and predict infilling vectors using a deep neural network to infill the discussions in 2D. In dynamic scenes, both the camera and object motion can create disocclusions in the next frame, which need to be infilled.

# Chapter 3

# ViP-NeRF: Visibility Prior for Sparse Input Neural Radiance Fields

## 3.1 Introduction

The key challenge with sparse input NeRF for static scenes is that the volume rendering equations in NeRF are under-constrained, leading to solutions that overfit the input views. This results in uncertain and inaccurate depth in the learned representation. Synthesized novel views in such cases contain extreme distortions such as blur, ghosting, and floater artifacts [128, 142]. Recent works have proposed different approaches to constrain the training of NeRF to output visually pleasing novel views. While a few recent works [207, 213, 218] focus on training NeRF models on a specific category of objects such as chairs or airplanes, we focus on training category agnostic sparse input NeRF models [128]. Such prior work can be broadly classified into generalized NeRF models and other regularization approaches.

In generalized NeRFs, the NeRF is additionally conditioned on a latent scene representation obtained using a convolutional neural network [34, 66, 76, 108, 137, 187, 189, 210]. The latent prior helps overcome the limitation on the number of views by enabling the NeRF model to effectively understand the scene. Such an approach is popular even when only a single image of the scene is available as input to the NeRF [27, 100, 201]. However, these models require a large multi-view dataset for pre-training and may suffer from generalization issues when used to render a novel scene [128]. Thus, we believe that there is a need to study the sparse-input NeRF without conditioning the NeRF on latent representations.

The other thread of work on sparse input NeRFs follows the original NeRF paradigm of training scene-specific NeRFs, and designs novel regularizations to assist NeRFs in converging to a better scene geometry [62, 127, 213]. One popular approach among such models is to supervise the depth estimated by the NeRF. RegNeRF [128] uses a depth smoothness prior to supervise the depth estimated by the NeRF. DS-NeRF [46] uses a sparse depth prior obtained from a Structure from Motion (SfM) model such as Colmap [145]. However, this prior is available at sparse keypoints only and hence does not sufficiently constrain the NeRF. On the other hand, DDP-NeRF [142] and SCADE [176] pre-train convolutional neural networks (CNN) on a large dataset of scenes to learn a dense depth prior. These approaches may also suffer from issues similar to those of the generalized models. This motivates the exploration of other reliable features for dense supervision to constrain the NeRF in addition to sparse depth supervision. In this chapter, we present our approach of regularizing the NeRF using a dense prior on the visibility of pixels in a pair of views.

## 3.2  NeRF Preliminaries

We first provide a brief introduction to NeRF and define the notations for subsequent use. A neural radiance field is an implicit representation of a scene using two multi-layer perceptrons (MLP). Given a set of images of a scene with corresponding camera poses, a pixel $\mathbf{q}$ is selected at random, and a ray $\mathbf{r}$ is passed from the camera center $\mathbf{o}$ through $\mathbf{q}$. Let $\mathbf{p}_1, \mathbf{p}_2, \ldots, \mathbf{p}_N$ be $N$ randomly sampled 3D points along $\mathbf{r}$. If $\mathbf{d}$ is the direction vector of $\mathbf{r}$ and $z_i$ is the depth of a 3D point $\mathbf{p}_i$, $i \in \{1, 2, \ldots, N\}$, then $\mathbf{p}_i = \mathbf{o} + z_i \mathbf{d}$. An

MLP $\mathcal{F}_1$ is trained to predict the volume density $\sigma_i$ at $\mathbf{p}_i$ as

$$\sigma_i, \mathbf{h}_i = \mathcal{F}_1(\mathbf{p}_i), \tag{3.1}$$

where $\mathbf{h}_i$ is a latent representation. A second MLP $\mathcal{F}_2$ then predicts the color using $\mathbf{h}_i$ and the viewing direction $\mathbf{v} = \mathbf{d}/\|\mathbf{d}\|$ as

$$\mathbf{c}_i = \mathcal{F}_2(\mathbf{h}_i, \mathbf{v}). \tag{3.2}$$

Let the distance between two consecutive samples $\mathbf{p}_i$ and $\mathbf{p}_{i+1}$ be $\delta_i = z_{i+1} - z_i$. The visibility or transmittance of $\mathbf{p}_i$ is then given by

$$T_i = \exp\left(-\sum_{j=1}^{i-1} \delta_j \sigma_j\right). \tag{3.3}$$

The weight or contribution of $\mathbf{p}_i$ in rendering the color $\hat{\mathbf{c}}$ of pixel $\mathbf{q}$ is computed as

$$w_i = T_i\left(1 - \exp(-\delta_i \sigma_i)\right) \tag{3.4}$$

to obtain

$$\widehat{\mathbf{c}} = \sum_{i=1}^{N} w_i \mathbf{c}_i. \tag{3.5}$$

The MLPs are trained using mean squared error loss with the true color $\mathbf{c}$ of $\mathbf{q}$ as

$$\mathcal{L}_{mse} = \|\mathbf{c} - \widehat{\mathbf{c}}\|^2. \tag{3.6}$$

## 3.3   Method

We illustrate the outline of our model in Fig. 3.1. The core idea of our work is that when only a few multiview images are available for NeRF training, the visibility of a pixel in different views can be more reliably densely estimated as compared to its absolute depth. In this regard, we introduce visibility regularization to train the NeRF with sparse input views in Sec. 3.3.1. To impose the visibility regularization, we obtain a binary visibility prior map for every pair of input training images, which we explain in Sec. 3.3.2. Finally, to reduce the training time, we design a method to efficiently predict the visibility of a given pixel in different views in Sec. 3.3.3. Sec. 3.3.4 summarizes the various loss functions used in training our model.

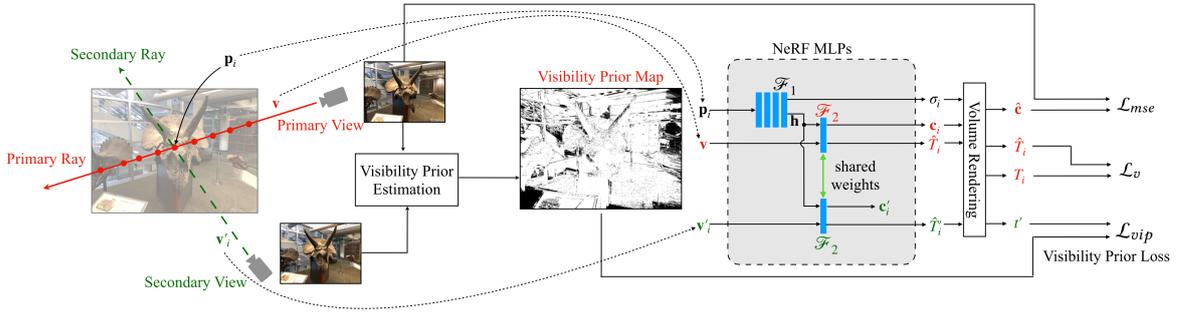

Figure 3.1: Overview of ViP-NeRF architecture. Given the images from <span style="color:red">primary</span> and <span style="color:green">secondary</span> views, we estimate a visibility prior map in the primary view and use it to supervise the visibility of pixels as predicted by the NeRF. Specifically, we cast a ray through a randomly selected pixel in the primary view and sample 3D points along the ray. For every point $\mathbf{p}_i$, we use the NeRF MLPs to obtain its visibility in primary and secondary views, along with volume density $\sigma_i$ and color $\mathbf{c}_i$. Volume rendering outputs visibility $t'$ of the chosen pixel in the secondary view which is supervised by the visibility prior. $\mathcal{L}_v$ constrains the visibilities $\hat{T}_i$ output by network and $T_i$ computed using volume rendering to be consistent with each other.

### 3.3.1   Visibility Regularization

Recall from Sec. 3.2 that NeRF trains MLPs by picking a random pixel $\mathbf{q}$ and predicting the color of $\mathbf{q}$ using the MLPs and volume rendering. Without loss of generality, we refer to the view corresponding to the ray $\mathbf{r}$ passing through $\mathbf{q}$ as the primary view and choose any other view as a secondary view. NeRF then samples $N$ candidate 3D points, $\mathbf{p}_1, \mathbf{p}_2, \ldots, \mathbf{p}_N$, along $\mathbf{r}$. Let $T_i'$ be the visibility of $\mathbf{p}_i$ from the secondary view, computed similar to Eq. (3.3). We define the visibility of pixel $\mathbf{q}$ in the secondary view, $t'(\mathbf{q})$, as the weighted visibilities of all the candidate 3D points $\mathbf{p}_i$ analogous to Eq. (3.5) as

$$t'(\mathbf{q}) = \sum_{i=1}^{N} w_i T_i' \quad \in [0, 1],  \tag{3.7}$$

where $w_i$ are obtained through Eq. (3.4). We omit the dependence of $w_i$ and $T_i'$ on $\mathbf{q}$ in the above equation for ease of reading. We obtain a prior $\tau'(\mathbf{q}) \in \{0, 1\}$ on the visibility $t'(\mathbf{q})$ as described in Sec. 3.3.2. We constrain the visibility $t'(\mathbf{q})$ to match the prior $\tau'(\mathbf{q})$. However, we find that the prior may be unreliable at pixels where $\tau' = 0$, as we describe in Sec. 3.3.2. Hence, we do not impose any visibility loss on such pixels and formulate our visibility prior loss as

$$\mathcal{L}_{vip}(\mathbf{q}) = \max(\tau'(\mathbf{q}) - t'(\mathbf{q}), 0).  \tag{3.8}$$

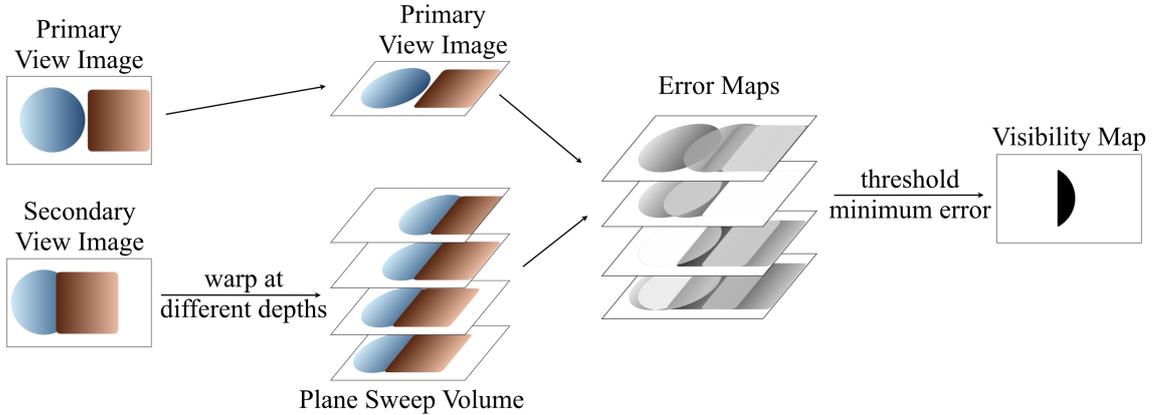

Figure 3.2: A toy example to illustrate the computation of visibility prior. The scene contains a blue sphere and a brown box and the relative pose between the views is a translation in **x** direction. The secondary view image is warped to the primary view at different depth planes to create a PSV and compared with the primary view image to obtain error maps. We observe that the brown square and the blue circle are matched better in the second and third planes respectively leading to lower error (denoted as white) in the respective error maps. The minimum error across all the planes is thresholded to obtain the visibility prior map corresponding to the primary view image. The right portion of the sphere which is occluded in the secondary view image is denoted in black in the visibility map.

Note that our loss function constrains the NeRF across pairs of views, unlike previous works which regularize [128, 142] in a given view alone. We believe that this leads to a better regularization for synthesizing novel views.

### 3.3.2 Visibility Prior

Given primary and secondary views, our goal is to estimate whether every pixel in the primary view is also visible in the secondary view through a binary visibility prior $\tau'(\mathbf{q})$. We employ plane sweep volumes to compute the visibility prior. We illustrate the computation of the visibility prior with a toy example in Fig. 3.2. Here, we warp the image in the secondary view to the primary view using the camera parameters at different depths varying between the near depth $z_{\min}$ and far depth $z_{\max}$. We sample $D$ depths uniformly in inverse depth similar to StereoMag [216]. The set of warped images is referred to as plane sweep volume (PSV) [71].

Let $I^{(1)}$ be the image in the primary view and $I_k^{(2)}$ be the set of $D$ warped images, where $k \in \{0, 1, \ldots, D-1\}$ denotes the plane index. We then compute the error map $E_k$ of the warped secondary image with the primary image at each plane $k$ of the PSV as

$$E_k = \|I^{(1)} - I_k^{(2)}\|_1, \tag{3.9}$$

where the norm is computed across the color channels. We determine the visibility prior $\tau'$ for pixel $\mathbf{q}$ by thresholding the minimum error across all the planes as

$$
\begin{aligned}
e(\mathbf{q}) &= \min_k E_k(\mathbf{q}), \\
\tau'(\mathbf{q}) &= \mathbb{1}_{\{\exp(-e(\mathbf{q})/\gamma) > 0.5\}},
\end{aligned}
\tag{3.10}
$$

where $\gamma$ is a hyper-parameter.

Intuitively, for a given pixel $\mathbf{q}$, a lower error in any of the planes indicates the presence of a matching pixel in the secondary view, i.e. $\mathbf{q}$ is visible in the secondary view. Note that this holds true when the intensity of pixels does not change significantly across views, which is typical for most of the objects in real-world scenes [96]. Consequently, the absence of a matching point across all the planes may indicate that $\mathbf{q}$ is not visible in the secondary view or $\mathbf{q}$ belongs to a highly specular object whose color varies significantly across different viewpoints. Thus, our prior is used to regularize the NeRF only in the first case above i.e. the pixels for which we find a match. Following the above procedure, we obtain the visibility prior for every pair of images obtained from the training set, by treating either image in the pair as the primary or the secondary view.

### 3.3.3 Efficient Prediction of Visibility

Recall that imposing $\mathcal{L}_{vip}$ in Eq. (3.8) requires computing visibility $T_i'$ in the secondary view for every $\mathbf{p}_i$. A naive approach to compute $T_i'$ involves sampling up to $N$ points along a secondary ray from the secondary view camera origin to $\mathbf{p}_i$ and querying the NeRF MLP $\mathcal{F}_1$ for each of these points. Thus, obtaining $t'(\mathbf{q})$ in Eq. (3.7) requires upto $N^2$ MLP queries, which increases the training time making it computationally prohibitive. We overcome this limitation by reformulating the NeRF MLP $\mathcal{F}_2$ to also output a view-dependent visibility of a given 3D point as,

$$\mathbf{c}_i, \hat{T}_i = \mathcal{F}_2(\mathbf{h}_i, \mathbf{v}); \quad \mathbf{c}_i', \hat{T}_i' = \mathcal{F}_2(\mathbf{h}_i, \mathbf{v}_i'), \tag{3.11}$$

where $\mathbf{v}_i'$ is the viewing direction of the secondary ray. We use the MLP output $\hat{T}_i'$ instead of $T_i'$ in Eq. (3.7).

Note that to output $\hat{T}_i'$, we need not query $\mathcal{F}_1$ again and can reuse $\mathbf{h}_i$ obtained from Eq. (3.1). We only need to query $\mathcal{F}_2$ additionally and since $\mathcal{F}_2$ is a single layer MLP and significantly smaller than $\mathcal{F}_1$, the additional computational burden is negligible. Thus, directly obtaining the secondary visibility $\hat{T}_i'$ of $\mathbf{p}_i$ through Eq. (3.11) allows us to compute $t'(\mathbf{q})$ in Eq. (3.7) using only $N$ queries of the MLP $\mathcal{F}_1$, as opposed to $N^2$ queries in the naive approach.

However, the use of $\hat{T}_i'$ in place of $T_i'$ regularizes the NeRF training only if the two quantities are close to each other. Thus, we introduce an additional loss to constrain the visibility $\hat{T}_i$ output by $\mathcal{F}_2$ to be consistent with the visibility $T_i$ computed using Eq. (3.3) as

$$\mathcal{L}_v = \sum_{i=1}^{N} \left( \left( \text{SG}(T_i) - \hat{T}_i \right)^2 + \left( T_i - \text{SG}(\hat{T}_i) \right)^2 \right), \tag{3.12}$$

where $\text{SG}(\cdot)$ denotes the stop-gradient operation. The first term in the above loss function uses $T_i$ as a target and brings $\hat{T}_i$ closer to it. On the other hand, since $\hat{T}_i$ gets additionally updated directly based on the visibility prior, the second term helps transfer such updates to $\mathcal{F}_1$ more efficiently than backpropagation through $\mathcal{F}_2$.

### 3.3.4  Overall Loss

Similar to DS-NeRF [46], we also use the sparse depth given by an SfM model to supervise the NeRF as

$$\mathcal{L}_{sd} = \|z - \hat{z}\|^2, \tag{3.13}$$

where $z$ is the depth provided by the SfM model, $\hat{z} = \sum_i w_i z_i$ is the depth estimated by NeRF and $w_i$ are obtained in Eq. (3.4). Our overall loss for ViP-NeRF is a linear combination of the losses obtained in Eq. (3.6), Eq. (3.8), Eq. (3.12) and Eq. (3.13) as

$$\mathcal{L} = \lambda_1 \mathcal{L}_{mse} + \lambda_2 \mathcal{L}_{sd} + \lambda_3 \mathcal{L}_{vip} + \lambda_4 \mathcal{L}_v, \tag{3.14}$$

where $\lambda_1, \lambda_2, \lambda_3$ and $\lambda_4$ are hyper-parameters. We note that $\mathcal{L}_{vip}$ is always employed in conjunction with $\mathcal{L}_v$ to make the learning computationally tractable.

## 3.4  Experiments

### 3.4.1  Evaluation Setup

We conduct experiments on two different datasets, namely RealEstate-10K and NeRF-LLFF. We evaluate all the models in the more challenging setup of 2, 3, or 4 input views, unlike prior work which use 9–18 input views [75, 142]. The test set is retained to be the same across all different settings for both datasets.

RealEstate-10K [216] dataset is commonly used to evaluate view synthesis models [67, 172] and contains videos of camera motion, both indoor and outdoor. The dataset also provides the camera intrinsics and extrinsics for all the frames. For our experiments, we choose 5 scenes from the test set, each containing 50 frames with a spatial resolution of $1024 \times 576$. In each scene, we reserve every $10^{\text{th}}$ frame for training and use the remaining 45 frames for testing.

NeRF-LLFF [122] dataset is used to evaluate the performance of various NeRF Models including sparse input NeRF models. It consists of 8 forward-facing scenes with a variable number of frames per scene at a spatial resolution of $1008 \times 756$. Following RegNeRF [128], we use every $8^{th}$ frame for testing. For training, we pick 2, 3 or 4 frames uniformly among the remaining frames following RegNeRF [128].

*Evaluation measures.* We quantitatively evaluate the methods using LPIPS [215], structural similarity (SSIM) [190], and peak signal to noise ratio (PSNR) measures. For LPIPS, we use the v0.1 release with the AlexNet [86] backbone as suggested by the authors.

### 3.4.2 Comparisons and Implementation Details

We compare the performance of our model with other sparse input NeRF models such as DDP-NeRF [142] and DietNeRF [75] which use learned priors to constrain the NeRF training. We also compare with DS-NeRF [46], InfoNeRF [83], and RegNeRF [128] that do not use learned priors. We train the models for 50k iterations on both datasets using the code provided by the respective authors.

For ViP-NeRF, we use Adam optimizer with a learning rate of 5e-4 that exponentially decays to 5e-6 following NeRF [123]. We set the loss weights such that the magnitudes of all the losses are of similar order after scaling. Specifically, we set $\lambda_1 = 1, \lambda_2 = 0.1, \lambda_3 = 0.001$ and $\lambda_4 = 0.1$. For visibility prior estimation, we set $D = 64$ and $\gamma = 10$. Since we require $\hat{T}'_i$ to be close to $T'_i$ while using $\hat{T}'_i$ to compute $\mathcal{L}_{vip}$, we impose $\mathcal{L}_{vip}$ after 20k iterations. We train our models on a single NVIDIA RTX A4000 16GB GPU.

### 3.4.3 Results

We show the quantitative performance of ViP-NeRF and other competing models on RealEstate-10K and NeRF-LLFF datasets in Tabs. 3.1 and 3.2. Our model outperforms

Table 3.1: Quantitative results on RealEstate-10K dataset.

| Model | learned prior | 2 views | | | 3 views | | | 4 views | | |
|---|---|---|---|---|---|---|---|---|---|---|
| | | LPIPS ↓ | SSIM ↑ | PSNR ↑ | LPIPS ↓ | SSIM ↑ | PSNR ↑ | LPIPS ↓ | SSIM ↑ | PSNR ↑ |
| InfoNeRF | | 0.6796 | 0.4653 | 12.30 | 0.6979 | 0.4024 | 11.15 | 0.6745 | 0.4298 | 11.52 |
| DietNeRF | ✓ | 0.5730 | 0.6131 | 15.90 | 0.5365 | 0.6190 | 16.60 | 0.5337 | 0.6282 | 16.89 |
| RegNeRF | | 0.5307 | 0.5709 | 16.14 | 0.4675 | 0.6096 | 17.38 | 0.4831 | 0.6068 | 17.46 |
| DS-NeRF | | 0.4273 | 0.7223 | 21.40 | 0.3930 | 0.7554 | 23.73 | 0.3961 | 0.7575 | 24.24 |
| DDP-NeRF | ✓ | 0.2527 | 0.7890 | 21.44 | 0.2240 | 0.8223 | 23.10 | 0.2190 | 0.8270 | 24.17 |
| ViP-NeRF | | **0.1704** | **0.8087** | **24.48** | **0.1441** | **0.8505** | **27.21** | **0.1386** | **0.8588** | **28.13** |

Table 3.2: Quantitative results on NeRF-LLFF dataset.

| Model | learned prior | 2 views | | | 3 views | | | 4 views | | |
|---|---|---|---|---|---|---|---|---|---|---|
| | | LPIPS ↓ | SSIM ↑ | PSNR ↑ | LPIPS ↓ | SSIM ↑ | PSNR ↑ | LPIPS ↓ | SSIM ↑ | PSNR ↑ |
| InfoNeRF | | 0.7561 | 0.2095 | 9.23 | 0.7679 | 0.1859 | 8.52 | 0.7701 | 0.2188 | 9.25 |
| DietNeRF | ✓ | 0.7265 | 0.3209 | 11.89 | 0.7254 | 0.3297 | 11.77 | 0.7396 | 0.3404 | 11.84 |
| RegNeRF | | 0.4402 | 0.4872 | 16.90 | 0.3800 | 0.5600 | 18.62 | **0.3446** | 0.6056 | 19.83 |
| DS-NeRF | | 0.4548 | 0.5068 | 17.06 | 0.4077 | 0.5686 | **19.02** | 0.3825 | 0.6016 | **20.11** |
| DDP-NeRF | ✓ | 0.4223 | **0.5377** | **17.21** | 0.4178 | 0.5610 | 17.90 | 0.3821 | 0.5999 | 19.19 |
| ViP-NeRF | | **0.4017** | 0.5222 | 16.76 | **0.3750** | **0.5837** | 18.92 | 0.3593 | **0.6085** | 19.57 |

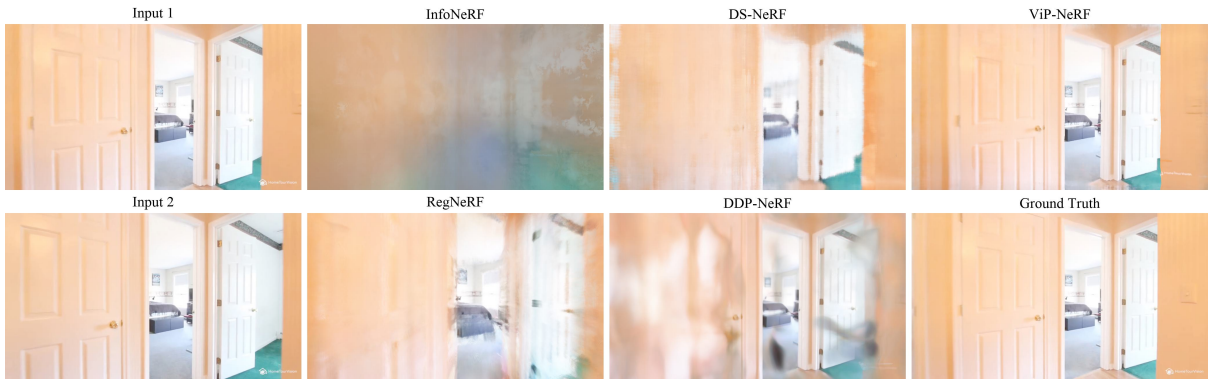

Figure 3.3: Qualitative examples on RealEstate-10K dataset with two input views. We observe that the predictions of ViP-NeRF are close to the ground truth, while those of other models suffer from various distortions. In particular, DDP-NeRF blurs regions of the frame near the left door and contains black floater artifacts.

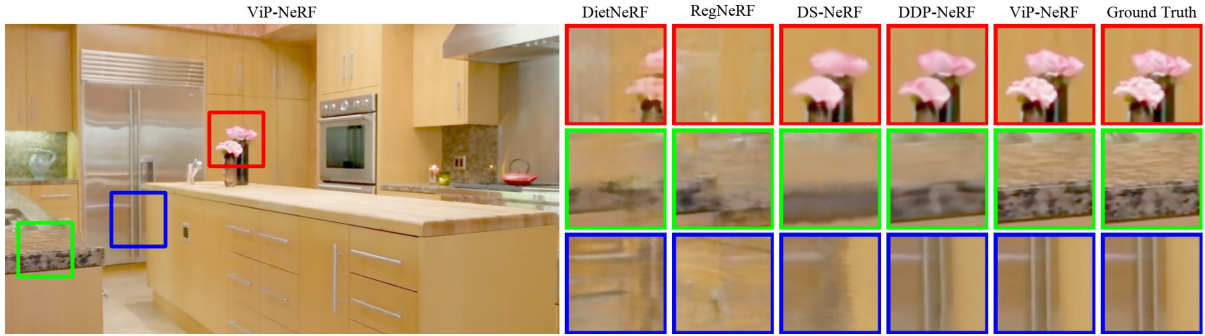

Figure 3.4: Qualitative examples on RealEstate-10K dataset with two input views. We observe sharp predictions by ViP-NeRF while predictions by other models suffer from blur and other artifacts. In particular, DDP-NeRF predictions contain blurred flowers (first row) and blurred tiles (second row).

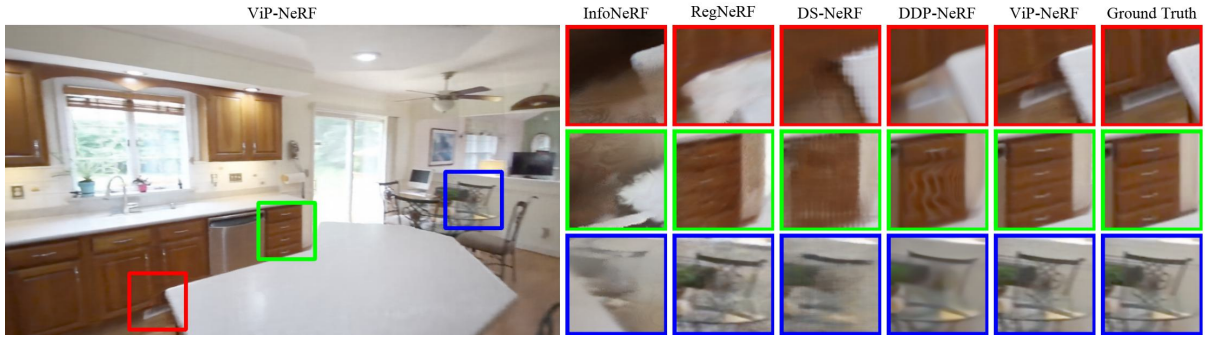

Figure 3.5: Qualitative examples on RealEstate-10K dataset with three input views. We find that ViP-NeRF is able to reconstruct novel views significantly better than the competing models. DDP-NeRF extends parts of the white table and fails to reconstruct the drawer handles accurately in the first and second examples. In the third example, DDP-NeRF fails to reconstruct thin objects in the chair.

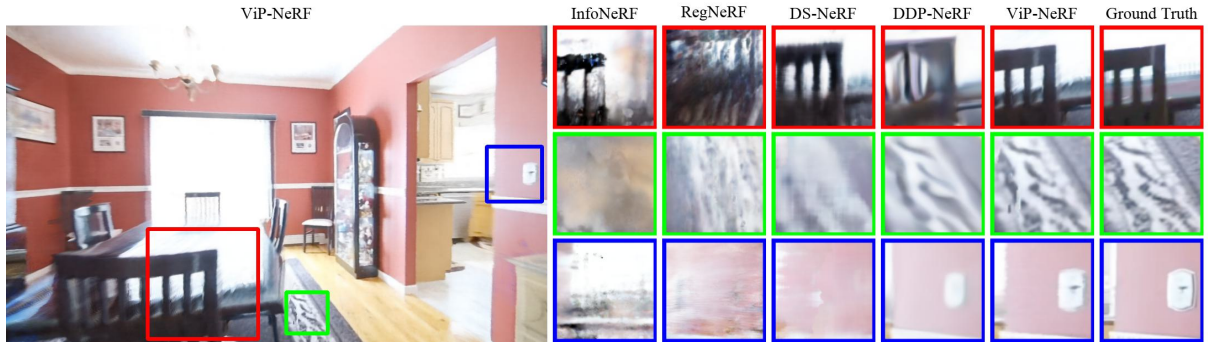

Figure 3.6: Qualitative examples on RealEstate-10K dataset with four input views. In the first example, DDP-NeRF fails to retain the structure of the chair while it blurs the texture of the carpet in the second example. We observe even more severe distortions among the predictions of other models.

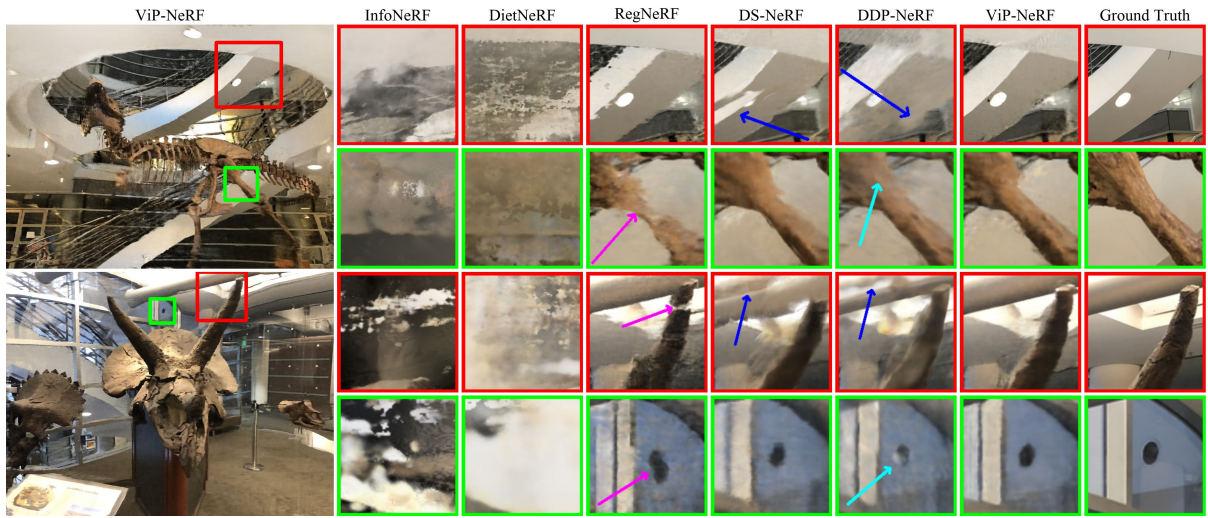

Figure 3.7: Qualitative examples on NeRF-LLFF dataset with two input views. In the first and third examples, we observe floater artifacts (blue arrows) in the predictions of DS-NeRF and DDP-NeRF, which are mitigated in the predictions of ViP-NeRF. We find that RegNeRF fails to capture thin t-rex bone in the second example and breaks the horn into two pieces in the third example (magenta arrows). Cyan arrows indicate color changes in the predictions of DDP-NeRF in the second and fourth examples. We note that predictions by our model do not suffer from the above described artifacts.

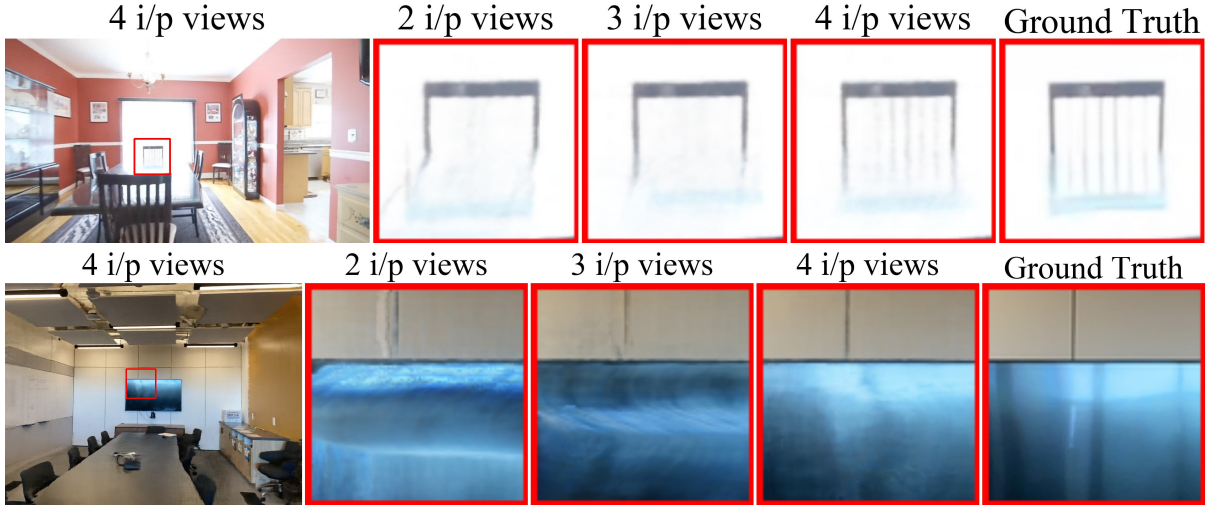

Figure 3.8: Qualitative examples on RealEstate-10K and NeRF-LLFF dataset with two, three, and four input views. We observe that ViP-NeRF models specular regions better as the number of input views increases. For example, in the first row, the reflection of the chair is better reconstructed as the number of views increases.

all the competing models, particularly in terms of the perceptual metric, LPIPS. ViP-NeRF even outperforms models such as DDP-NeRF and DietNeRF which involve pretraining on a large dataset. Fig. 3.3 shows qualitative comparisons on a scene from the RealEstate-10K dataset, where we observe significantly better synthesis by our model as compared to the competing models. We show more qualitative comparisons in Figs. 3.4 to 3.7. In these samples, we find that ViP-NeRF removes most of the floater artifacts and successfully retains the shapes of objects. Video comparisons are available on our project webpage[1].

In Fig. 3.8, we qualitatively compare the predictions of our model with different numbers of input views. We observe that ViP-NeRF estimates the geometry reasonably well with even two input views. However, with more input views, the performance of ViP-NeRF improves in reflective or specular regions. Fig. 3.9 visualizes the visibility map predicted by ViP-NeRF, where we observe that it is able to accurately predict the regions in the primary image which are visible and occluded in the secondary image.

---

| Primary View | Primary View | Secondary View | ViP-NeRF |

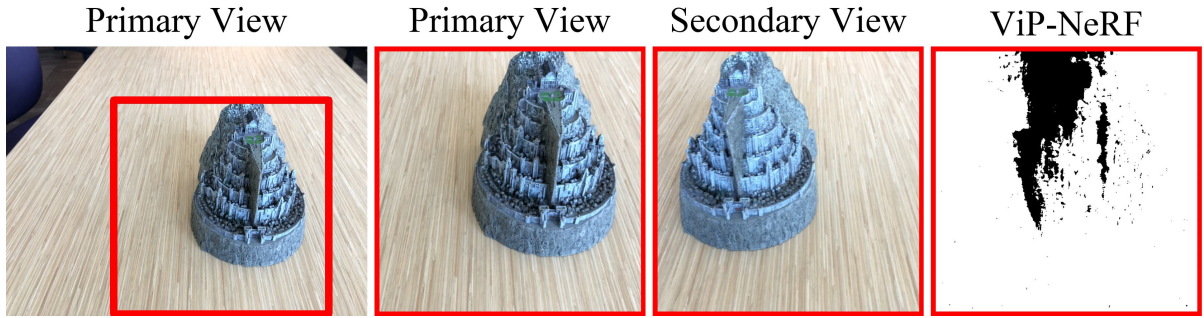

Figure 3.9: Visualization of the visibility map predicted by ViP-NeRF. White indicates the regions of the 'Primary View' which are visible in the 'Secondary View' and black indicates the occluded regions. From the primary and secondary views, we observe that the left part of the fortress and the neighboring portion of the wood are hidden in the secondary view. ViP-NeRF is able to reasonably determine the visible and occluded regions.

Table 3.3: Comparison of reliability of priors used in different models. The reference visibility is obtained using NeRF trained with dense input views.

|         | RealEstate-10K |          |        | NeRF-LLFF |          |        |
|---------|----------------|----------|--------|-----------|----------|--------|
| model   | **Prec. ↑**    | **Rec. ↑** | **F1 ↑** | **Prec. ↑** | **Rec. ↑** | **F1 ↑** |
| ViP-NeRF | 0.97          | **0.83** | **0.89** | 0.82     | **0.85** | **0.83** |
| DDP-NeRF | **0.98**      | 0.53     | 0.66   | **0.86**  | 0.33     | 0.47   |

*Dense depth vs dense visibility.* The key idea of our work is that it may be possible to reliably estimate dense visibility than dense depth. From Tab. 3.1, we find that ViP-NeRF outperforms DDP-NeRF consistently, which indicates that the dense visibility prior we compute without any pre-training is superior to the learned dense depth prior used by DDP-NeRF. Further from Tab. 3.2, we observe that ViP-NeRF consistently improves over DS-NeRF in terms of LPIPS and SSIM, whereas DDP-NeRF does not. This may be due to the domain shift between the training dataset of DDP-NeRF and the LLFF dataset, resulting in no performance improvement over DS-NeRF. Thus, we conclude that augmenting sparse depth with dense visibility leads to better view synthesis performance than dense completion of the sparse depth. We further validate this conclusion by comparing the two priors in the following.

*Validating priors.* We compare the reliability of the dense visibility prior used in our model against the dense depth prior from DDP-NeRF. For this comparison, we convert the dense depth to visibility and compare it with the visibility prior of our approach. Specifically, we warp the image in the secondary view to the primary view using the dense depth prior and compute the visibility map similar to Eq. (3.10). We compare the visibility maps obtained using dense depth and our approach with the visibility map predicted by a NeRF model trained with dense input views. We evaluate the visibility maps in terms of precision, recall, and F1 score.

From Tab. 3.3, we observe that our approach significantly outperforms DDP-NeRF prior in terms of the recall and F1 score, while performing similarly in terms of precision. A high precision of our prior indicates that it makes very few mistakes when imposing $\mathcal{L}_{vip}$. On the other hand, a high recall shows that our prior is able to capture most of the visible regions where $\mathcal{L}_{vip}$ needs to be imposed. On the contrary, a low recall for the DDP-NeRF prior indicates that large regions that are actually visible in the secondary view are marked as occluded by the dense depth prior. Consequently, this indicates the presence of a large number of pixels with inaccurate depth in the prior of DDP-NeRF. Thus, we conclude that our visibility prior is more reliable than the dense depth prior from DDP-NeRF for training the NeRF.

Table 3.4: Evaluation of depth estimated by different models with two input views. The reference depth is obtained using NeRF trained with dense input views. The depth RMSE on the two datasets are of different orders on account of different depth ranges.

| model | RealEstate-10K | | NeRF-LLFF | |
|---|---|---|---|---|
| | RMSE ↓ | SROCC ↑ | RMSE ↓ | SROCC ↑ |
| ViP-NeRF | **1.6411** | **0.7702** | **45.6314** | **0.6184** |
| DDP-NeRF | 1.7211 | 0.7544 | 46.6268 | 0.6136 |

As discussed in Sec. 3.1, visibility is related to relative depth, and thus a prior on visibility only constrains the relative depth ordering of the objects. On the other hand, the dense depth prior constrains the absolute depth, perhaps incorrectly. Thus the visibility prior provides more freedom to the NeRF in reconstructing the 3D geometry and is also more reliable compared to the depth prior. This may explain the superior performance of visibility regularization over dense depth regularization.

*Evaluation of estimated depth.* It is believed that better performance in synthesizing novel views is directly correlated with the accuracy of depth estimation [46]. Thus, we compare our model with DDP-NeRF on their ability to estimate absolute depth correctly using root mean squared error (RMSE). We also evaluate the models on their ability to estimate the relative depth of the scene correctly using spearman rank-order correlation coefficient (SROCC) [43], which computes the linear correlation between ranks of the estimated pixel depths with that of the ground truth depth. Due to the unavailability of ground truth depth on both the datasets, we train a NeRF model with dense input views and use its predicted depth as a pseudo ground truth. From Tab. 3.4, we observe that our model consistently outperforms DDP-NeRF both in terms of absolute and relative depth. Fig. 3.10 shows that the depth estimated by DDP-NeRF is smooth in textured regions, which may be leading to blur in the synthesized frame. In contrast, the dense visibility prior used in our model allows NeRF to predict sharp depth in such regions leading to sharper frame predictions.

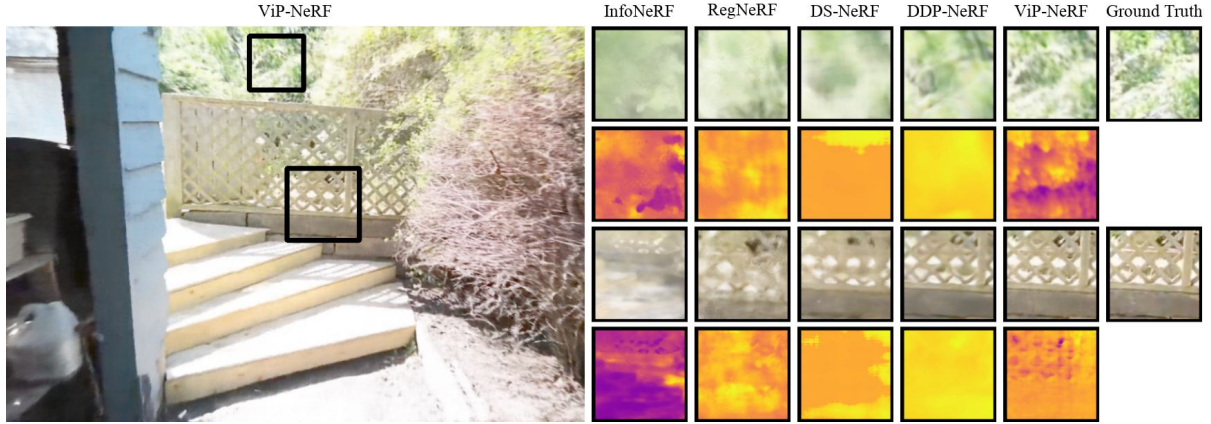

Figure 3.10: Estimated depth map on RealEstate-10K dataset with two input views. We find that ViP-NeRF is better in both frame synthesis and depth estimation compared to the competing models. For example, in the first row, the depth estimated by DDP-NeRF is smooth which may be leading to a loss of sharpness in synthesizing the shrubs. In contrast, ViP-NeRF predictions are sharper. For better visualization, we show inverse depth and normalize it to set the maximum value to unity.

Table 3.5: Ablation experiments on both the datasets with two input views.

| model | RealEstate-10K | | NeRF-LLFF | |
|---|---|---|---|---|
| | **LPIPS ↓** | **SSIM ↑** | **LPIPS ↓** | **SSIM ↑** |
| ViP-NeRF | **0.1704** | **0.8087** | **0.4017** | **0.5222** |
| w/o sparse depth | 0.2754 | 0.7588 | 0.5056 | 0.4631 |
| w/o dense visibility | 0.4273 | 0.7223 | 0.4548 | 0.5068 |

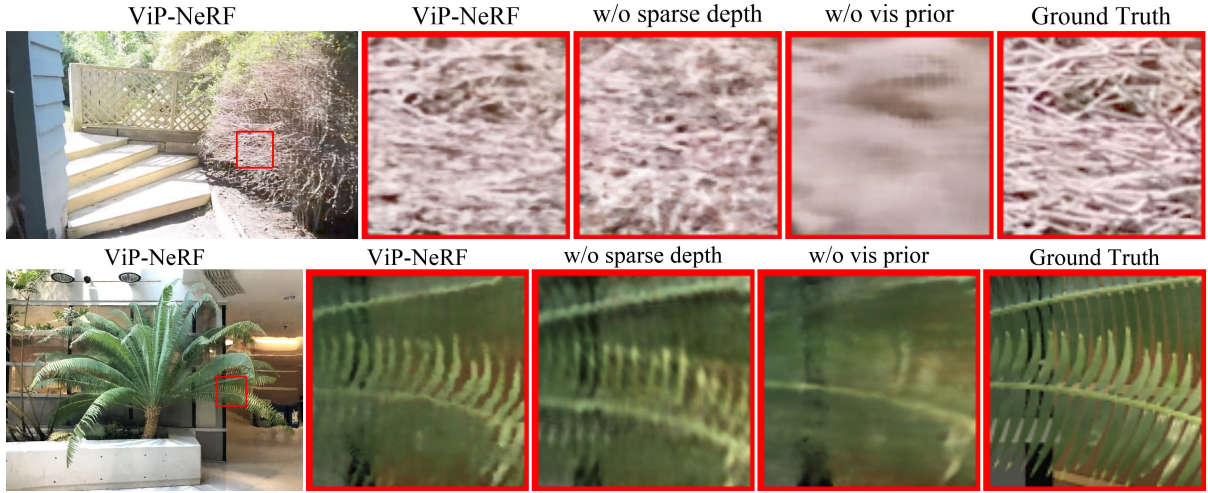

(a) Qualitative examples for ablations on RealEstate-10K and NeRF-LLFF dataset. We observe that the absence of dense visibility prior leads to significant blur in the predicted frames. While the reconstruction is reasonable without the sparse depth prior, we obtain the best reconstructions when using both the priors.

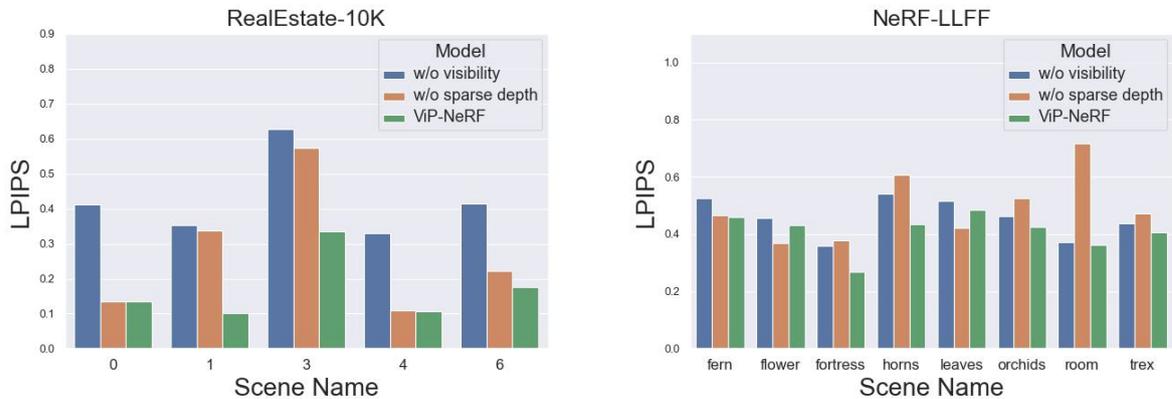

(b) Scene-wise LPIPS scores of ViP-NeRF and the ablated models. Note that lower LPIPS scores are better. ViP-NeRF performs better than both the ablated models in most cases leading to overall better performance.

Figure 3.11:  Qualitative and quantitative comparisons of ablated models on both RealEstate-10K and NeRF-LLFF datasets.

Table 3.6: Training times for various sparse input NeRF methods.

| Model | Training Time (hours) |
|---|---|
| InfoNeRF | 8 |
| RegNeRF | 10 |
| DS-NeRF | 8 |
| DDP-NeRF | 4 |
| ViP-NeRF | 8.5 |

*Ablations.* We analyze the contributions of dense visibility and sparse depth priors in ViP-NeRF, by disabling them one at a time. From Tab. 3.5 and Fig. 3.11a, we find that removing either priors leads to a drop in performance on both the datasets. This suggests that the dense visibility prior may be providing information that is complementary to the sparse depth prior. For a more fine-grained analysis, we compare the LPIPS scores on individual scenes in Fig. 3.11b. We observe that the addition of dense visibility prior over sparse depth prior leads to an improvement in the performance on all the scenes. Further, we find that our model with dense visibility prior alone is able to achieve impressive performance, especially on the RealEstate-10K dataset.

We now analyze the importance of the sparse depth prior against the dense visibility prior across different scenes. The performance of the NeRF with sparse depth prior is better than that of the visibility prior in the scenes where the sparse depth is available for a large number of pixels. Specifically, for highly textured scenes such as those in the NeRF-LLFF dataset, we can obtain the sparse depth for more pixels (762, 1045, and 1524 pixels on average with 2, 3, and 4 input views). On the other hand, for scenes with smooth objects such as walls, floor or ceiling in RealEstate-10K dataset, we can only obtain the sparse-depth prior for fewer pixels (328, 496, and 614 pixels on average with 2, 3, and 4 input views). In such scenes, the visibility prior tends to be more useful. Incorporating both the priors leads to the best performance in all the scenes.

*Comparison of training times.* Tab. 3.6 shows the training times for various sparse in-put NeRF methods. Since we use DS-NeRF as our base model and add only a single layer to predict visibility, we observe very little overhead as compared to DS-NeRF. However, DDP-NeRF employs a single NeRF instead of coarse and fine NeRFs and achieves the hierarchical sampling of 3D points using the estimated depth and its variance. Hence, DDP-NeRF requires lesser time for training.

## 3.5   Summary

We study the problem of training NeRFs in sparse input scenarios, where the NeRF tends to overfit the input views and learn incorrect geometry. We propose a prior on the visibility of pixels in other viewpoints to regularize the training and mitigate such errors. The visibility prior obtained using a plane sweep volume is more reliable as compared to the depth prior estimated using pre-trained networks. We reformulate the NeRF MLPs to additionally output visibility to compute the visibility prior loss in a time-efficient manner. ViP-NeRF outperforms prior work on two commonly used datasets for novel view synthesis.

# Chapter 4

# Simple-RF: Regularizing Sparse Input Radiance Fields with Simpler Solutions

## 4.1 Introduction

In Chapter 3, we find that a dense visibility prior is better than dense depth priors obtained using deep depth estimation networks for training Neural Radiance Fields (NeRF) with sparse input views. However, the visibility prior is not as rich as the depth prior. In this chapter, we explore learning scene-specific dense depth supervision that does not suffer from generalization issues. Further, NeRFs have been enhanced to optimize and render quickly [125], reduce aliasing artifacts [16], and learn on unbounded scenes [17]. However, all these models require tens to hundreds of images per scene to learn the scene geometry accurately, and their quality deteriorates significantly when only a few training images are available [75]. In this chapter, we focus on training both implicit radiance fields such as NeRF and explicit radiance fields such as TensoRF [33] and Zip-NeRF [18] with a sparse set of input images. We achieve this by designing augmented models that learn better depth supervision in certain regions of the scene and train these augmented models in tandem with the main radiance field. The depth estimated by these augmented models can be used to effectively supervise the main radiance field in the sparse input setting. Fig. 4.1 qualitatively shows the improvements achieved through our regularizations on NeRF, TensoRF and ZipNeRF on three different datasets.

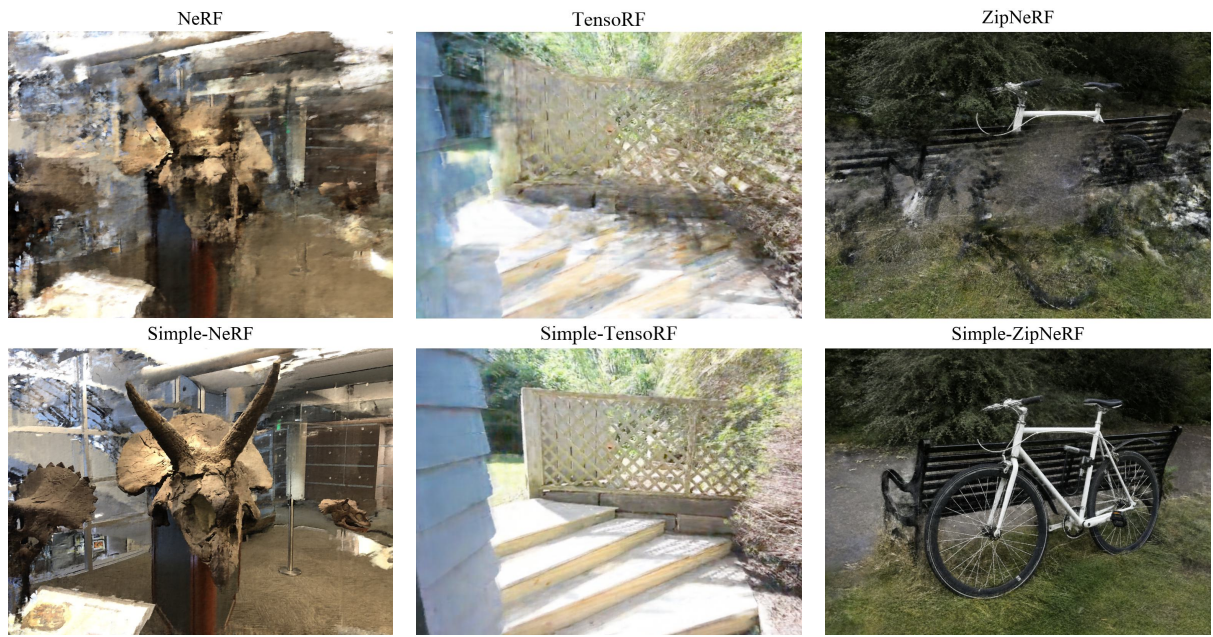

Figure 4.1: We show the improvements achieved by our regularizations on the NeRF, Tensor-RF and ZipNeRF models on NeRF-LLFF, RealEstate-10K and MipNeRF360 datasets respectively. We observe that the vanilla radiance fields suffer from various distortions. Regularizing the radiance fields with simpler solutions leads to significantly better reconstructions with all the three radiance fields.

We note that despite the recent work on sparse input 3DGS models, we do not explore designing augmentations for 3DGS. As noted in the recent literature, 3DGS mainly suffers from poor initialization with few input views [38]. We believe 3DGS requires a combination of good initialization and supervision from augmentations to learn from few input views. This necessitates a separate study on designing better initializations for 3DGS, which is beyond the scope of this work.

## 4.2 Radiance Fields and Volume Rendering Preliminaries

We first provide a brief recap of the radiance fields and volume rendering. We also describe the notation required for other sections in this chapter. To render a pixel $\mathbf{q}$, we shoot the corresponding ray into the scene and sample $N$ 3D points $\mathbf{p}_1, \mathbf{p}_2, \ldots, \mathbf{p}_N$, where $\mathbf{p}_1$ and $\mathbf{p}_N$ are the closest to and farthest from the camera, respectively. At every 3D point $\mathbf{p}_i$, the radiance field $\mathcal{F} = \mathcal{F}_1 \circ \mathcal{F}_2$ is queried to obtain a view-independent volume density $\sigma_i$ and a view-dependent color $\mathbf{c}_i$ as

$$\sigma_i, \mathbf{h}_i = \mathcal{F}_1(\mathbf{p}_i), \quad \mathbf{c}_i = \mathcal{F}_2(\mathbf{h}_i, \mathbf{v}),\tag{4.1}$$

where $\mathbf{v}$ is the viewing direction and $\mathbf{h}_i$ is a latent feature of $\mathbf{p}_i$. Volume rendering is then applied along every ray to obtain the color for each pixel as $\mathbf{c} = \sum_{i=1}^{N} w_i \mathbf{c}_i$, where

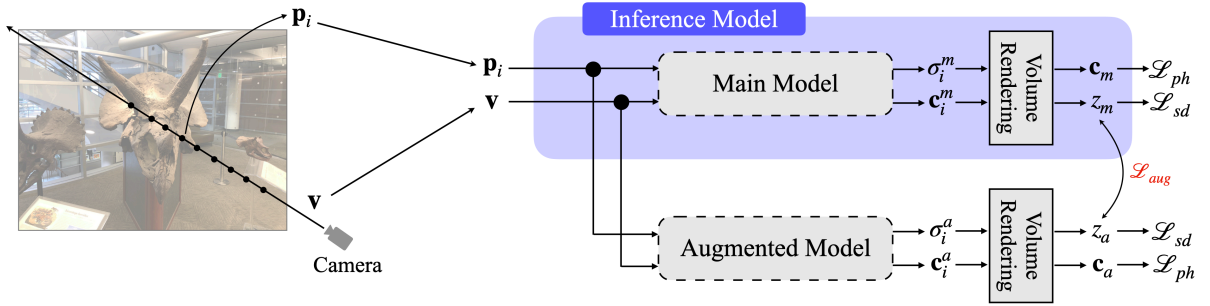

Figure 4.2: Architecture of Simple-RF family of models. We train the augmented model that only learns simpler solutions in tandem with the main model. The augmented models learn better depth in certain regions, which is propagated to the main model through the depth supervision loss $\mathcal{L}_{aug}$. During inference, only the Main Model is employed.

the weights $w_i$ are computed as

$$w_i = \exp\left(-\sum_{j=1}^{i-1} \delta_j \sigma_j\right) \cdot \left(1 - \exp\left(-\delta_i \sigma_i\right)\right),  \tag{4.2}$$

and $\delta_i$ is the distance between $\mathbf{p}_i$ and $\mathbf{p}_{i+1}$. The expected ray termination length is computed as $z = \sum_{i=1}^{N} w_i z_i$, where $z_i$ is the depth of $\mathbf{p}_i$. $z$ is typically also used as the depth of the pixel $\mathbf{q}$ [46]. $\mathcal{F}_1$ and $\mathcal{F}_2$ are modelled differently for NeRF, TensoRF and ZipNeRF, and are trained using the photometric loss $\mathcal{L}_{ph} = \|\mathbf{c} - \hat{\mathbf{c}}\|^2$, where $\hat{\mathbf{c}}$ is the true color of $\mathbf{q}$.

## 4.3  Method

Learning a radiance field with sparse input views leads to overfitting on the input views with severe distortions in novel views. Our key observation is that most of the distortions are due to the sub-optimal use of the high capabilities of the radiance field model. Further, we find that reducing the capability of the radiance field helps constrain the model to learn only simpler solutions, which can provide better depth supervision in certain regions of the scene. However, the lower capability models are not optimal either since they cannot learn complex solutions where necessary. Our solution here is to use the higher capability model as the main model and employ the lower capability models as augmentations to provide guidance on where to use simpler solutions. The challenge is that, it is not known apriori where one needs to employ supervision from the augmented model. We determine the more accurate model among the main and augmented models in terms of the estimated depth and use the more reliable depth to supervise the other. We note that the augmented models are employed only during the learning phase and not during inference. Thus, there is no additional overhead during inference. The augmented models are similar to the main model, but we modify their parameters to reduce their capability, and train them in tandem with the main model.

To design the augmented models, we first analyze the shortcomings of the radiance field with sparse input views. Specifically, we determine the components of the model that cause overfitting with fewer input views, causing distortions in novel views. We then design the augmented models by reducing the model capability with respect to such components. Thus, designing the augmented models is non-trivial, and the design may need to be different for different radiance fields based on the architecture of the radiance fields and the distortions observed. Nonetheless, the core idea of designing augmentations by reducing their capability to learn simpler solutions is common across all radiance fields.

We discuss the design of augmentations for NeRF, TensoRF and ZipNeRF in Secs. 4.3.1 to 4.3.3 respectively. We describe our approach to determining the reliability of the depth estimates in Sec. 4.3.1.4. Finally, Sec. 4.3.4 summarizes all the loss functions used to train our full model. Fig. 4.2 shows the architecture of our family of simple radiance fields.

### 4.3.1 Simple-NeRF

We start by discussing the specific details of the NeRF that are relevant for the design of augmentations in Sec. 4.3.1.1, then analyze the shortcomings of the NeRF with sparse input views in Sec. 4.3.1.2 and finally discuss the design of augmentations in Sec. 4.3.1.3. Fig. 4.3 shows the detailed architecture of Simple-NeRF.

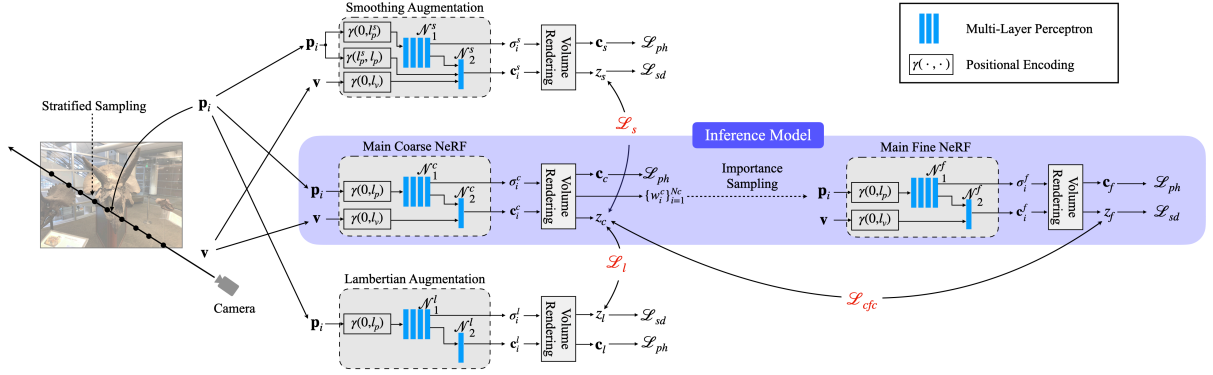

Figure 4.3: Architecture of Simple-NeRF. We train two augmented NeRF models in tandem with the main NeRF. In smoothing augmentation, we reduce the positional encoding frequencies that are input to $\mathcal{N}_1^s$ and concatenate the remaining frequencies to the input of $\mathcal{N}_2^s$. For Lambertian augmentation, we ask $\mathcal{N}_2^l$ to output the color based on position alone, independent of the viewing direction. We add depth supervision losses $\mathcal{L}_s$ and $\mathcal{L}_l$ between the coarse NeRFs of the main and augmented models and a consistency loss $\mathcal{L}_{cfc}$ between the coarse and fine NeRFs of the main model. During inference, only the Main Model is employed.

### 4.3.1.1   NeRF Preliminaries

The NeRF learns the radiance field $\mathcal{F}$ using two neural networks $\mathcal{N}_1, \mathcal{N}_2$ and predicts the view-independent volume density $\sigma_i$ and view-dependent color $\mathbf{c}_i$ as

$$\sigma_i, \mathbf{h}_i = \mathcal{N}_1\left(\gamma(\mathbf{p}_i, 0, l_p)\right); \qquad \mathbf{c}_i = \mathcal{N}_2\left(\mathbf{h}_i, \gamma(\mathbf{v}, 0, l_v)\right), \qquad (4.3)$$

where $\mathbf{v}$ is the viewing direction, $\mathbf{h}_i$ is a latent feature of $\mathbf{p}_i$ and

$$\gamma(x, d_1, d_2) = [\sin(2^{d_1}x), \cos(2^{d_1}x), \dots, \sin(2^{d_2-1}x), \cos(2^{d_2-1}x)] \qquad (4.4)$$

is the positional encoding from frequency $d_1$ to $d_2$. $l_p$ and $l_v$ are the highest positional encoding frequencies for $\mathbf{p}_i$ and $\mathbf{v}$ respectively. When $d_1 = 0$, $x$ is concatenated to the positional encoding features in Eq. (4.4). NeRF circumvents the need for the dense sampling of 3D points along a ray by employing two sets of MLPs, a coarse NeRF and a fine NeRF, both trained using $\mathcal{L}_{ph}$. The coarse NeRF is trained with a coarse stratified sampling, and the fine NeRF with dense sampling around object surfaces, where object surfaces are coarsely localized based on the predictions of the coarse NeRF. Since the scene geometry is mainly learned by the coarse NeRF, we add the augmentations only to the coarse NeRF.

### 4.3.1.2 Analysing Sparse Input NeRF

With sparse input views, we find that two components of the NeRF, namely positional-encoding and view-dependent radiance, can cause overfitting, leading to distortions in novel views. Both positional encoding and view-dependent radiance are elements designed to increase the capability of the NeRF to explain different complex phenomena. For example, the former helps in learning thin objects against a farther background, and the latter helps in learning specular objects. However, when training with sparse views, the fewer constraints coupled with the higher capacity of the NeRF lead to solutions that overfit the observed images and learn implausible scene geometries. Specifically, the high positional encoding degree leads to undesired depth discontinuities in smooth-depth regions resulting in floater artifacts, where a part of an object is broken away from it and floats freely in space [17], as shown in Fig. 4.4a. The view-dependent radiance causes shape-radiance ambiguity, leading to duplication artifacts in the novel views as shown in Fig. 4.4b. With sparse input views, the NeRF explains different objects by varying the color of the same 3D points based on the viewing direction, thereby giving us an illusion of the object without learning the correct geometry of the object. This is, in a way, similar to the illusion created by lenticular images and can be observed better in the supplementary video on our project page. Our augmentations consist of reducing the capability of the NeRF model with respect to the positional encoding and

RGB frame  DS-NeRF  Simple-NeRF

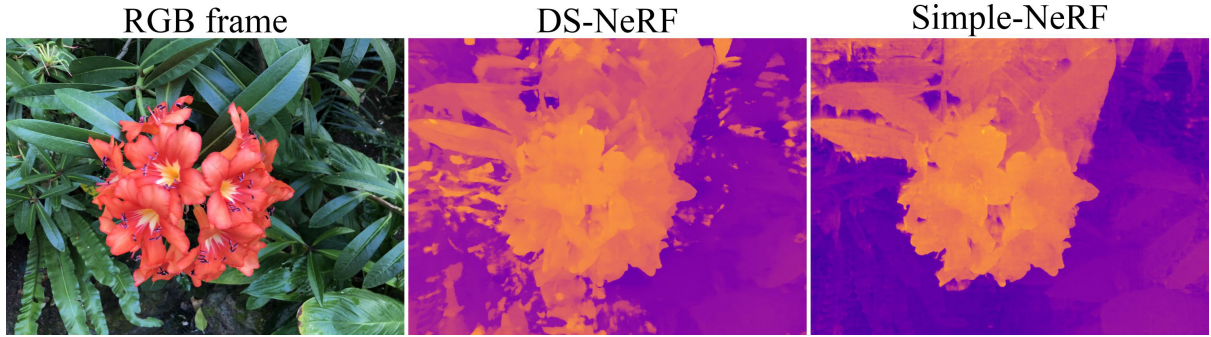

(a) *Floater artifacts:* We visualize the depth learned by the NeRF model for an input frame from the NeRF-LLFF flower scene.

Simple-NeRF  DS-NeRF  Simple-NeRF

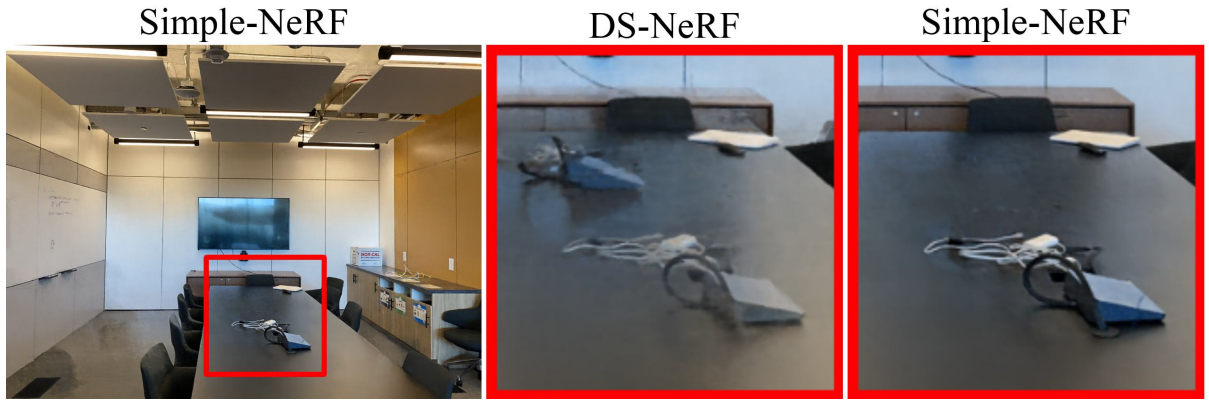

(b) *Duplication artifacts:* To visualize the duplication artifacts that arise due to the shape-radiance ambiguity in sparse-input NeRF, we render an input frame by only changing the viewing direction. This is an example from the NeRF-LLFF room scene.

Figure 4.4: **Failure of sparse-input NeRF:** We show two shortcomings of the NeRF when trained with two input views on the NeRF-LLFF dataset. In Fig (a), we observe the floaters as small orange regions in the depth map. In Fig (b), we observe the duplication of the object on the table caused by the NeRF trying to blend the input images. Simple-NeRF introduces regularizations to mitigate these distortions as seen in both figures. We note that the models used to synthesize the above images include the sparse depth supervision (Sec. 4.3.4).

view-dependent radiance to obtain better depth supervision.

### 4.3.1.3 Design of Augmentations

We employ two augmentations, one each for regularizing positional encoding and view-dependent radiance, which we describe in the following. We refer to the two augmentations as smoothing and Lambertian augmentations, respectively.

*Smoothing augmentation:* The positional encoding maps two nearby points in $\mathbb{R}^3$ to two farther away points in $\mathbb{R}^{3(2l_p+1)}$ allowing the NeRF to learn sharp discontinuities in volume density between the two points in $\mathbb{R}^3$ as a smooth function in $\mathbb{R}^{3(2l_p+1)}$. We reduce the depth discontinuities, which are caused by discontinuities in the volume density, by reducing the highest positional encoding frequency for $\mathbf{p}_i$ to $l_p^s < l_p$ as

$$\sigma_i, \mathbf{h}_i = \mathcal{N}_1^s(\gamma(\mathbf{p}_i, 0, l_p^s)), \qquad (4.5)$$

where $\mathcal{N}_1^s$ is the MLP of the augmented model. The main model is more accurate where depth discontinuities are required, and the augmented model is more accurate where discontinuities are not required. We determine the respective locations as binary masks and use only the reliable depth estimates from one model to supervise the other model, as we explain in Sec. 4.3.1.4.

Since color tends to have more discontinuities than depth in regions such as textures, we include the remaining high-frequency positional encoding components of $\mathbf{p}_i$ in the input for $\mathcal{N}_2$ as

$$\mathbf{c}_i = \mathcal{N}_2^s(\mathbf{h}_i, \gamma(\mathbf{p}_i, l_p^s, l_p), \gamma(\mathbf{v}_i, 0, l_v)). \qquad (4.6)$$

Note that $\mathbf{h}_i$ already includes the low-frequency positional encoding components of $\mathbf{p}_i$.

*Lambertian Augmentation:* The ability of the NeRF to predict view-dependent radiance helps it learn non-Lambertian surfaces. With fewer images, the NeRF can simply learn any random geometry and change the color of 3D points in accordance with the input viewpoint to explain away the observed images [214]. To guard the NeRF against this, we disable the view-dependent radiance in the second augmented NeRF model to output color based on $\mathbf{p}_i$ alone as

$$\sigma_i, \mathbf{h}_i = \mathcal{N}_1^l(\gamma(\mathbf{p}_i, 0, l_p)); \qquad \mathbf{c}_i = \mathcal{N}_2^l(\mathbf{h}_i), \tag{4.7}$$

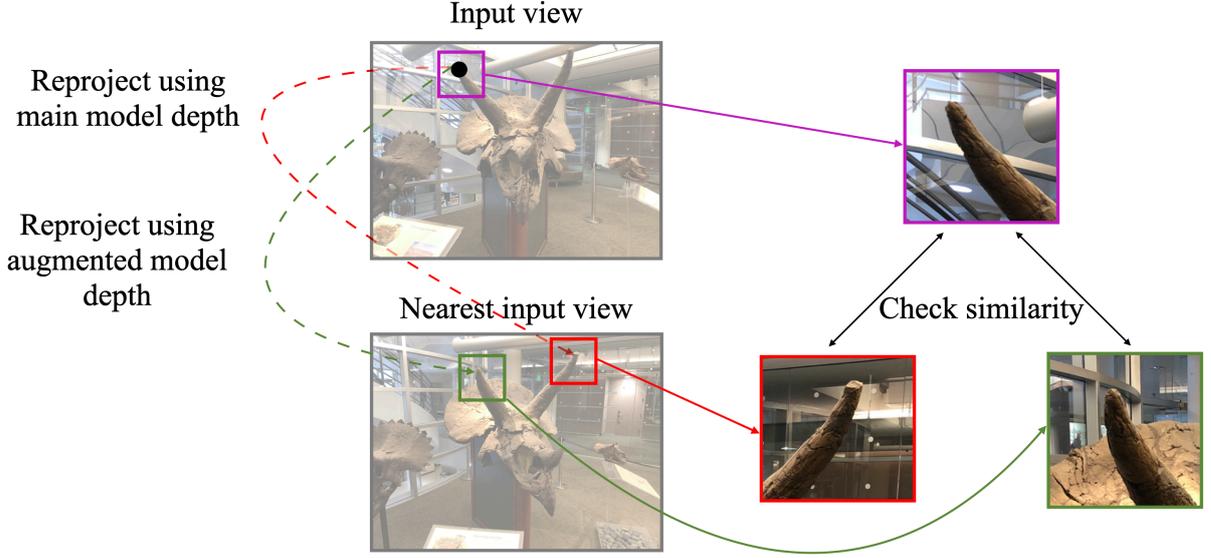

Figure 4.5: **Determining the reliability of depths for supervision:** We choose the depth that has higher similarity, with respect to the patches reprojected to the nearest input view, to supervise the other model (Sec. 4.3.1.4). The patches are only representative and are not to scale.

We note that while the augmented model is more accurate in Lambertian regions, the main model is better equipped to handle specular objects. We determine the respective locations as we explain in the following and use only the reliable depth estimates for supervision.

### 4.3.1.4   Determining Reliable Depth Estimates

Let the depths estimated by the main and augmented models for pixel $\mathbf{q}$ be $z_m$ and $z_a$ respectively. We now seek to determine the more accurate depth among the two. Fig. 4.5 shows our approach to determining the reliability of the estimated depth. Specifically, we reproject a $k \times k$ patch around $\mathbf{q}$ to the nearest training view using both $z_m$ and $z_a$. We then compute the similarity of the reprojected patch with the corresponding patch in the first image using the mean squared error (MSE) in intensities. We choose the depth corresponding to lower MSE as the reliable depth. To filter out the cases where both the main and augmented models predict incorrect depth, we define a threshold $e_\tau$ and mark the depth to be reliable only if its MSE is also less than $e_\tau$. If $e_m$ and $e_a$ are the reprojection MSE corresponding to $z_m$ and $z_a$ respectively, we compute a mask $m_a$ that indicates where the augmented model is more reliable as

$$m_a = \begin{cases} 1 & \text{if } (e_a \leq e_m) \text{ and } (e_a \leq e_\tau) \\ 0 & \text{otherwise .} \end{cases} \qquad (4.8)$$

We compute the reliability mask $m_m$ for the main model similarly. We now impose the depth supervision as

$$\mathcal{L}_{aug} = \mathbb{1}_{\{m_a=1\}} \odot \|z_m - \overline{\nabla}(z_a)\|^2 + \mathbb{1}_{\{m_m=1\}} \odot \|\overline{\nabla}(z_m) - z_a\|^2, \qquad (4.9)$$

where $\odot$ denotes element-wise product, $\mathbb{1}$ is the indicator function and $\overline{\nabla}$ is the stop-gradient operator. We impose two losses, $\mathcal{L}_s$ for the smoothing augmentation and $\mathcal{L}_l$ for the Lambertian augmentation. The final depth supervision loss is the sum of the two losses.

For specular regions, the intensities of the reprojected patches may not match, leading to the masks being zero. This only implies supervision for fewer pixels and not supervision with incorrect depth estimates.

### 4.3.1.5 Hierarchical Sampling

Since multiple solutions can explain the observed images in the few-shot setting, the coarse and fine MLP of the NeRF may converge to different depth estimates for a given pixel as shown in Fig. 4.6b. Thus, dense sampling may not be employed around the

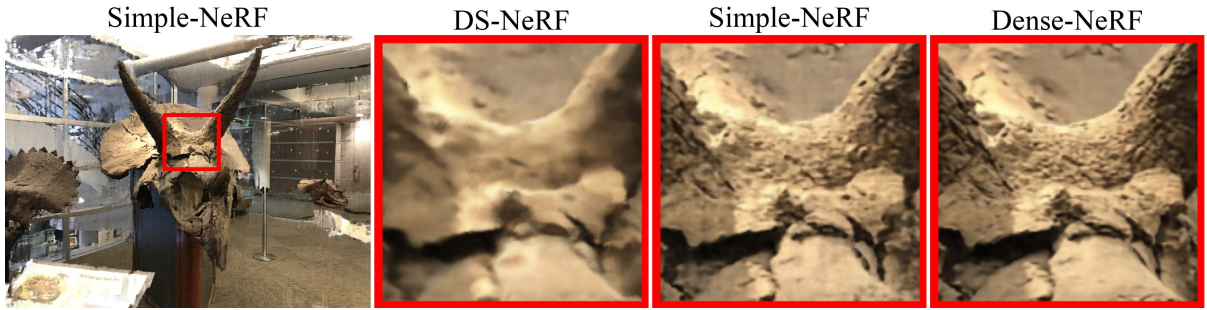

(a) The above images correspond to the NeRF-LLFF horns scene. We enlarge a small region of the frame to better observe the improvement in sharpness.

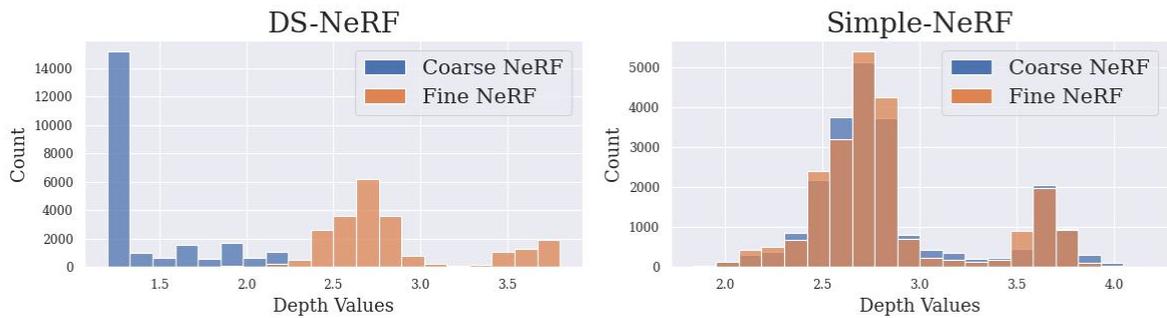

(b) Histogram of depth values predicted by the coarse and fine NeRF models for the image patch shown in Fig (a).

Figure 4.6: **Ineffective hierarchical sampling in sparse-input NeRF:** Fig (b) shows that the coarse and fine models in the NeRF converge to different depth estimates when training with sparse input views. This leads to ineffective hierarchical sampling, resulting in blurry predictions in Fig (a). By predicting consistent depth estimates with the help of $\mathcal{L}_{cfc}$, Simple-NeRF predicts consistent depth estimates leading to sharp reconstructions. We note that the models used to synthesize the above images include the sparse depth supervision (Sec. 4.3.4).

region where the fine NeRF predicts the object surface, which is equivalent to using only the coarse sampling for the fine NeRF. This can lead to blur in rendered images as seen in Fig. 4.6a. To prevent such inconsistencies, we drive the two NeRFs to be consistent in their solutions by imposing an MSE loss between the depths predicted by the two NeRFs. If $z_c$ and $z_f$ are the depths estimated by the coarse and fine NeRFs respectively, we define the coarse-fine consistency loss as

$$\mathcal{L}_{cfc} = \mathbb{1}_{\{m_f=1\}} \odot \|z_c - \overline{\nabla}(z_f)\|^2 + \mathbb{1}_{\{m_c=1\}} \odot \|\overline{\nabla}(z_c) - z_f\|^2, \qquad (4.10)$$

where the masks $m_c$ and $m_f$ are determined as we describe in Sec. 4.3.1.4.

Apart from enforcing consistency between the coarse and fine NeRF models, $\mathcal{L}_{cfc}$ provides two additional benefits. Without $\mathcal{L}_{cfc}$, the augmentations need to be imposed on the fine NeRF as well, leading to an increase in the training time and memory requirements. Secondly, if one of the coarse or fine NeRFs converges to the correct solution, $\mathcal{L}_{cfc}$ helps quickly convey the knowledge to the other NeRF, thereby facilitating faster convergence.

### 4.3.2 Simple-TensoRF

We first provide a brief overview of TensoRF [33] in Sec. 4.3.2.1 and also describe the notation required to explain the design of our augmentations. We discuss the distortions observed with sparse input TensoRF in Sec. 4.3.2.2 and then discuss the design of augmentations in Sec. 4.3.2.3.

#### 4.3.2.1 TensoRF Preliminaries

TensoRF models the fields $\mathcal{F}_1$ and $\mathcal{F}_2$ with a tensor $\mathcal{G}_1$ and a tiny MLP $\mathcal{N}_2$, respectively. The 3D tensor $\mathcal{G}_1$ is factorized as the sum of outer products of 1D vectors $\mathbf{v}$ and 2D matrices $\mathbf{M}$. Specifically, $\mathcal{G}_1$ consists of two 3D tensors, $\mathcal{G}_\sigma$ to learn the volume density and $\mathcal{G}_c$ to learn the latent features of the color as

$$\mathcal{G}_\sigma = \sum_{r=1}^{R_\sigma} \mathbf{v}_{\sigma,r}^X \circ \mathbf{M}_{\sigma,r}^{YZ} + \sum_{r=1}^{R_\sigma} \mathbf{v}_{\sigma,r}^Y \circ \mathbf{M}_{\sigma,r}^{XZ} + \sum_{r=1}^{R_\sigma} \mathbf{v}_{\sigma,r}^Z \circ \mathbf{M}_{\sigma,r}^{XY}, \qquad (4.11)$$

$$\mathcal{G}_c = \sum_{r=1}^{R_c} \mathbf{v}_{c,r}^X \circ \mathbf{M}_{c,r}^{YZ} \circ \mathbf{a}_{3r-2} + \sum_{r=1}^{R_c} \mathbf{v}_{c,r}^Y \circ \mathbf{M}_{c,r}^{XZ} \circ \mathbf{a}_{3r-1} \qquad (4.12)$$

$$+ \sum_{r=1}^{R_c} \mathbf{v}_{c,r}^Z \circ \mathbf{M}_{c,r}^{XY} \circ \mathbf{a}_{3r},$$

$$\sigma_i = \text{sigmoid}(\mathcal{G}_\sigma(\mathbf{p}_i)); \qquad\qquad \mathbf{h}_i = \mathcal{G}_c(\mathbf{p}_i), \qquad (4.13)$$

where $\circ$ represents the outer product, and $R_\sigma$ and $R_c$ represent the number of components in the factorization of sigma and color grids, respectively. $\mathcal{G}(\mathbf{p}_i)$ is obtained by trilinearly interpolating $\mathcal{G}$ at $\mathbf{p}_i$. $\mathbf{v}^X \in \mathbb{R}^I$ and $\mathbf{M}^{YZ} \in \mathbb{R}^{J \times K}$ represent the vector along the x-axis and the matrix in the yz-plane respectively and so on, where $I, J$ and $K$ represent the resolution of the tensor in the $x, y$ and $z$ dimensions respectively. Thus, the total number of voxels in the tensor is $N_{vox} = I \times J \times K$. Note that $\mathcal{G}_c$ uses an additional vector $\mathbf{a}_r \in \mathbb{R}^D$ to learn appearance as a latent feature of dimension $D$.

TensoRF assumes that the entire scene is contained within a 3D bounding box $\mathbf{b}$ as shown in Fig. 4.7b, whose vertices are given by $\{(b_{x_1}, b_{x_2}), (b_{y_1}, b_{y_2}), (b_{z_1}, b_{z_2})\}$. TensoRF handles unbounded forward-facing scenes by transforming the space into normalized device coordinates (ndc) similar to the NeRF. The coarse to fine training is implemented by using lower resolution tensors $\mathcal{G}_\sigma$ and $\mathcal{G}_c$ during the initial stages of the optimization and gradually increasing the resolution as the training progresses. Finally, the color at $\mathbf{p}_i$ is obtained using the tiny MLP $\mathcal{N}_2$ as

$$\mathbf{c}_i = \mathcal{N}_2(\mathbf{h}_i, \gamma(\mathbf{v}, 0, l_v)), \qquad (4.14)$$

where $\gamma$ is the positional encoding described by Eq. (4.4). Thus, we note that $\mathcal{F}_2 = \gamma \circ \mathcal{N}_2$. The color of the pixel is then obtained through volume rendering using $\sigma_i$ and $\mathbf{c}_i$ as in Sec. 4.2. For further details, we refer the readers to TensoRF [33].

RGB frame   DS-TensoRF   Simple-TensoRF

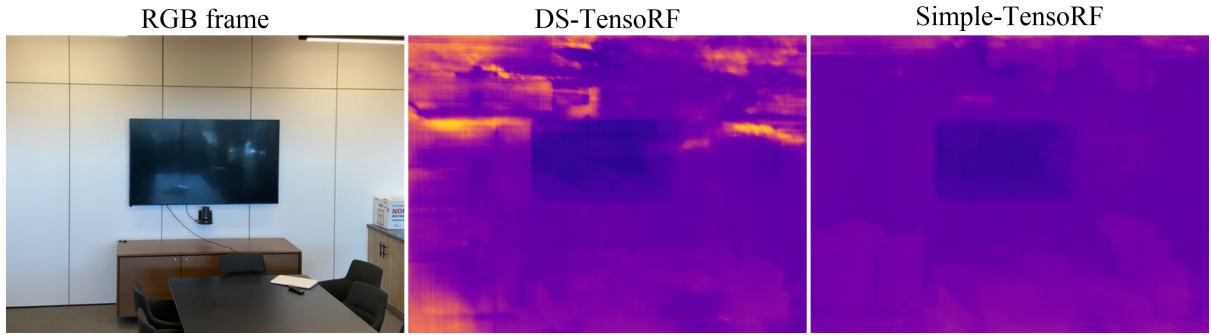

(a) *Floater artifacts:* We visualize the depth learned by the TensoRF model for an input frame from the NeRF-LLFF room scene.

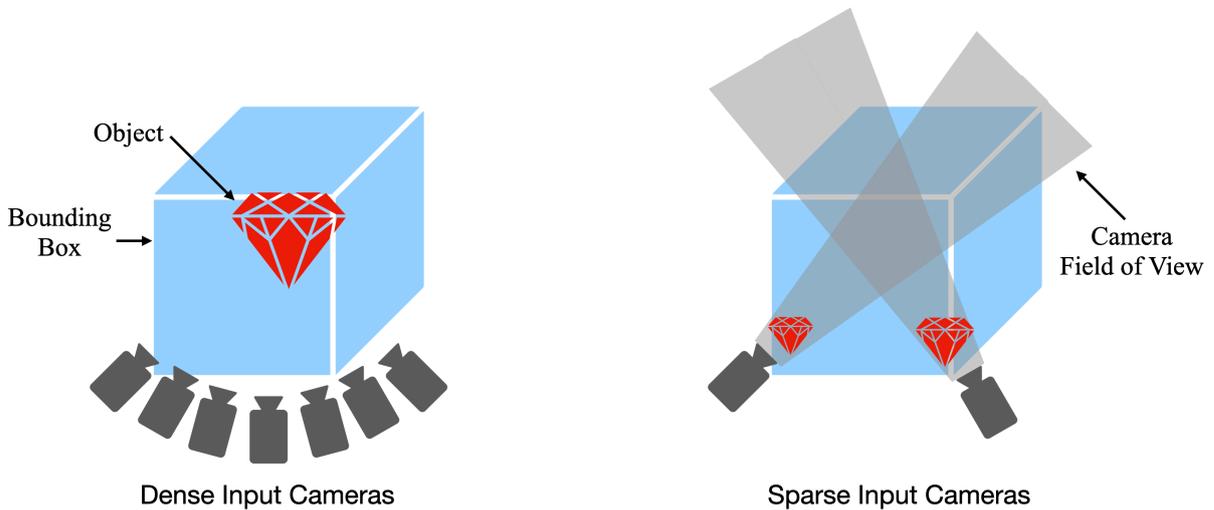

(b) *Objects close to camera:* We illustrate TensoRF incorrectly placing objects close to the cameras using a toy example.

Figure 4.7: **Failure of sparse-input TensoRF**: We show the two shortcomings of TensoRF when trained with few input views. In Fig (a), the orange regions indicate the floaters. For reference, we also show the depth learned by Simple-TensoRF, which is free from floaters. We note that the models used to synthesize these images include the sparse depth supervision (Sec. 4.3.4). In Fig (b), the image on the left depicts the true scene, which can be accurately learned by the TensoRF model provided with dense input views. The image on the right illustrates how TensoRF can incorrectly place the objects yet perfectly reconstruct the input views, when training with few input views.

#### 4.3.2.2  Analysing Sparse Input TensoRF

When training a TensoRF model with sparse input views, we find that three of its components cause overfitting, leading to distortions in novel views. Employing a higher resolution tensor $\mathcal{G}_\sigma$ with a large number of components $R_\sigma$ allows the TensoRF to learn sharp depth edges, but results in undesired depth discontinuities in smooth regions causing floaters as shown in Fig. 4.7a. Further, the large bounding box $\mathbf{b}$ allows the TensoRF to handle objects that are truly very close to the camera. On account of large distances between cameras when only a few input views are available, it may be possible to place objects close to one camera such that they are out of the field of view of the other cameras, even for objects visible in multiple input views. Specifically, TensoRF learns multiple copies of the same object, each visible in only one input view, thereby explaining the observations without learning the geometry of the objects as shown in Fig. 4.7b. We design the augmentations to reduce the capability of the TensoRF model with respect to these three components.

#### 4.3.2.3  Design of Augmentations

Employing a high-resolution and high-rank tensor $\mathcal{G}_\sigma$ enables TensoRF to learn significantly different $\sigma$ values for two nearby points in $\mathbb{R}^3$ leading to undesired depth discontinuities in smooth regions. We constrain the augmented TensoRF to learn only smooth and continuous depth surfaces by reducing the number of components to $R_\sigma^s < R_\sigma$ and also reducing the number of voxels of $\mathcal{G}_\sigma$ from to $N_{vox}^s < N_{vox}$. We note that modifying only one of these components is insufficient to achieve the desired smoothing. For example, only reducing the resolution of the grid allows TensoRF to learn sharp changes in $\sigma$ at the voxel edges, leading to block artifacts. On the other hand, only reducing the number of components allows TensoRF to learn sharp changes in $\sigma$ on account of the high-resolution grid.

Further, we find that reducing $R_\sigma$ to $R_\sigma^s$ leads to the augmented TensoRF learning cloudy volumes instead of hard object surfaces. We encourage the augmented TensoRF to learn hard surfaces by employing a mass concentration loss that minimizes the entropy of mass, grouped into $N_{mc}$ intervals as

$$\mathcal{L}_{mc} = H\left(\left\{\sum_{i=(j-1)(N/N_{mc})+1}^{j(N/N_{mc})} w_i\right\}_{j=1}^{N_{mc}}\right), \tag{4.15}$$

where $H(w_1, w_2, \ldots, w_n) = \sum_{i=1}^{n}(-w_i \log w_i)$ is the entropy operator, $N$ is the number of 3D points $\mathbf{p}_i$ along a ray and $w_i$ is the weight corresponding to $\mathbf{p}_i$ as described in Eq. (4.2).

Objects that are incorrectly placed close to the camera due to a large bounding box are typically smooth in depth and hence are not mitigated by the above augmentation. We design a second augmentation to mitigate such distortions by reducing the size of the bounding box $\mathbf{b}$ along the z-axis by increasing $b_{z_1}$ to $b_{z_1}^s$. We note that replicating the same in the main TensoRF model could lead to distortions in objects that are truly close to the camera. In practice, we find that including both the augmentations in a single augmented TensoRF model works reasonably well, and hence we employ a single augmented model. We then use the reliable depth estimates from the augmented model to supervise the main model as in Eq. (4.9).

### 4.3.3 Simple-ZipNeRF

ZipNeRF [18] integrates the iNGP model [125], which achieves significant improvements in optimization and rendering times, with the anti-aliasing ability of MipNeRF [16] and the ability to handle unbounded 360° scenes of MipNeRF360 [17]. Our contributions to enable the training of ZipNeRF with sparse input views are mainly with respect to the components of the iNGP model, and hence, we believe that the augmentations designed for ZipNeRF are relevant to iNGP as well. We discuss the specific components of ZipNeRF that are relevant in our augmentations in Sec. 4.3.3.1, analyze the limitations of ZipNeRF with sparse input views in Sec. 4.3.3.2 and then discuss the design of our augmentations in Sec. 4.3.3.3.

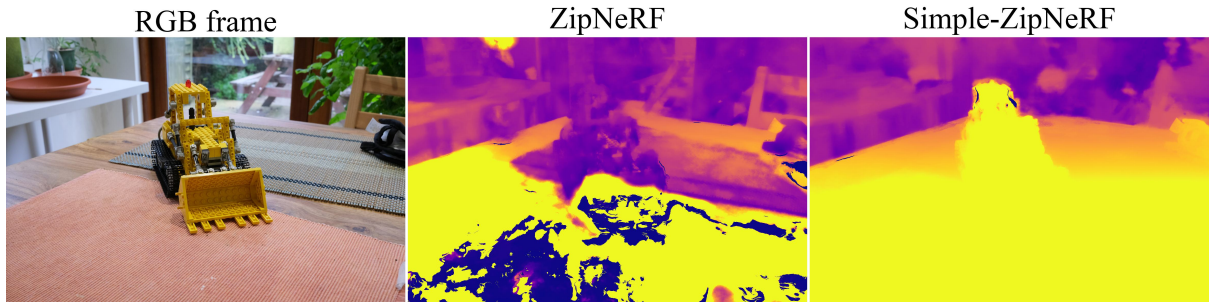

RGB frame    ZipNeRF    Simple-ZipNeRF

(a) *Floater artifacts:* We visualize the depth learned by the ZipNeRF model for an input frame from the MipNeRF360 kitchen scene.

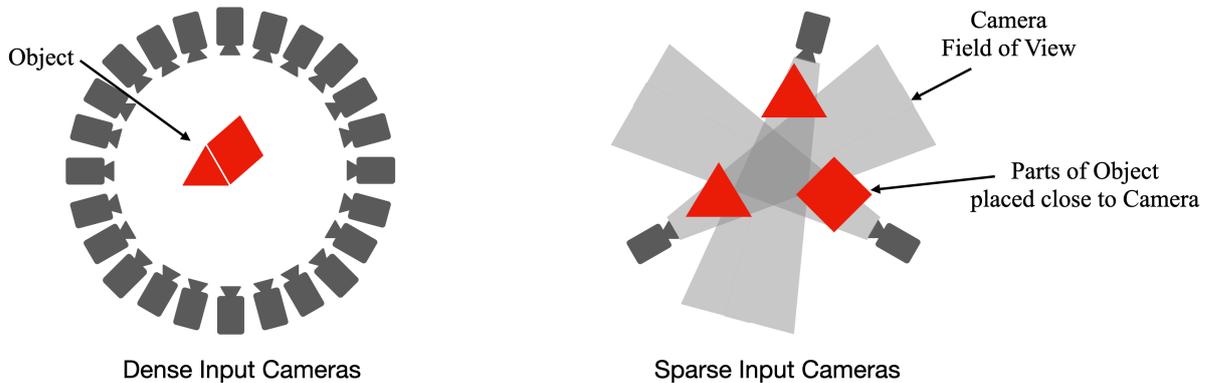

(b) *Objects close to camera:* We illustrate ZipNeRF incorrectly placing objects close to the cameras using a toy example.

Figure 4.8: We show two shortcomings of ZipNeRF when trained with few input views. In Fig (a), while the RGB frame for an input view is reconstructed perfectly, we observe floaters in the depth image, shown by the dark-blue regions. For reference, we also show the depth learned by Simple-ZipNeRF, which is free from floaters and better reconstructs the scene. In Fig (b), the image on the left depicts the true scene, which can be accurately learned by the ZipNeRF model provided with dense input views. The image on the right illustrates how the sparse-input ZipNeRF model can incorrectly place parts of the object close to the cameras, yet perfectly reconstruct the input views.

### 4.3.3.1 ZipNeRF Preliminaries

ZipNeRF employs a multi-resolution grid and a hash function that maps every vertex of the grid to an entry in a hash table. The hash table contains the latent features representing the volume density and the radiance. Concretely, given a point $\mathbf{p}_i \in \mathbb{R}^3$, the vertices of the voxel enclosing $\mathbf{p}_i$ are mapped to an entry in a hash table of length $T$ through the use of hash function $\mathcal{H}_1$ as,

$$\mathcal{H}_1(\mathbf{p}) = \left( \bigoplus_{j=1}^{3} p_j \pi_j \right) \mod T, \tag{4.16}$$

where $\oplus$ denotes the bit-wise XOR operation, $\pi_j$ is a prime number, and $p_j$ is the $j$-th coordinate of $\mathbf{p}$. The feature vectors corresponding to the eight vertices of the voxel are trilinearly interpolated. The same procedure is repeated for every level of the multi-resolution grid, and the corresponding interpolated features are concatenated to form the latent feature $\mathcal{H}_1(\mathbf{p}_i)$. Two tiny MLPs are employed to decode $\mathcal{H}_1(\mathbf{p}_i)$ into the volume density and the radiance as

$$\sigma_i, \mathbf{h}_i = \mathcal{N}_1(\mathcal{H}_1(\mathbf{p}_i)); \qquad \mathbf{c}_i = \mathcal{N}_2(\mathbf{h}_i, \gamma(\mathbf{v}, 0, l_v)), \tag{4.17}$$

where $\gamma$ is the positional encoding as defined in Eq. (4.4). We note that $\mathcal{F}_1$ and $\mathcal{F}_2$ in Eq. (4.1) are thus represented as $\mathcal{F}_1 = \mathcal{H}_1 \circ \mathcal{N}_1$ and $\mathcal{F}_2 = \gamma \circ \mathcal{N}_2$. The color of the pixel is then obtained through volume rendering using $\sigma_i$ and $\mathbf{c}_i$ as in Sec. 4.2. Note that multiple vertices in the grid could map to the same entry in the hash table at every level. iNGP and ZipNeRF rely on the MLP $\mathcal{N}_1$ to resolve such collisions based on multi-resolution features. Unbounded scenes are handled by employing a contraction function that maps the distance along the ray from $z \in [z_{\text{near}}, z_{\text{far}}]$ to a normalized distance $s \in [0, 1]$. The 3D points $\mathbf{p}_i$ are then sampled in $s$-domain. For further details, we refer the readers to iNGP [125] and ZipNeRF [18].

### 4.3.3.2   Analysing Sparse Input ZipNeRF

We find that two components of ZipNeRF tend to cause overfitting when trained with sparse input views, leading to distortions in novel views. Firstly, employing a hash table with a large size $T$ enables ZipNeRF to learn sharp depth edges, but introduces undesired depth discontinuities in smooth regions, causing floaters as shown in Fig. 4.8a. Secondly, since ZipNeRF handles unbounded 360° scenes, it learns the radiance field over the entire 3D space. Similar to TensoRF, ZipNeRF tends to incorrectly place multiple copies of objects very close to the camera without learning the correct geometry as shown in Fig. 4.8b. Thus, we design the augmentations to reduce the capability of the ZipNeRF model with respect to these two components.

### 4.3.3.3   Design of Augmentations

Employing a hash table of larger size $T$ allows ZipNeRF to avoid collisions and not share features across multiple 3D points. This enables ZipNeRF to map two nearby points in $\mathbb{R}^3$ to two independent entries in the hash table, thus mapping them to two farther away points in the latent feature space. This allows the MLP $\mathcal{N}_1$ to learn discontinuities in the volume density, resulting in sharp depth edges. We encourage the augmented model to share features across more 3D points by reducing the size of the hash table to $T^s < T$.

To mitigate the objects being placed close to the camera incorrectly, we cannot reduce the size of the bounding box as in TensoRF, since ZipNeRF handles unbounded scenes. We achieve a similar effect by sampling the 3D points $\mathbf{p}_i$ along the ray in $s$-domain in the range $s \in [s_{\text{near}}, 1]$ instead of $s \in [0, 1]$. This ensures that the objects are placed at least at a certain distance away from the camera in the augmented model. However, we note that employing the above modification in the main model is detrimental to learning or rendering any objects that are truly close to the camera. In practice, we find that including both the augmentations in a single augmented ZipNeRF model works reasonably well, and hence, we employ a single augmented model. We then use these depth estimates as in Eq. (4.9).

### 4.3.4 Overall Loss

Let $\mathcal{L}_m$ denote the combination of the losses employed by the corresponding radiance fields. For example, while the NeRF employs only the photometric loss $\mathcal{L}_{ph}$, TensoRF employs a total variation regularization in addition to $\mathcal{L}_{ph}$. We refer the readers to the corresponding papers for the details of all the losses imposed. We impose all such losses on the augmented models as well and denote them by $\mathcal{L}_a$. In addition, we also include the sparse depth loss on both the main and augmented models as,

$$\mathcal{L}_{sd} = \|z_m - \hat{z}\|^2 + \|z_a - \hat{z}\|^2, \tag{4.18}$$

where $z_m$ and $z_a$ are the depths obtained from the main and augmented models respectively, and $\hat{z}$ is the sparse depth given by the SfM model [46]. Our final loss is a combination of all the losses as,

$$\mathcal{L} = \lambda_m \mathcal{L}_m + \lambda_a \mathcal{L}_a + \lambda_{sd} \mathcal{L}_{sd} + \lambda_{aug} \mathcal{L}_{aug} + \tag{4.19}$$
$$\lambda_{cfc} \mathcal{L}_{cfc} + \lambda_{mc} \mathcal{L}_{mc},$$

where $\mathcal{L}_{cfc}$ and $\mathcal{L}_{mc}$ are respectively imposed for the main NeRF and augmented TensoRF models only, and $\lambda_m, \lambda_a, \lambda_{sd}, \lambda_{aug}, \lambda_{cfc}$ and $\lambda_{mc}$ are hyper-parameters.

## 4.4 Experimental Setup

### 4.4.1 Datasets

We evaluate the performance of our models on four popular datasets, namely NeRF-LLFF [122], RealEstate-10K [216], MipNeRF360 [17] and NeRF-Synthetic [123]. We assume the camera parameters are known for the input images, since in applications such as robotics or extended reality, external sensors or a pre-calibrated set of cameras may provide the camera poses.

Table 4.1: Train and test frame numbers of RealEstate-10K dataset used in the three different settings.

| No. of i/p frames | Train frame nos. | Test frame nos. |
|:---:|:---:|:---:|
| 2 | 10, 20 | 5–9, 11–19, 21–25 |
| 3 | 10, 20, 30 | 5–9, 11–19, 21–29, 31–35 |
| 4 | 0, 10, 20, 30 | 1–9, 11–19, 21–29, 31–35 |

*NeRF-LLFF* dataset contains eight real-world forward-facing scenes typically consisting of an object at the centre against a complex background. Each scene contains a varying number of images ranging from 20 to 60, each with a spatial resolution of $1008 \times 756$. Following prior work [128], we use every 8th view as the test view and uniformly sample 2, 3 or 4 input views from the remaining.

*RealEstate-10K* dataset contains a large number of real-world forward-facing scenes, from which we select 5 test scenes for our experiments. We include both indoor and unbounded outdoor scenes and select 50 temporally continuous frames from each scene. The frames have a spatial resolution of $1024 \times 576$. Following prior work [159], we reserve every 10th frame for training and choose 2, 3 or 4 input views among them. In the remaining 45 frames, we use those frames that are not very far from the input frames for testing. Specifically, we choose all the frames between the training views that correspond to interpolation and five frames on either side that correspond to extrapolation. Tab. 4.1 shows the train and test frame numbers we use for the three different settings.

*MipNeRF360* dataset contains seven publicly available unbounded 360° real-world scenes including both indoor and outdoor scenes. Each scene contains 100 to 300 images. The four indoor scenes have a spatial resolution of approximately $1560 \times 1040$, and the three outdoor scenes have an approximate spatial resolution of $1250 \times 830$. Following prior work [18], we reserve every 8th view for testing and uniformly sample 12, 20 and 36 input views from the remaining. We use more input views on this dataset as compared to the other datasets owing to the significantly larger fields of view.

*NeRF-Synthetic* dataset contains eight bounded 360° synthetic scenes, each containing 100 train images and 200 test images. All the scenes have a spatial resolution of $800 \times 800$. For training, we uniformly sample 4, 8 and 12 input views from the training set and test on all 200 test images.

### 4.4.2 Evaluation measures

We quantitatively evaluate the predicted frames from various models using the peak signal-to-noise ratio (PSNR), structural similarity (SSIM) [190] and LPIPS [215] measures. For LPIPS, we use the v0.1 release with the AlexNet [86] backbone, as suggested by the authors. We also employ depth mean absolute error (MAE) to evaluate the models on their ability to predict absolute depth in novel views. In addition, we also evaluate the models with regard to their ability to predict better relative depth using the spearman rank order correlation coefficient (SROCC). Obtaining better relative depth might be more crucial in downstream applications such as 3D scene editing. Since the ground truth depth is not provided in the datasets, we train NeRF and ZipNeRF models with dense input views on forward-facing and 360° datasets respectively and use their depth predictions as pseudo ground truth. On the NeRF-LLFF and MipNeRF360 datasets, we normalize the predicted depths by the median ground truth depth, since the scenes have different depth ranges. With very few input views on forward-facing datasets, the test views could contain regions that are not visible in the input views, and hence, we also evaluate both the view synthesis and depth performance in visible regions only. To determine such regions, we use the depth estimated by a NeRF trained with dense input views and compute the visible region mask through reprojection error in depth. On the other hand, the input views cover most of the scene in the 360° datasets, and hence we evaluate the performance on full frames. We do not evaluate the rendered depth on the NeRF-Synthetic dataset since the depth estimated with the dense input ZipNeRF is unreliable, especially in the white background regions, and the ground truth depth is not provided in the dataset either.

Table 4.2: Quantitative results of NeRF based models on the NeRF-LLFF dataset.

| Model | 2 views | | | 3 views | | | 4 views | | |
|---|---|---|---|---|---|---|---|---|---|
| | LPIPS ↓ | SSIM ↑ | PSNR ↑ | LPIPS ↓ | SSIM ↑ | PSNR ↑ | LPIPS ↓ | SSIM ↑ | PSNR ↑ |
| InfoNeRF | 0.6024 | 0.2219 | 9.16 | 0.6732 | 0.1953 | 8.37 | 0.6985 | 0.2270 | 9.18 |
| DietNeRF | 0.5465 | 0.3283 | 11.94 | 0.6120 | 0.3405 | 11.76 | 0.6506 | 0.3496 | 11.86 |
| RegNeRF | 0.3056 | 0.5712 | 18.52 | 0.2908 | 0.6334 | 20.22 | 0.2794 | 0.6645 | 21.32 |
| FreeNeRF | **0.2638** | 0.6322 | 19.52 | 0.2754 | 0.6583 | 20.93 | 0.2848 | 0.6764 | 21.91 |
| DS-NeRF | 0.3106 | 0.5862 | 18.24 | 0.3031 | 0.6321 | 20.20 | 0.2979 | 0.6582 | 21.23 |
| DDP-NeRF | 0.2851 | 0.6218 | 18.73 | 0.3250 | 0.6152 | 18.73 | 0.3042 | 0.6558 | 20.17 |
| ViP-NeRF | 0.2768 | 0.6225 | 18.61 | 0.2798 | 0.6548 | 20.54 | 0.2854 | 0.6675 | 20.75 |
| Simple-NeRF | 0.2688 | **0.6501** | **19.57** | **0.2559** | **0.6940** | **21.37** | **0.2633** | **0.7016** | **21.99** |

Table 4.3: Quantitative results of NeRF based models on the RealEstate-10K dataset.

| Model | 2 views | | | 3 views | | | 4 views | | |
|---|---|---|---|---|---|---|---|---|---|
| | LPIPS ↓ | SSIM ↑ | PSNR ↑ | LPIPS ↓ | SSIM ↑ | PSNR ↑ | LPIPS ↓ | SSIM ↑ | PSNR ↑ |
| InfoNeRF | 0.5924 | 0.4342 | 12.27 | 0.6561 | 0.3792 | 10.57 | 0.6651 | 0.3843 | 10.62 |
| DietNeRF | 0.4381 | 0.6534 | 18.06 | 0.4636 | 0.6456 | 18.01 | 0.4853 | 0.6503 | 18.01 |
| RegNeRF | 0.4129 | 0.5916 | 17.14 | 0.4171 | 0.6132 | 17.86 | 0.4316 | 0.6257 | 18.34 |
| FreeNeRF | 0.5036 | 0.5354 | 14.70 | 0.4635 | 0.5708 | 15.26 | 0.5226 | 0.6027 | 16.31 |
| DS-NeRF | 0.2709 | 0.7983 | 26.26 | 0.2893 | 0.8004 | 26.50 | 0.3103 | 0.7999 | 26.65 |
| DDP-NeRF | 0.1290 | 0.8640 | 27.79 | 0.1518 | 0.8587 | 26.67 | 0.1563 | 0.8617 | 27.07 |
| ViP-NeRF | 0.0687 | 0.8889 | 32.32 | 0.0758 | 0.8967 | 31.93 | 0.0892 | 0.8968 | 31.95 |
| Simple-NeRF | **0.0635** | **0.8942** | **33.10** | **0.0726** | **0.8984** | **33.21** | **0.0847** | **0.8987** | **32.88** |

## 4.5   Experimental Results

We present the main results of our work with Simple-NeRF in Sec. 4.5.1 and then show the extension of our ideas to explicit models in Secs. 4.5.2 and 4.5.3.

Table 4.4: Evaluation of depth estimated by different NeRF based models with two input views. The reference depth is obtained using NeRF with dense input views. The depth MAE on the two datasets is of different orders on account of different depth ranges.

| model | NeRF-LLFF | | RealEstate-10K | |
|---|---|---|---|---|
| | MAE ↓ | SROCC ↑ | MAE ↓ | SROCC ↑ |
| DS-NeRF | 0.2074 | 0.7230 | 0.7164 | 0.6660 |
| DDP-NeRF | 0.2048 | 0.7480 | 0.4831 | 0.7921 |
| ViP-NeRF | 0.1999 | 0.7344 | 0.3856 | 0.8446 |
| Simple-NeRF | **0.1420** | **0.8480** | **0.3269** | **0.9215** |

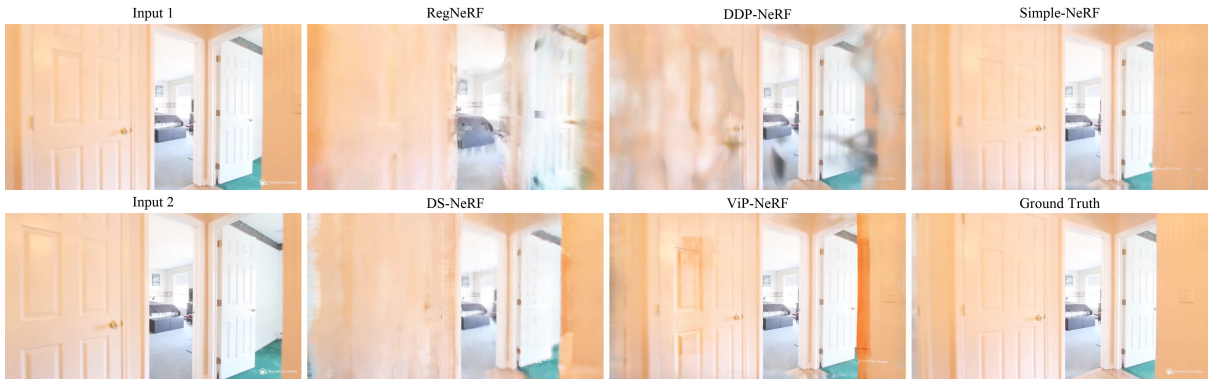

Figure 4.9: **Qualitative examples of NeRF based models on the RealEstate-10K dataset with two input views.** While DDP-NeRF predictions contain blurred regions, ViP-NeRF predictions are color-saturated in certain regions of the door. Simple-NeRF does not suffer from these distortions and synthesizes a clean frame. For reference, we also show the input images.

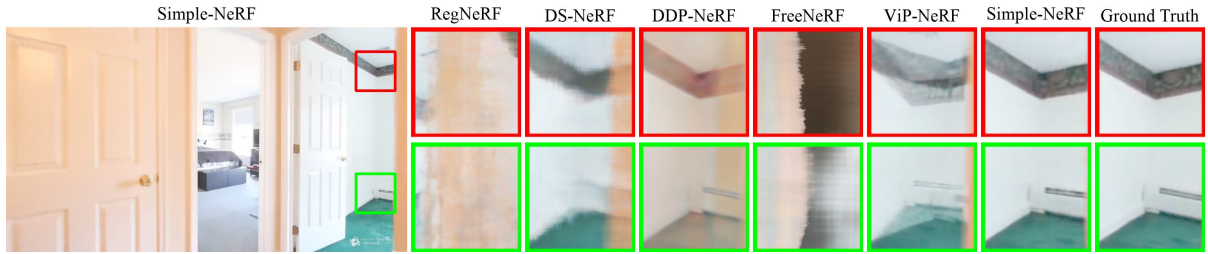

Figure 4.10: **Qualitative examples of NeRF based models on RealEstate-10K dataset** with three input views. Simple-NeRF predictions are closest to the ground truth among all the models. In particular, DDP-NeRF predictions have a different shade of color and ViP-NeRF suffers from shape-radiance ambiguity, creating ghosting artifacts.

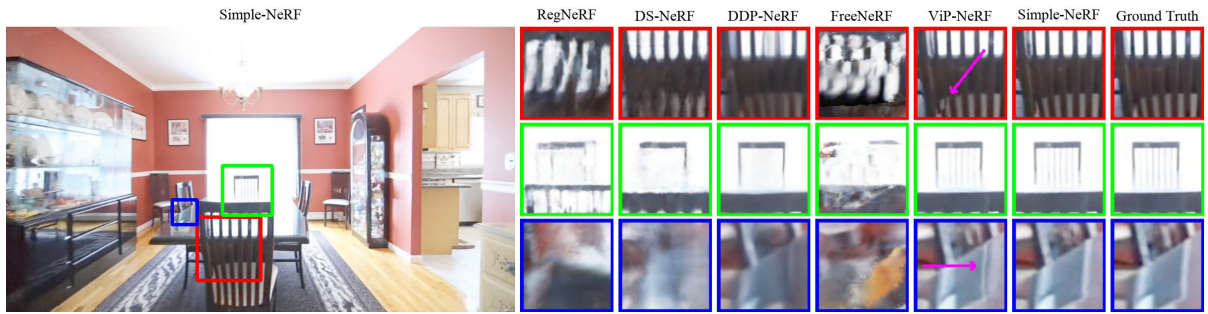

Figure 4.11: **Qualitative examples of NeRF based models on the RealEstate-10K dataset with four input views.** We find that Simple-NeRF and ViP-NeRF perform the best among all the models. However, ViP-NeRF predictions contain minor distortions, as pointed out by the magenta arrow, which is rectified by Simple-NeRF.

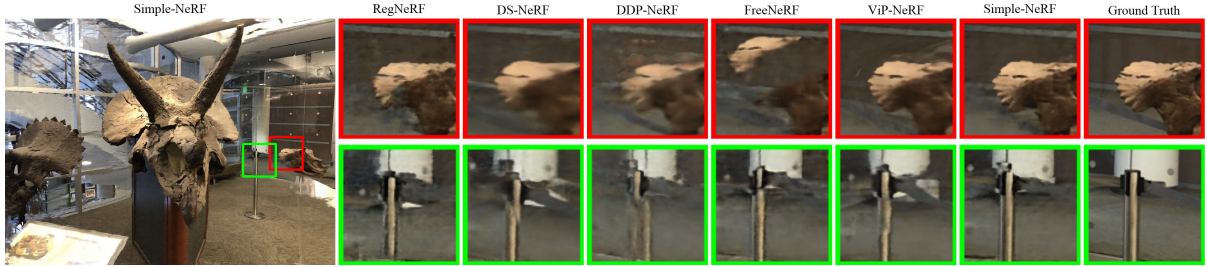

Figure 4.12: **Qualitative examples of NeRF based models on the NeRF-LLFF dataset with two input views.** DDP-NeRF and ViP-NeRF synthesize frames with broken objects in the second row, and FreeNeRF breaks the object in the first row due to incorrect depth estimations. Simple-NeRF produces sharper frames devoid of such artifacts.

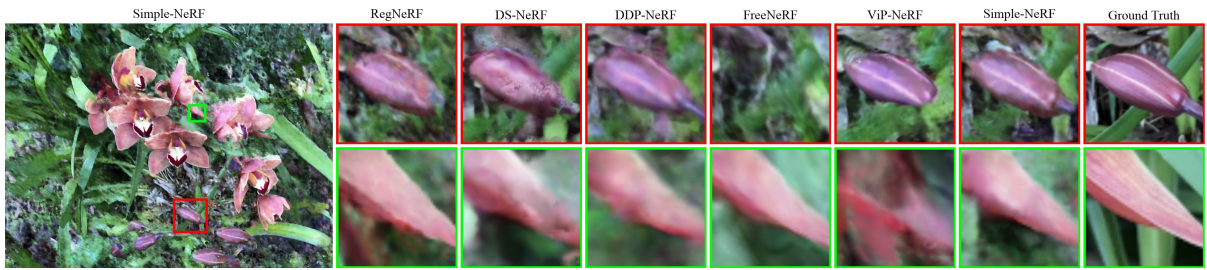

Figure 4.13: **Qualitative examples of NeRF based models on the NeRF-LLFF dataset with three input views.** In the first row, the orchid is displaced out of the cropped box in the FreeNeRF prediction, due to incorrect depth estimation. ViP-NeRF and RegNeRF fail to predict the complete orchid accurately and contain distortions at either end. In the second row, ViP-NeRF prediction contains severe distortions. Simple-NeRF reconstructs the best among all the models in both examples.

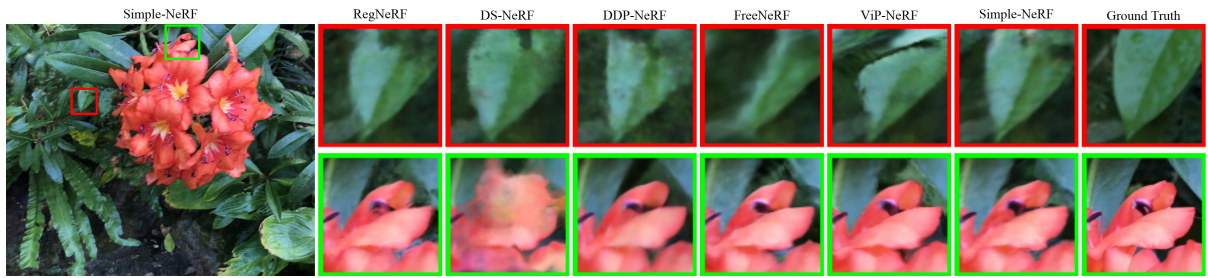

Figure 4.14: **Qualitative examples of NeRF based models on the NeRF-LLFF dataset with four input views.** In the first row, we find that ViP-NeRF, FreeNeRF, and DDP-NeRF struggle to reconstruct the shape of the leaf accurately. In the second row, DS-NeRF introduces floaters. Simple-NeRF does not suffer from such artifacts and reconstructs the shapes better.

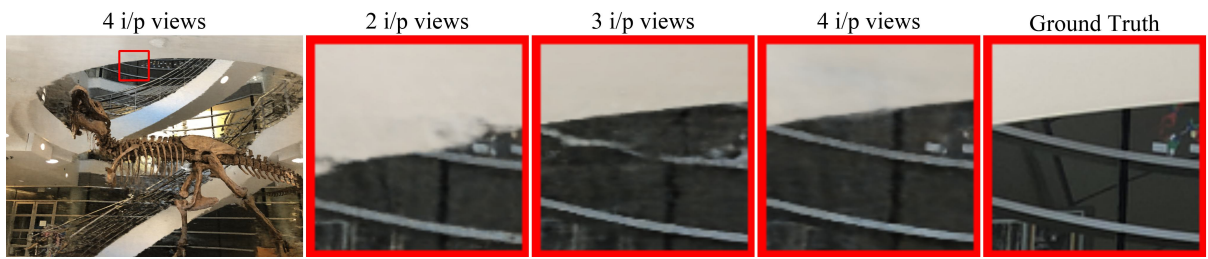

Figure 4.15: **Qualitative examples of Simple-NeRF on the NeRF-LLFF dataset with two, three, and four input views.** We observe errors in depth estimation with two input views, causing a change in the position of the roof. While this is corrected with three input views, there are a few shape distortions in the metal rods. With four input views, even such distortions are corrected.

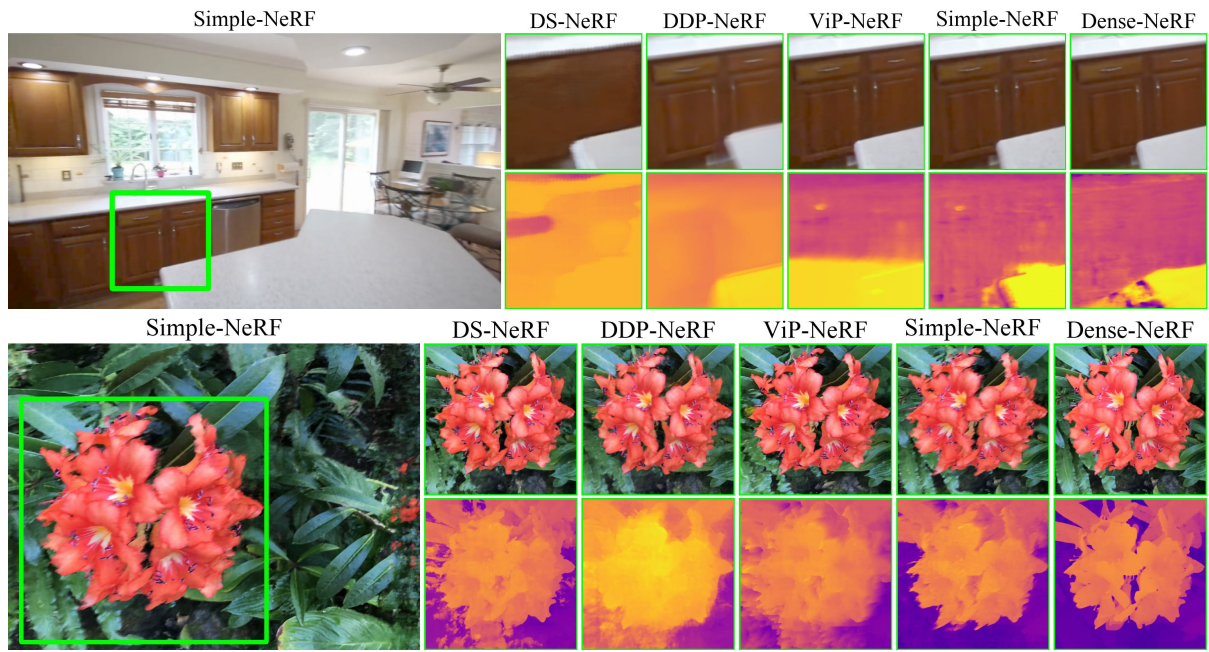

Figure 4.16: **Estimated depth maps of NeRF based models** on RealEstate-10K and NeRF-LLFF datasets with two input views. In both examples, the two rows show the predicted images and the depths respectively. We find that Simple-NeRF is significantly better at estimating the scene depth. Also, DDP-NeRF synthesizes the left table edge at a different angle due to incorrect depth estimation.

### 4.5.1  Simple-NeRF

#### 4.5.1.1  Comparisons

We evaluate the performance of our Simple-NeRF on the two forward-facing datasets only since the NeRF does not natively support unbounded 360 scenes. We evaluate the performance of our model against various sparse input NeRF models. We compare with DS-NeRF [46], DDP-NeRF [142] and RegNeRF [128] which regularize the depth estimated by the NeRF. We also evaluate DietNeRF [75] and InfoNeRF [83] that regularize the NeRF in hallucinated viewpoints. We also include two recent models, FreeNeRF [205] and ViP-NeRF [159], among the comparisons. We train the models on both datasets using the codes provided by the respective authors.

#### 4.5.1.2  Implementation details

We develop our code in PyTorch and on top of DS-NeRF [46]. We employ the Adam Optimizer with an initial learning rate of 5e-4 and exponentially decay it to 5e-6. We adjust the weights for the different losses such that their magnitudes after scaling are of similar orders. For the first 10k iterations of the training, we only impose $\mathcal{L}_m, \mathcal{L}_a$ and $\mathcal{L}_{sd}$. $\mathcal{L}_{aug}$ and $\mathcal{L}_{cfc}$ are imposed after 10k iterations. We set the hyper-parameters as follows: $l_p = 10$, $l_v = 4$, $l_p^s = 3$, $k = 5$, $e_\tau = 0.1$, $\lambda_m = \lambda_a = 1$, $\lambda_{sd} = \lambda_{aug} = \lambda_{cfc} = 0.1$ and $\lambda_{mc} = 0$. The network architecture is exactly the same as DS-NeRF. For the augmented models, we only change the input dimension of the MLPs $\mathcal{N}_1$ and $\mathcal{N}_2$ appropriately. The augmented models are employed only during training, and the network is exactly the same as Vanilla NeRF for inference. We train the models on a single NVIDIA RTX 2080 Ti GPU for 100k iterations.

#### 4.5.1.3  Quantitative and Qualitative Results

Tabs. 4.2 and 4.3 show the view-synthesis performance of Simple-NeRF and other prior art on NeRF-LLFF and RealEstate-10K datasets respectively. We find that Simple-NeRF achieves state-of-the-art performance on both datasets in most cases. The higher performance of all the models on the RealEstate-10K dataset is perhaps due to the scenes being simpler. Hence, the performance improvement is also smaller as compared to the NeRF-LLFF dataset. Fig. 4.9 shows predictions of various models on an example scene from the RealEstate-10K dataset, where we observe that Simple-NeRF is the best in reconstructing the novel view. Figs. 4.10 to 4.15 show more comparisons on both datasets with 2, 3, and 4 input views. Further, Simple-NeRF improves significantly in estimating the depth of the scene as seen in Tab. 4.4 and Fig. 4.16. We provide video comparisons on our project webpage[1].

We note that the quantitative results in Tabs. 4.2 and 4.3 differ from the values reported in Chapter 3(ViP-NeRF) on account of the following two differences. Firstly, the quality evaluation metrics are computed on full frames in ViP-NeRF. However, we exclude the regions not seen in the input views as explained in Sec. 4.4.2. Secondly, while we use the same train set as that of ViP-NeRF on the RealEstate-10K dataset, we modify the test set as shown in Tab. 4.1. We change the test set since the test views that are very far away from the train views may contain large unobserved regions.

### 4.5.1.4 Ablations

We test the importance of each of the components of our model by disabling them one at a time. We disable the smoothing and Lambertian augmentations and coarse-fine consistency loss individually. When disabling $\mathcal{L}_{cfc}$, we additionally add augmentations to the fine NeRF since the knowledge learned by coarse NeRF may not efficiently propagate to the fine NeRF. We also analyze the need to supervise with only the reliable depth estimates by disabling the mask and stop-gradients in $\mathcal{L}_{aug}$ and $\mathcal{L}_{cfc}$. In addition, we also analyze the effect of including residual positional encodings $\gamma(\mathbf{p}_i, l_p^s, l_p)$ while predicting the color in the smoothing augmentation model. Tab. 4.5 shows a quantitative comparison between the ablated models. We observe that each of the components is crucial, and disabling any of them leads to a drop in performance. Further, using all the

---

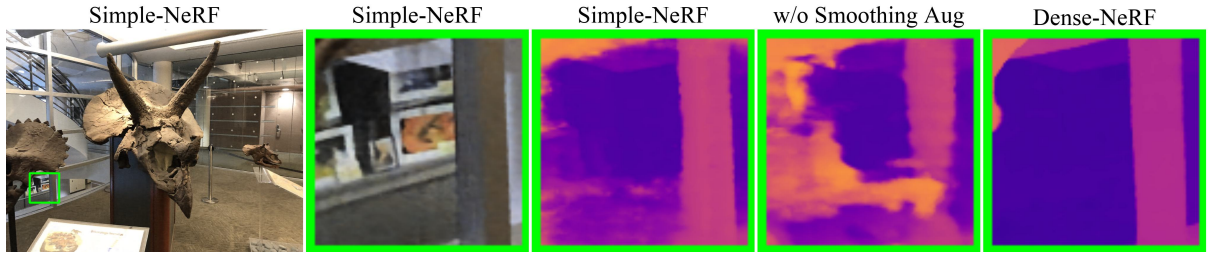

(a) *Without smoothing augmentation:* The ablated model introduces floaters that are significantly reduced by using the smoothing augmentation model.

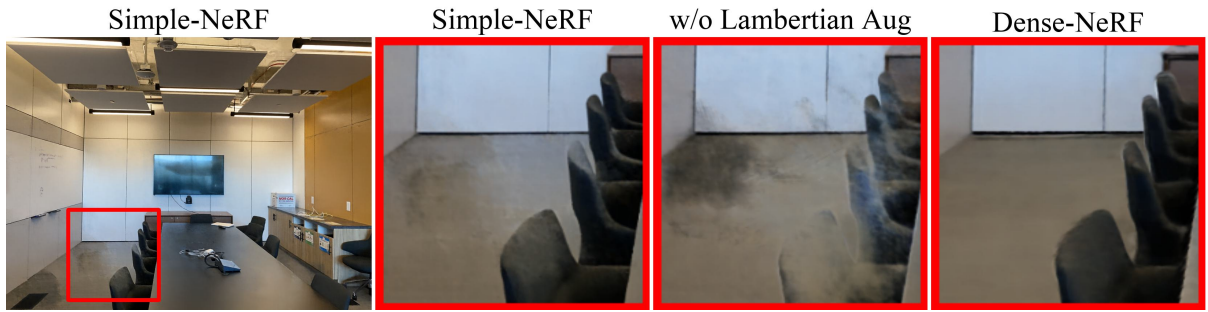

(b) *Without Lambertian augmentation:* The ablated model suffers from shape-radiance ambiguity and produces ghosting artifacts.

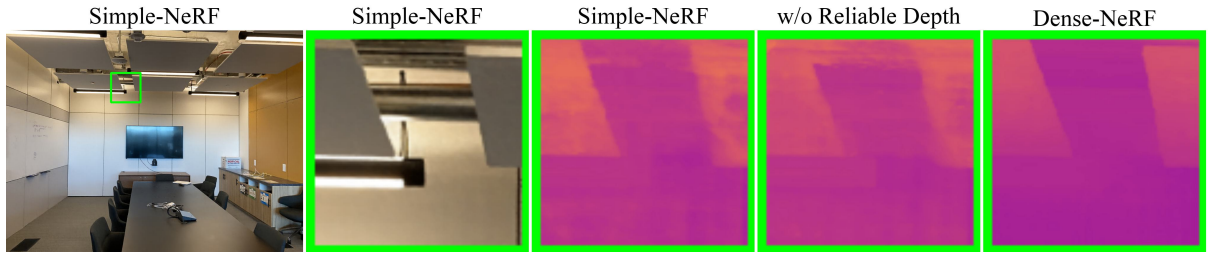

(c) *Without reliability of depth supervision:* The smoothing augmentation model struggles to learn sharp depth discontinuities at true depth edges. Supervising the main model using such depths without determining their reliability causes the main model to learn incorrect depth. As a result, the ablated model fails to learn sharp depth discontinuities at certain regions.

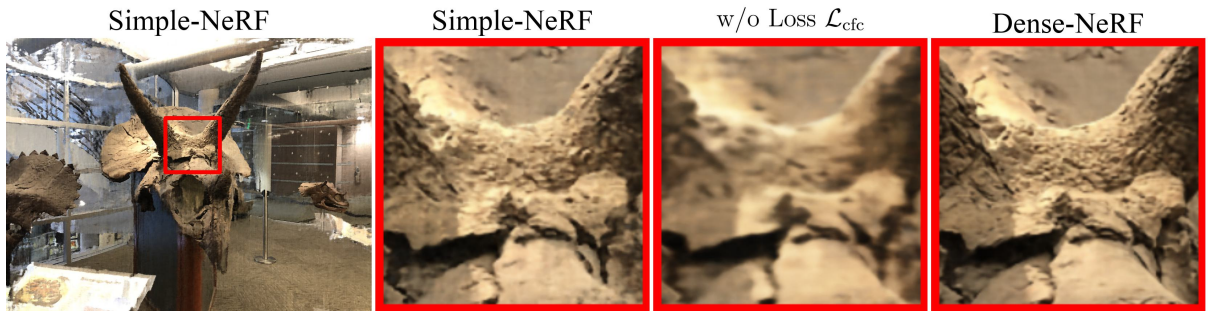

(d) *Without coarse-fine consistency:* We observe that while Simple-NeRF predictions are sharper, the ablated model without coarse-fine consistency loss, $\mathcal{L}_{cfc}$ produces blurred renders. This is similar to Fig. 4.6a, where we observe DS-NeRF also produce blurred renders.

Figure 4.17: **Qualitative examples for Simple-NeRF ablated models on the NeRF-LLFF dataset with two input views.**

Table 4.5: SimpleNeRF ablation experiments on RealEstate-10K and NeRF-LLFF datasets with two input views.

| model | RealEstate-10K | | NeRF-LLFF | |
|---|---|---|---|---|
| | LPIPS ↓ | MAE ↓ | LPIPS ↓ | MAE ↓ |
| Simple-NeRF | **0.0635** | **0.33** | **0.2688** | **0.14** |
| w/o smoothing augmentation | 0.0752 | 0.38 | 0.2832 | 0.15 |
| w/o Lambertian augmentation | 0.0790 | 0.39 | 0.2834 | 0.15 |
| w/o coarse-fine consistency | 0.0740 | 0.42 | 0.3002 | 0.19 |
| w/o reliable depth | 0.0687 | 0.45 | 0.3020 | 0.22 |
| w/o residual pos enc | 0.0790 | 0.40 | 0.2837 | 0.16 |
| w/ identical augmentations | 0.0777 | 0.40 | 0.2849 | 0.15 |
| w/ smaller n/w as smoothing aug | 0.0740 | 0.38 | 0.2849 | 0.15 |

depths for supervision instead of only the reliable depths leads to a significant drop in performance. Finally, disabling $\mathcal{L}_{cfc}$ also leads to a drop in performance in addition to increasing the training time by almost 2× due to the inclusion of augmentations for the fine NeRF.

Since we design our regularizations on top of DS-NeRF [46] baseline, our framework can be seen as a semi-supervised learning model by considering the sparse depth from a Structure from Motion (SfM) module as providing limited depth labels and the remaining pixels as the unlabeled data. Our approach of using augmented models in tandem with the main radiance field model is perhaps closest to the Dual-Student architecture [80] that trains another identical model in tandem with the main model and imposes consistency regularization between the predictions of the two models. However, our augmented models have complementary abilities as compared to the main radiance field model. We now analyze if there is a need to design augmentations that learn "simpler" solutions by replacing our novel augmentations with identical replicas of the NeRF as augmentations. The seventh row of Tab. 4.5 shows a performance drop when using identical augmentations.

Finally, we analyze the need for an augmentation that explicitly achieves depth smoothing. In other words, we ask if naively reducing the model capacity in the augmented model achieves a similar effect to that of our smoothing augmentation. We test this by replacing the smoothing augmentation with an augmented model that has a smaller MLP $\mathcal{N}_1$. Specifically, we reduce the number of layers from eight to four in the augmented model. From the results in the last row of Tab. 4.5, we conclude that reducing the positional encoding degree is more effective, perhaps because the MLP with fewer layers may still be capable of learning floaters on account of using all the positional encoding frequencies.

### 4.5.1.5   Visualization of Depth Reliability Masks

In Fig. 4.18, we present visualizations that motivate the design of our augmentations in Simple-NeRF, namely the smoothing and Lambertian augmentations. We train our model without augmentations and the individual augmentations separately with only $\mathcal{L}_{ph}$ and $\mathcal{L}_{sd}$ for 100k iterations. Using the depth maps predicted by the models for an input training view, we determine the mask that indicates which depth estimates are more accurate, as explained in Sec. 4.3.1.4. For two scenes from the LLFF dataset, we show an input training view and focus on a small region to visualize the corresponding masks.

We observe that the smoothing augmentation is determined to have estimated better depths in smooth regions. At edges, the depth estimated by the main model is more accurate. Similarly, the Lambertian augmentation estimates better depth in Lambertian regions, while the main model estimates better depth in specular regions. We note that the masks shown are not the masks obtained by our final model. Since the masks are computed at every iteration, and the training of the main and augmented models are coupled, it is not possible to determine the exact locations where the augmented models help the main model learn better.

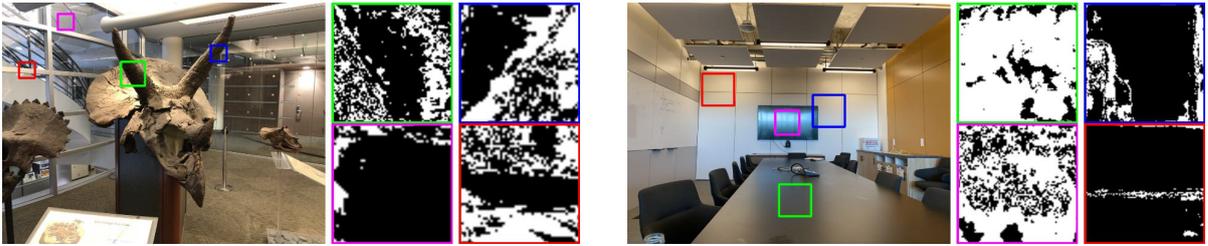

(a) **Smoothing augmentation:** The green and blue boxes focus on the two horns, where we observe that the augmented model depth is preferred in the depth-wise smooth regions on horns, and the main model depth is preferred at the edges. The magenta box focuses on a completely smooth region, so the augmented model depth is preferred for most pixels. In the red box, augmented model depth is preferred along the horizontal bar. The main model depth is preferred on either side of the bar that contains multiple depth discontinuities.

(b) **Lambertian augmentation:** The green and magenta boxes focus on the TV and the table, respectively, which are highly specular in this scene (please view the supplementary videos of the room scene to observe the specularity of these objects). In these regions, the main model depth is determined to be more accurate since the main model can handle specular regions. The red and blue boxes focus on Lambertian regions of the scene where the depth estimated by the augmented model is preferred.

Figure 4.18: Visualizations of depth reliability mask for the two augmentations of Simple-NeRF. White pixels in the mask indicate that the main model depth is determined to be more accurate at the corresponding locations. Black pixels indicate that the augmented model depth is determined to be more accurate.

Table 4.6: Quantitative results of TensoRF based models with three input views.

| Model | NeRF-LLFF | | | | | RealEstate-10K | | | | |
|---|---|---|---|---|---|---|---|---|---|---|
| | LPIPS ↓ | SSIM ↑ | PSNR ↑ | Depth MAE ↓ | Depth SROCC ↑ | LPIPS ↓ | SSIM ↑ | PSNR ↑ | Depth MAE ↓ | Depth SROCC ↑ |
| TensoRF | 0.5474 | 0.3163 | 12.29 | 0.67 | 0.03 | 0.0986 | 0.8532 | 29.62 | 0.44 | 0.63 |
| DS-TensoRF | 0.2897 | 0.6291 | 18.58 | 0.23 | 0.73 | 0.0739 | 0.8872 | 32.50 | 0.27 | 0.75 |
| Simple-TensoRF | **0.2461** | **0.6749** | **20.22** | **0.17** | **0.83** | **0.0706** | **0.8920** | **32.70** | **0.22** | 0.80 |
| $R^s_\sigma = R_\sigma$ | 0.2536 | 0.6677 | 19.85 | 0.18 | 0.81 | 0.085 | 0.8821 | 30.94 | 0.27 | 0.77 |
| $N^s_{vox} = N_{vox}$ | 0.2568 | 0.6579 | 19.95 | 0.19 | 0.79 | 0.0735 | 0.8896 | 32.22 | **0.22** | **0.82** |
| $R^s_\sigma = R_\sigma; N^s_{vox} = N_{vox}$ | 0.2728 | 0.6424 | 19.50 | 0.22 | 0.74 | 0.0787 | 0.8871 | 31.73 | 0.23 | 0.79 |

### 4.5.2  Simple-TensoRF

#### 4.5.2.1  Implementation Details

Building on the original TensoRF code base, we employ Adam Optimizer with an initial learning rate of $2e - 2$ and $1e - 3$ for the tensor and MLP parameters respectively, which decay to $2e - 3$ and $1e - 4$. We employ the same hyper-parameters as the original implementation for the main model as follows: $R_\sigma = 24$, $R_c = 72$, $\mathbf{b} = \{(-1.5, 1.5), (-1.67, 1.67), (-1.0, 1.0)\}$, $N_{vox} = 640^3$, $D = 27$, and $l_v = 0$. We set $R^s_\sigma = 12$, $b^s_{z_1} = -0.5$ and $N^s_{vox} = 160^3$, $N_{mc} = 5$, $k = 5$, $e_\tau = 0.1$ for the augmented model and the remaining hyper-parameters are the same as the main model. We weigh the losses as $\lambda_m = \lambda_a = 1$, $\lambda_{sd} = \lambda_{aug} = 0.1$, $\lambda_{mc} = 0.01$ and $\lambda_{cfc} = 0$. We train the models on a single NVIDIA RTX 2080 Ti 11GB GPU for 25k iterations and enable $\mathcal{L}_{aug}$ after 5k iterations.

#### 4.5.2.2  Quantitative and Qualitative Results

Tab. 4.6 shows the view-synthesis performance of Simple-TensoRF on the NeRF-LLFF and RealEstate-10K datasets. We compare the performance of our model against the vanilla TensoRF and a baseline we create by adding sparse depth loss on TensoRF, which we refer to as DS-TensoRF. We find that Simple-TensoRF significantly improves performance over TensoRF and DS-TensoRF on both datasets. Fig. 4.19 compares the three models visually, where we observe that Simple-TensoRF mitigates multiple distortions

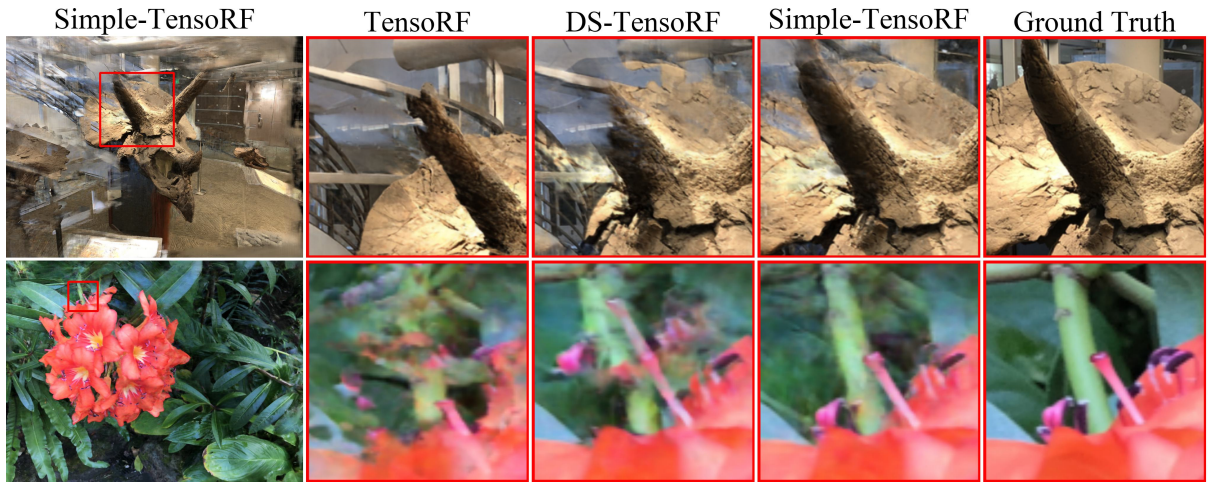

(a) *NeRF-LLFF dataset:* In the first example, we find that the horn is broken and almost half of the bony frill is missing in the renders of TensoRF and DS-TensoRF. In the second example, TensoRF and DS-TensoRF extend the red stigma and break the green stem. Simple-TensoRF does not introduce such distortions and is closest to the ground truth.

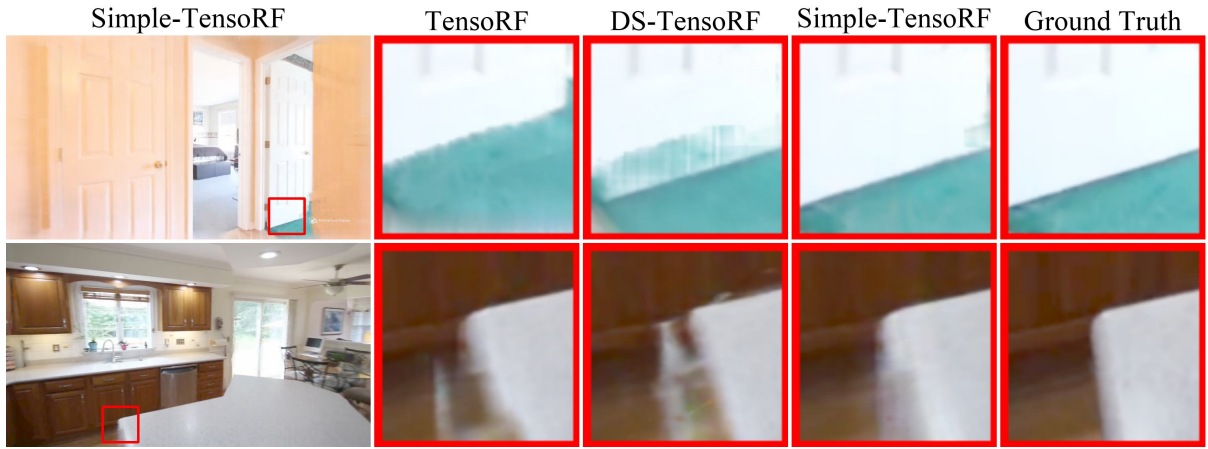

(b) *RealEstate-10K dataset:* In the first example, we observe a shift in the position of the door due to incorrect depth estimation. With sparse depth supervision, DS-TensoRF moves the door to the correct position, but only partially. Adding our augmentations provides the best result. Similarly, we see distortions in the frames rendered by TensoRF and DS-TensoRF in the second example, which are reduced significantly by Simple-TensoRF.

Figure 4.19: **Qualitative examples of TensoRF based models with three input views.**

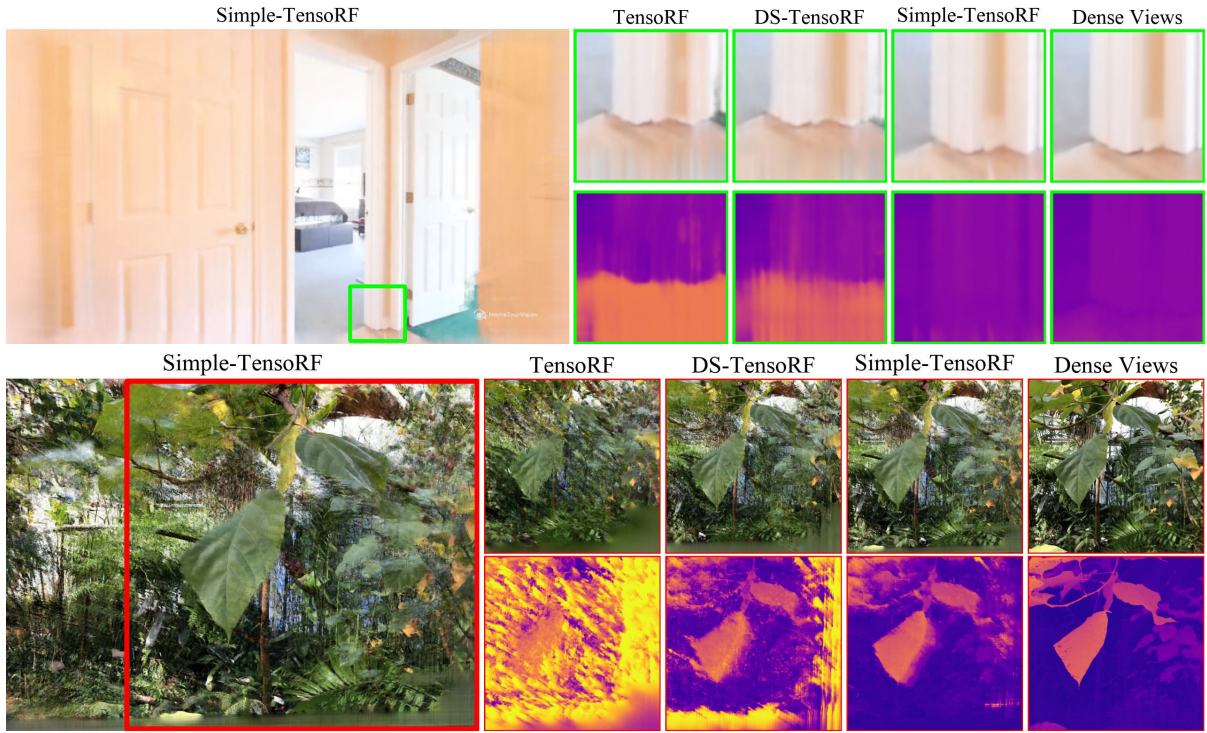

Figure 4.20: **Estimated depth maps of TensoRF based models** on RealEstate-10K and NeRF-LLFF datasets with three input views. In both examples, the two rows show the predicted images and the depths respectively. In the first example, TensoRF and DS-TensoRF incorrectly estimate the depth of the floor as shown by the orange regions. In the second row, while TensoRF is unable to estimate the scene geometry, DS-TensoRF is unable to mitigate all the floaters in orange color. We find that Simple-TensoRF is significantly better at estimating the scene depth.

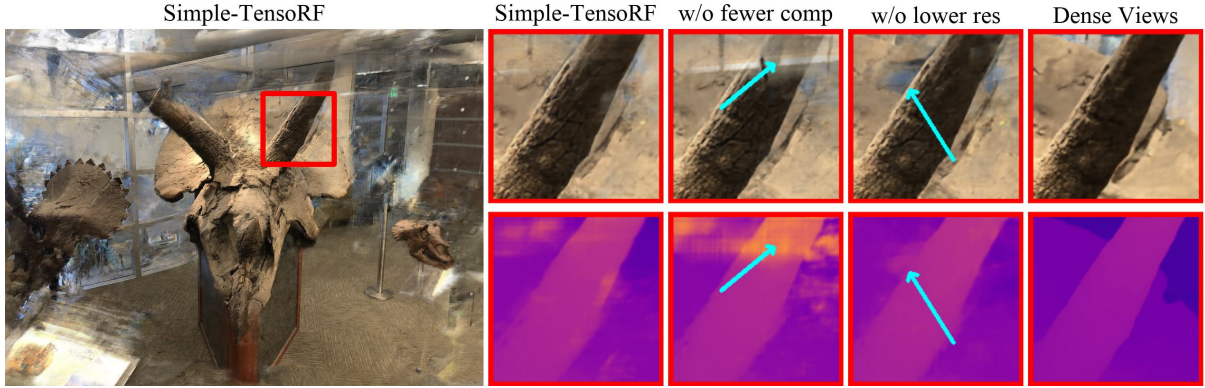

Figure 4.21: **Qualitative examples of Simple-TensoRF ablations** on NeRF-LLFF dataset with three input views. Reducing the tensor resolution only leads to translucent floaters as shown by the arrows in the second column. On the other hand, only reducing the number of tensor decomposed components leads to small opaque floaters as shown by the arrows in the third column.

observed in the renders of TensoRF and DS-TensoRF. From Tab. 4.6 and Fig. 4.20, we observe that Simple-TensoRF is significantly better at estimating the scene depth than both TensoRF and DS-TensoRF. While we observe that Simple-NeRF performs marginally better than Simple-TensoRF in most cases, Simple-TensoRF achieves a lower depth MAE on the RealEstate-10K dataset.

We test the need for the different components of our augmentation by disabling them one at a time and show the quantitative results in the second half of Tab. 4.6. Specifically, we disable the reduction in the number of tensor decomposition components and the number of voxels in the first two rows respectively. In the third row, we disable both the components, where the augmented model consists of the reduction in the bounding box size only. We find that disabling either or both of the components leads to a drop in performance. In particular, Fig. 4.21 shows that reducing only the tensor resolution and not reducing the number of tensor decomposition components leads to translucent blocky floaters. On the other hand, reducing only the number of components causes small and completely opaque floaters. Further, we find that reducing the number of components $R_\sigma$ is more crucial in obtaining simpler solutions on the RealEstate-10K dataset.

Table 4.7: Quantitative results of ZipNeRF based models on the MipNeRF360 dataset.

| Model | 12 input views | | | | | 20 input views | | | 36 input views | | |
|---|---|---|---|---|---|---|---|---|---|---|---|
| | LPIPS ↓ | SSIM ↑ | PSNR ↑ | Depth MAE ↓ | Depth SROCC ↑ | LPIPS ↓ | SSIM ↑ | PSNR ↑ | LPIPS ↓ | SSIM ↑ | PSNR ↑ |
| ZipNeRF | 0.5614 | 0.4616 | 15.86 | 7.43 | 0.28 | 0.435 | 0.5911 | 18.89 | 0.3316 | 0.6737 | 21.78 |
| Augmented ZipNeRF | 0.6825 | 0.4462 | 16.27 | 96.42 | 0.49 | 0.619 | 0.5244 | 19.31 | 0.5917 | 0.5646 | 21.21 |
| Simple-ZipNeRF | **0.4876** | **0.5245** | **17.60** | **3.54** | **0.51** | **0.3421** | **0.6456** | **21.03** | **0.239** | **0.7458** | **24.19** |

Table 4.8: Quantitative results of ZipNeRF based models on the NeRF-Synthetic dataset.

| Model | 4 input views | | | 8 input views | | | 12 input views | | |
|---|---|---|---|---|---|---|---|---|---|
| | LPIPS ↓ | SSIM ↑ | PSNR ↑ | LPIPS ↓ | SSIM ↑ | PSNR ↑ | LPIPS ↓ | SSIM ↑ | PSNR ↑ |
| ZipNeRF | 0.4263 | 0.7548 | 11.04 | 0.2877 | 0.7973 | 15.01 | 0.1625 | 0.8528 | 20.12 |
| Simple-ZipNeRF | **0.3878** | **0.7715** | **11.50** | **0.2461** | **0.8063** | **15.88** | **0.1532** | **0.8531** | **20.51** |

### 4.5.3 Simple-ZipNeRF

#### 4.5.3.1 Implementation Details

We build our code in PyTorch on top of an unofficial ZipNeRF implementation[2]. For the main model, we retain the hyper-parameters of the original ZipNeRF. For the augmented model, we reduce the size of the hash table from $T = 2^{21}$ to $T^s = 2^{11}$ and set $s_{near} = 0.3$. We impose $\mathcal{L}_{aug}$ after 5k iterations and use $k = 5$, $e_\tau = 0.2$. The rest of the hyper-parameters for the augmented model are the same as the main model. We weigh the losses as $\lambda_m = \lambda_a = 1$, $\lambda_{aug} = 10$ and $\lambda_{sd} = \lambda_{cfc} = \lambda_{mc} = 0$. We do not impose the sparse depth loss $\mathcal{L}_{sd}$ since we find that Colmap either fails in sparse reconstruction or provides noisy sparse depth for 360° scenes. We train the models on a single NVIDIA RTX 2080 Ti 11GB GPU for 25k iterations.

#### 4.5.3.2 Quantitative and Qualitative Results

We compare the performance of ZipNeRF with and without our augmentations on the MipNeRF360 and NeRF-Synthetic datasets in Tabs. 4.7 and 4.8 respectively. We observe that including our augmentations improves performance significantly on both datasets

---

[2]ZipNeRF implementation: https://github.com/SuLvXiangXin/zipnerf-pytorch

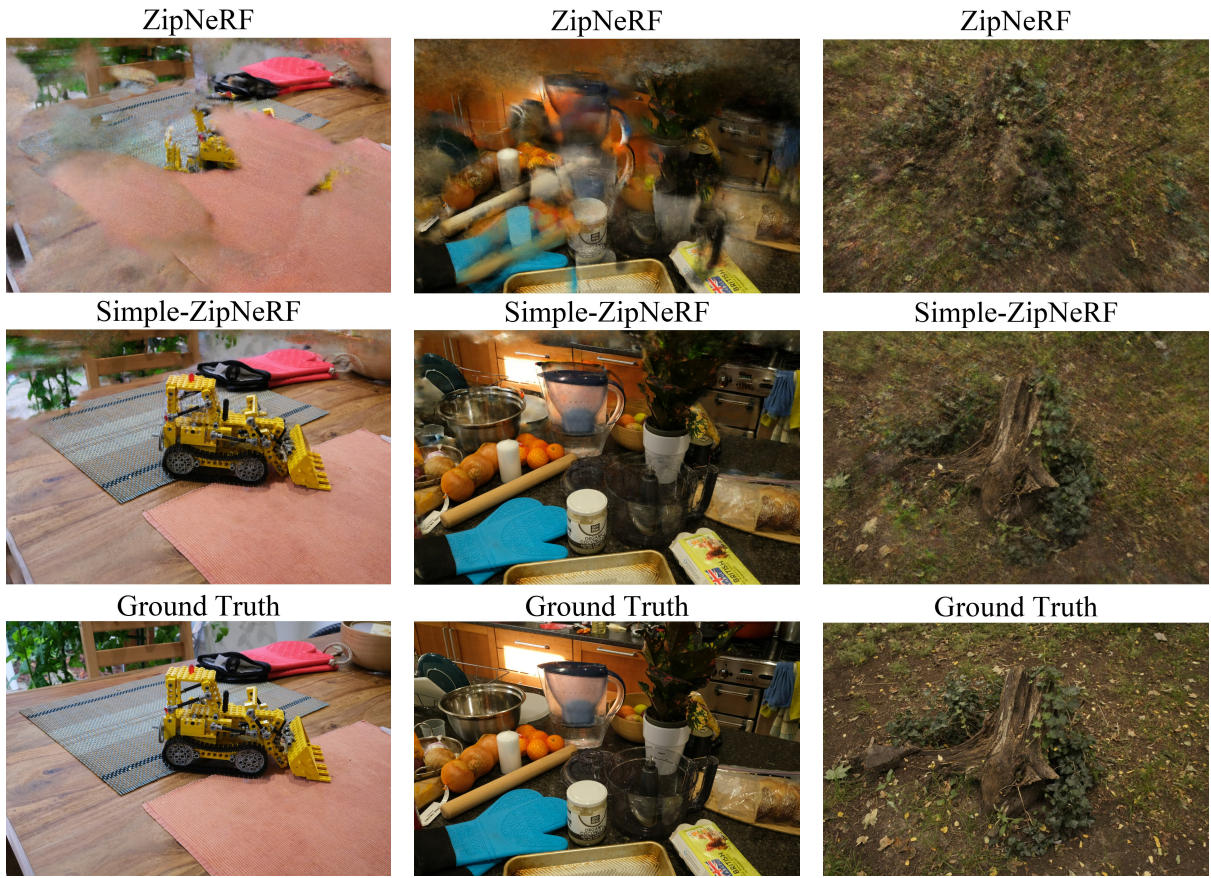

(a) 12 input views.          (b) 20 input views.          (c) 36 input views.

Figure 4.22: **Qualitative examples of ZipNeRF and Simple-ZipNeRF on the MipNeRF360 dataset.** In the first column, we observe that ZipNeRF places large regions of the pink mat close to the camera, occluding the bulldozer. In the second example, we observe objects being broken or placed at incorrect positions due to incorrect depth estimation, as well as translucent floaters in ZipNeRF predictions. Finally, in the third column, we observe that ZipNeRF fails to reconstruct the tree stump. In all the cases, Simple-ZipNeRF produces very good reconstructions without any floaters.

Simple-ZipNeRF          ZipNeRF          Simple-ZipNeRF          Dense-ZipNeRF

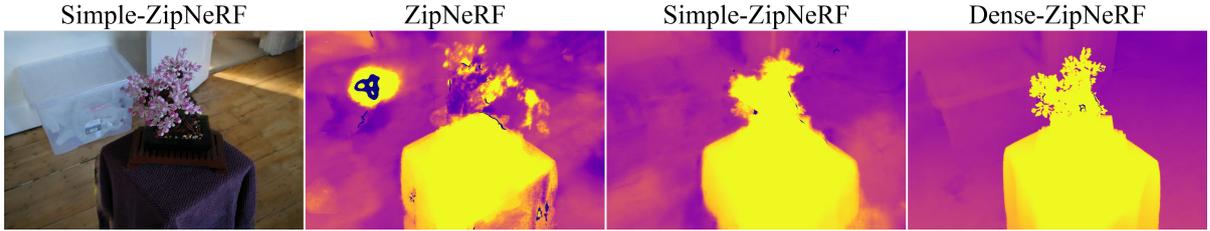

Figure 4.23: **Simple-ZipNeRF estimated depth maps** on MipNeRF360 dataset with 20 input views. We observe that the depth map estimated by ZipNeRF contains floaters and that the depth estimates for the bonsai are incorrect. However, Simple-ZipNeRF does not suffer from such issues and the estimated depth is very close to that of ZipNeRF with dense input views.

in terms of all the evaluation measures. This observation is further supported by the qualitative examples in Figs. 4.22 to 4.24, where we observe a clear improvement in the quality of the rendered novel views and depth when employing our augmentations. In addition, Tab. 4.7 and Fig. 4.25 also shows the performance of the augmented model on the MipNeRF360 dataset. We observe a significant reduction in distortions in the renders of the augmented model; however, the same does not reflect in the quantitative evaluation due to the blur introduced by the augmented model.

Further, in Fig. 4.26, we show the performance of ZipNeRF and Simple-ZipNeRF as the number of input views increases. We observe that the performance of ZipNeRF is too low with very few input images, where our augmentation does not help improve the performance. As the number of input views increases, the performance of ZipNeRF improves, and our augmentation helps improve the performance significantly. Further, with a large number of input views, the performance of ZipNeRF saturates, and our augmentation does not help improve the performance. This shows that our augmentations are highly effective when the performance of the base model is moderately good.

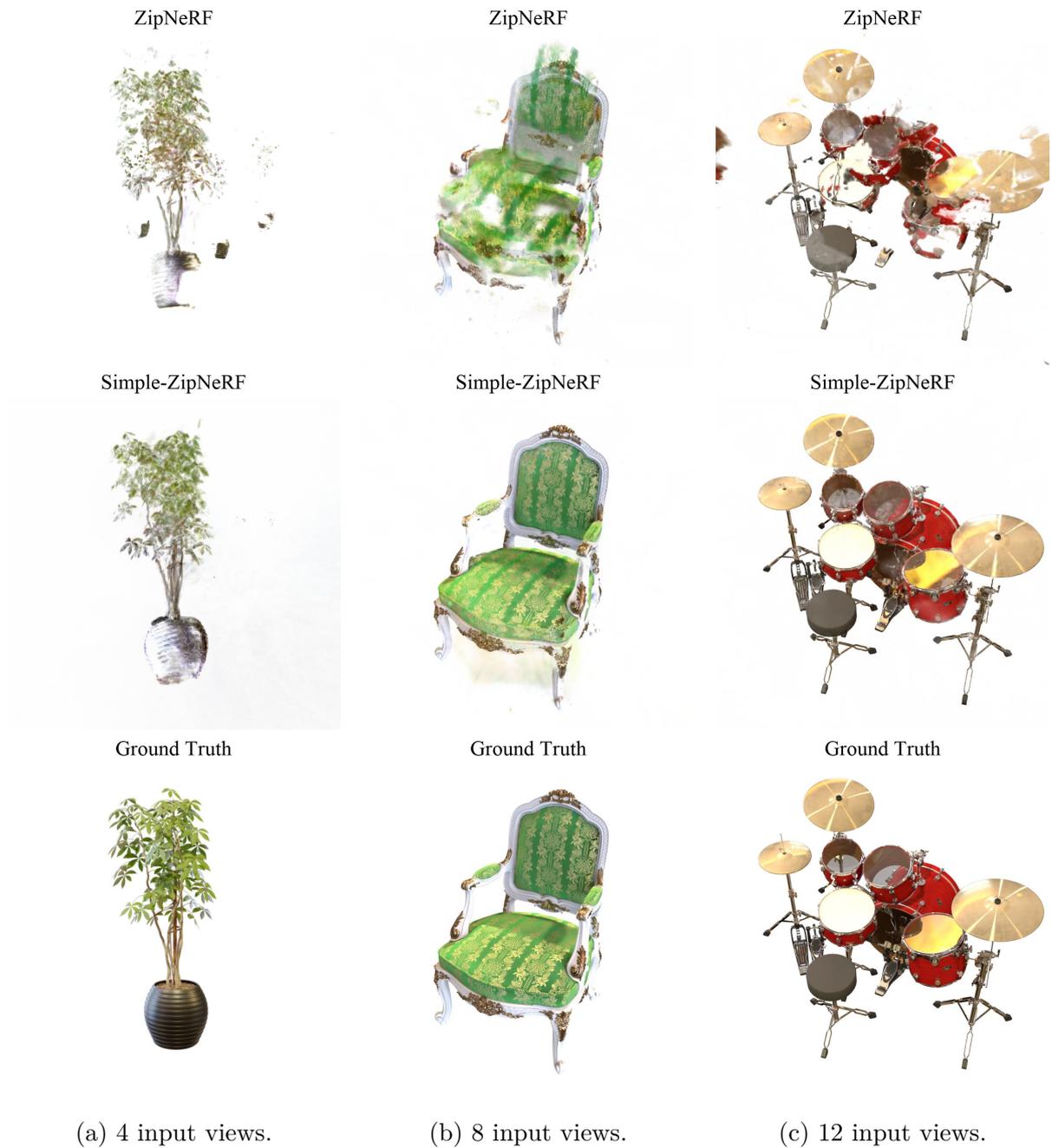

(a) 4 input views.              (b) 8 input views.              (c) 12 input views.

Figure 4.24: **Qualitative examples of ZipNeRF and Simple-ZipNeRF on the NeRF-Synthetic dataset.** While the renders of ZipNeRF contain multiple floaters, Simple-ZipNeRF outputs are cleaner and free from such artifacts.

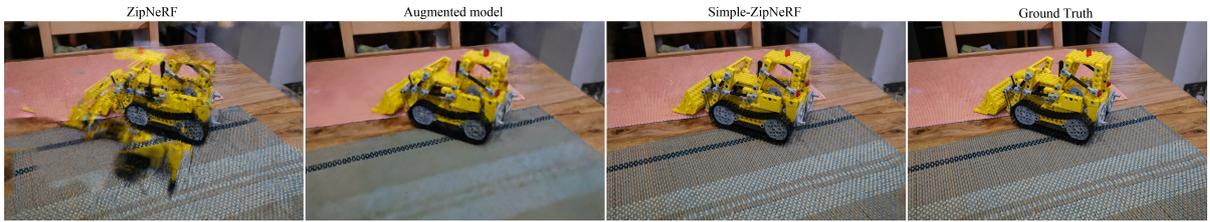

Figure 4.25: **Qualitative examples to visualize the effect of our augmentation.**
We observe that the ZipNeRF render contains severe distortions.  The output of our
augmented model is significantly better in reconstructing the scene, but the render con-
tains severe blur on account of smoothing introduced by small hash table.  Learning from
the depth provided by the augmented model, Simple-ZipNeRF is able to reconstruct the
scene better as well as retain sharpness by utilizing the larger hash table.

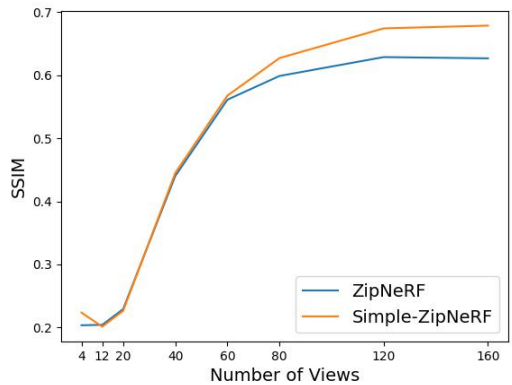

(a) MipNeRF360 Bicycle scene.

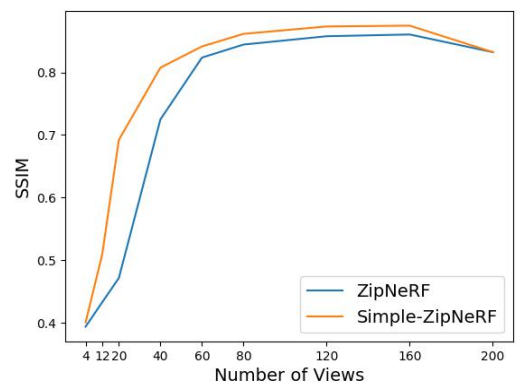

(b) MipNeRF360 Counter scene.

Figure 4.26: **Performance of ZipNeRF and Simple-ZipNeRF with increasing
number of input views.**  We observe that our augmentation improves performance
significantly over ZipNeRF, when the performance of the base model is moderate.  When
the performance of the base model is extremely poor or extremely good, the augmen-
tation does not have a significant impact.  However, our augmentation does not lead to
significant degradation in performance in either case.

Table 4.9: Training and inference (per frame) time and memory comparison of various models.

| Model | Training | | Inference | |
|---|---|---|---|---|
| | Time (hrs) | Mem (GB) | Time (sec) | Mem (GB) |
| NeRF | 14 | 6.1 | 54 | 0.8 |
| Simple-NeRF | 21 | 8.8 | 54 | 0.8 |
| TensoRF | 2.1 | 6.8 | 21 | 4.0 |
| Simple-TensoRF | 3.7 | 7.2 | 21 | 4.0 |
| ZipNeRF | 2.0 | 6.7 | 13 | 4.8 |
| Simple-ZipNeRF | 4.2 | 8.6 | 13 | 4.8 |

## 4.6 Discussion

### 4.6.1 Computational Complexity

We report the approximate GPU memory utilization and time taken for training and inference of our family of Simple-RF models in Tab. 4.9. We observe that Simple-NeRF with two augmentations takes only 1.5 times more time than NeRF for training on account of employing augmentations on the coarse NeRF only. While coarse NeRF queries the MLPs 64 times, the fine NeRF queries the MLPs 192 times, giving a combined 256 queries per pixel. Simple-NeRF queries the coarse MLPs 192 times and the fine MLPs 192 times, with a total of 384 queries per pixel. On the other hand, Simple-TensoRF and Simple-ZipNeRF take twice the time as TensoRF and ZipNeRF respectively on account of employing a single augmentation with exactly the same number of queries as the main model. We note that it could be possible to further reduce the training time for ZipNeRF by employing the augmentation only on the proposal MLP. However, this requires the proposal MLPs to output color and to be trained with the photometric loss instead of the interval loss [18]. The effect of such a change is unclear and is left for future work. Interestingly, Simple-TensoRF requires only a little more memory than TensoRF during training, perhaps due to the low resolution tensor employed by the augmented model. Further, while the NeRF models require significantly less memory during inference, grid based models such as TensoRF and ZipNeRF require more memory due to the use of a voxel grid in place of MLPs. Finally, we note that at inference time, Simple-RF models take exactly the same time and memory as the baseline models since the augmentations are disabled during inference. All the above experiments are conducted on a single NVIDIA RTX 2080 Ti 11GB GPU.

## 4.7   Summary

We address the problem of few-shot radiance fields by obtaining depth supervision from simpler solutions learned by lower capability augmented models that are trained in tandem with the main radiance field model. We show that augmentations can be designed for both implicit models, such as NeRF, and explicit radiance fields, such as TensoRF and ZipNeRF. Since the shortcomings of various radiance fields are different, we design the augmentations appropriately for each model. We show that our augmentations improve performance significantly on all three models, and we achieve state-of-the-art performance on forward-facing scenes as well as 360° scenes. Notably, our models achieve a significant improvement in the depth estimation of the scene, which indicates a superior geometry estimation.

# Chapter 5

# Factorized Motion Fields for Fast Sparse Input Dynamic View Synthesis

## 5.1 Introduction

Although Neural Radiance Fields (NeRF) [123] brought in a seminal shift in novel view synthesis by incorporating differential volume rendering with a compact continuous-depth model, it has several limitations such as the inability to handle object motion in the scene as well as long training and rendering times. Recently, K-Planes [53] significantly reduced the time complexity for dynamic view synthesis by proposing a factorized 4D volume representation. However, K-Planes requires a large number of input viewpoints to render photo-realistic novel views when employing a multi-view camera setup. In this chapter, we overcome this limitation and design a fast dynamic radiance field that can effectively learn the dynamic scene with few input viewpoints. We mainly focus on a sparse multi view camera setting, where a video from each viewpoint is available. Tab. 5.1 shows how our model relates to other models in the literature.

---

This chapter is based on the work accepted at SIGGRAPH 2024 [155].

Optimizing a dynamic radiance field with few input viewpoints is highly under-constrained. A popular approach to regularize an under-constrained system is to impose additional priors during the optimization. In this chapter, we explore the use of motion priors to regularize the dynamic radiance field. However, K-Planes employs a 4D volumetric representation without a motion model, it does not allow the motion implicitly learned by the model to be regularized using motion priors. Thus, there is a need to design a dynamic radiance field with an explicit motion field that lends itself to be constrained with motion supervision. Further, we desire the radiance field to be compact and allow faster optimization and rendering.

Fast optimization of radiance fields allows wider usage in real-world applications. Voxel grid based approaches trade memory for speed by replacing the MLP in NeRF with a voxel grid [54, 165]. Prior work on fast and compact representation for learning radiance fields can be broadly classified into two categories. One category of models employs a factorized volume representation [28, 33, 53] to exploit the spatial correlation of the entities to be learned, such as volume density and color. The other category of models [18, 81, 125] mainly exploit the sparsity of the scene to reduce the memory footprint. However, the scene flow for a moving object is non-zero at every time instant, but its variation is temporally correlated. Since the motion field is not sparse temporally, we believe that the sparsification based approaches may tend to learn an independent motion field for every time instant. For example, with a 4DGS model, a single 4D Gaussian may not be able to model the motion across a few time instants and can model the motion at a single time instant only, similar to Lee et al. [92]. On the other hand, factorized volumes can effectively exploit the spatio-temporal correlation of the motion field and hence we employ a 4D factorized model to learn the motion field. We then supervise the motion learned by the motion model using reliable sparse flow priors.

Table 5.1: **Related work overview:** We compare our work with prior works based on various aspects. Sparse input views refers to models that can handle data from few stationary multi-view cameras. Explicit refers to models that primarily employ explicit models followed by an optional tiny MLP.
[†] TiNeuVox and SWAGS use an explicit model to represent the scene, but use implicit neural networks to model the motion. [*] NSFF predicts motion as an auxiliary task.

| | Dynamic Scenes | Sparse Views | Explicit Model | Motion Model |
|---|:---:|:---:|:---:|:---:|
| NeRF | ✗ | ✗ | ✗ | – |
| TensoRF, i-ngp, 3DGS | ✗ | ✗ | ✓ | – |
| DS-NeRF, SimpleNeRF | ✗ | ✓ | ✗ | – |
| NSFF[*], DyNeRF | ✓ | ✗ | ✗ | ✗ |
| D-NeRF | ✓ | ✗ | ✗ | ✓ |
| TiNeuVox, SWAGS | ✓ | ✗ | ✗[†] | ✓ |
| K-Planes, HexPlane | ✓ | ✗ | ✓ | ✗ |
| Ours | ✓ | ✓ | ✓ | ✓ |

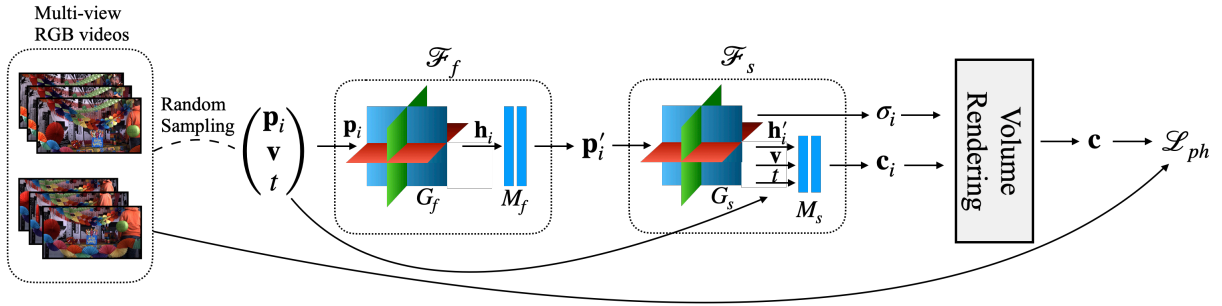

Figure 5.1: **Model architecture:** We decompose the dynamic radiance field into a 4D scene flow or deformation field $\mathcal{F}_f$ that maps a 3D point $\mathbf{p}_i$ at time $t$ to the corresponding 3D point $\mathbf{p}'_i$ at canonical time $t'$, and a 5D radiance field $\mathcal{F}_s$ that models the scene at canonical time $t'$. Both the fields are modeled using a factorized volume followed by a tiny MLP, which allows fast optimization and rendering. We note that $\mathbf{G}_f$ is modeled using six planes, although we show only three owing to the difficulty in visualizing four dimensions. The MLP $\mathbf{M}_s$ is conditioned on time and viewing direction to model time-dependent color variations such as shadows and view-dependent color variations such as specularities. The output of $\mathcal{F}_s$ is volume rendered to obtain the color of the pixel and the photometric loss is used to train both the fields. The explicitly modeled motion field $\mathcal{F}_f$ is additionally regularized using the flow priors as shown in Fig. 5.3.

## 5.2   Method

Given a set of posed video frames $I_t^v$, where $t = \{1, 2, \ldots, N_f\}$ denotes the time instant or the frame index among $N_f$ frames per camera and $v = \{1, 2, \ldots N_c\}$ denotes the camera index among $N_c$ stationary cameras, the goal is to synthesize the dynamic scene at a specified time instant $t = \{1, 2, \ldots, N_f\}$ and in any novel view. We focus on the sparse multi-view setting, where the number of cameras $N_c$ is small. The main challenges here include the design of an explicit motion model that allows regularization with motion priors and the choice of reliable motion priors itself when training with few input views. We first describe the design of our dynamic radiance field (Sec. 5.2.1), and then discuss its training with reliable flow priors (Sec. 5.2.2).

### 5.2.1   Dynamic Radiance Field

We modify the K-Planes [53] implementation for dynamic radiance fields to incorporate an explicit motion model. K-Planes employs a multi-resolution grid $\mathbf{G}$ followed by a tiny multi-layer perceptron (MLP) $\mathbf{M}$ to represent a radiance field. As shown in Fig. 5.1, we model our dynamic radiance field by employing two factorized tensorial models, a radiance field $\mathcal{F}_s = \mathbf{G}_s \circ \mathbf{M}_s$ that represents the 3D scene at a canonical time instant $t'$, and a 4D scene flow or deformation field $\mathcal{F}_f = \mathbf{G}_f \circ \mathbf{M}_f$ that represents the scene flow for a 3D point $\mathbf{p}$ from time $t$ to $t'$. Mapping the 3D scene at every time instant to a canonical volume helps our model enforce temporal consistency of objects in the scene [138].

Here, $\mathbf{G}_f$ consists of six planes $\{\mathbf{S}_{xy}, \mathbf{S}_{yz}, \mathbf{S}_{xz}, \mathbf{S}_{xt}, \mathbf{S}_{yt}, \mathbf{S}_{zt}\}$ at every resolution, where the first three planes model the spatial correlation and the next three model the spatio-temporal correlation of the motion field. To obtain the scene flow for a 3D point $\mathbf{p}$ from time $t$ to $t'$, we first project the 4D point $(\mathbf{p}, t)$ onto the six planes and bilinearly interpolate the feature vectors in each of the six planes. We then combine the features using the Hadamard product to obtain the final feature vector $\mathbf{h}_i$ as

$$\mathbf{h}_i = \mathbf{G}_f(\mathbf{p}) = \prod_{c \in \{xy, yz, xz, xt, yt, zt\}} S_c(\mathbf{p}), \qquad (5.1)$$

which is then fed to the tiny MLP $\mathbf{M}_f$ that outputs the scene flow. Modeling the scene flow field $\mathcal{F}_f$ using the hex-plane representation and the tiny MLP makes our motion model fast and compact. For more details on the hex-plane representation, we refer the readers to K-Planes [53]. We model the scene grid $\mathbf{G}_s$ using a similar factorization with three spatial planes. While it is beneficial to model the motion field using a factorized volume, the canonical scene can be learned using any explicit representation such as 3DGS [81]. We employ factorized volumes to model both scene and motion using a unified framework.

We train our model, similar to K-Planes, by rendering randomly sampled pixels $\mathbf{q}_t^v$ in view $v$ at time $t$ and using the ground truth color as supervision. To render a pixel $\mathbf{q}_t^v$, we sample $N_p$ points $\{\mathbf{p}_i\}_{i=1}^{N_p}$ at depths $\{z_i\}_{i=1}^{N_p}$ along the corresponding ray. For every 3D point $\mathbf{p}_i$, we first obtain the corresponding 3D point $\mathbf{p}_i'$ at canonical time $t'$ by computing the scene flow from time $t$ to $t'$ using $\mathcal{F}_f$ as

$$\mathbf{p}_i' = \mathcal{F}_f\left(\mathbf{p}_i, t\right) + \mathbf{p}_i. \tag{5.2}$$

We then query $\mathbf{G}_s$ at $\mathbf{p}_i'$ to obtain the volume density $\sigma_i$ and a latent feature $\mathbf{h}_i'$ corresponding to $\mathbf{p}_i$ as

$$\sigma_i, \mathbf{h}_i' = \mathbf{G}_s\left(\mathbf{p}_i'\right). \tag{5.3}$$

A tiny MLP $\mathbf{M}_s$ maps $\mathbf{h}_i'$, encoded viewing direction $\mathbf{v}$ and encoded time $t$ to the color $\mathbf{c}_i$ of $\mathbf{p}_i$ as

$$\mathbf{c}_i = \mathbf{M}_s\left(\mathbf{h}_i', \gamma\left(\mathbf{v}\right), \gamma\left(t\right)\right), \tag{5.4}$$

where $\gamma$ denotes the encoding [125] of the viewing direction and the time instant. Conditioning $\mathbf{M}_s$ additionally on time allows our model to capture illumination changes due to object motion. $\mathbf{c}_i$ are then volume rendered to obtain the color $\mathbf{c}$ of $\mathbf{q}$ as

$$w_i = \left(\Pi_{j=1}^{i-1} \exp\left(-\delta_j \sigma_j\right)\right) \left(1 - \exp\left(-\delta_i \sigma_i\right)\right), \tag{5.5}$$

$$\mathbf{c} = \sum_{i=1}^{N_p} w_i \mathbf{c}_i, \quad z = \sum_{i=1}^{N_p} w_i z_i, \tag{5.6}$$

where $\delta_i = z_i - z_{i-1}$ and $z$ gives the expected depth of $\mathbf{q}_t^v$. Both $\mathcal{F}_f$ and $\mathcal{F}_s$ are optimized through the photometric loss, $\mathcal{L}_{\mathrm{ph}} = \|\mathbf{c} - \hat{\mathbf{c}}\|^2$, where $\hat{\mathbf{c}}$ is the ground truth color. Decomposing the scene into a canonical scene field and a flow field allows us to regularize the flow field using flow priors when only a sparse set of viewpoints are available.

### 5.2.2 Training with Flow Priors

A popular choice for motion priors is the dense optical flow estimated using deep optical flow networks [170]. However, we find that the dense optical flow estimates are unreliable due to generalization issues as seen in Fig. 5.2a. On the other hand, matching keypoints using robust SIFT [111] descriptors are more reliable across cameras as seen in Fig. 5.2b. Specifically, for a pair of frames across time instants $t$ and $s$ and cameras $v$ and $u$, we extract the SIFT keypoints in individual images using Colmap [145] and match the keypoints using SIFT descriptors. The difference between the locations of the matched keypoints gives the sparse flow prior $\mathcal{P}_{\mathrm{sf}}$. While Colmap provides a reasonable number of matches, newer models such as R2D2 [141] could be employed to improve the richness of the sparse flow prior.

Employing flow priors from any time $t$ to canonical time $t'$ for large $t - t'$ may be less efficient since large regions of the scene may not be visible. Thus, it is desirable to utilize flow priors across a short duration of time. However, the motion model $\mathcal{F}_f$ provides the scene flow from $t$ to $t'$ only and not the flow between any two arbitrary time instants. We resolve this by using the flow prior to encourage $\mathcal{F}_f$ to map a pair of matched points at different time instants to the same object location in 3D. Specifically, we obtain the matching pixels using the flow priors and constrain the 3D points corresponding to the

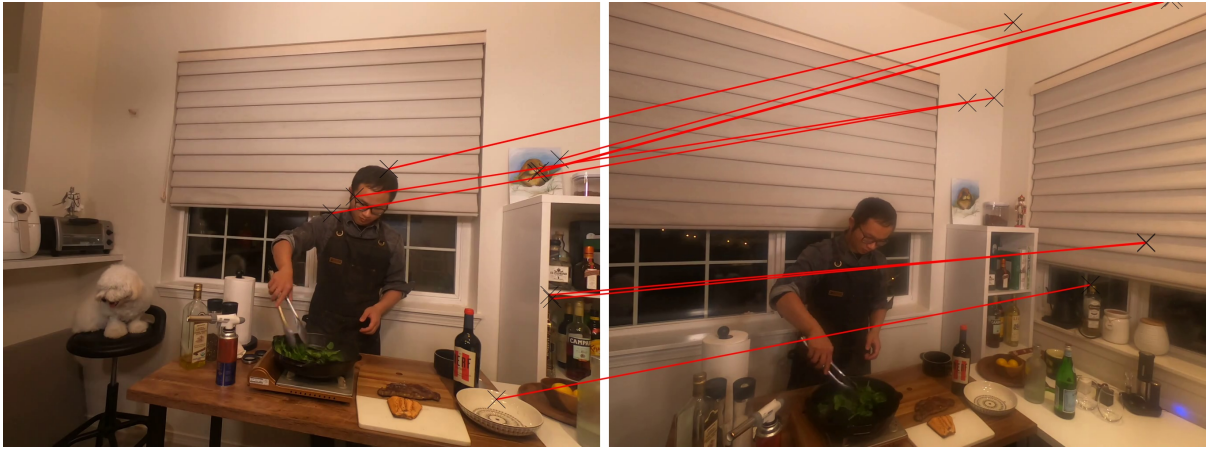

(a) Dense flow across cameras estimated by RAFT.

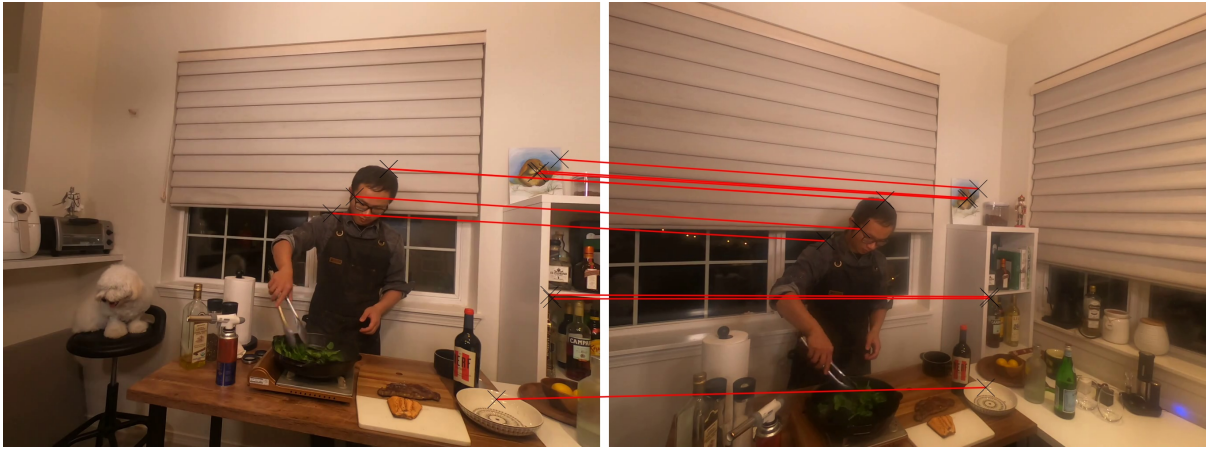

(b) Sparse flow across cameras estimated by SIFT.

Figure 5.2: **Visualization of different flow priors:** We show the matched pixels as provided by different flow priors. The pixels in the first view are randomly picked from those for which sparse flow is available and the same pixels are used for dense flow. Note that the second view in the first two examples has more blur as compared to the first view. (a) We show that the dense flow priors across cameras obtained using deep optical flow networks such as RAFT [169] are prone to erroneous matches, due to variations in camera parameters and lighting. We observed similar trends with other deep optical flow networks as well such as AR-Flow [105]. (b) Matching pixels across cameras using robust SIFT features provides reliable matches, albeit sparse.

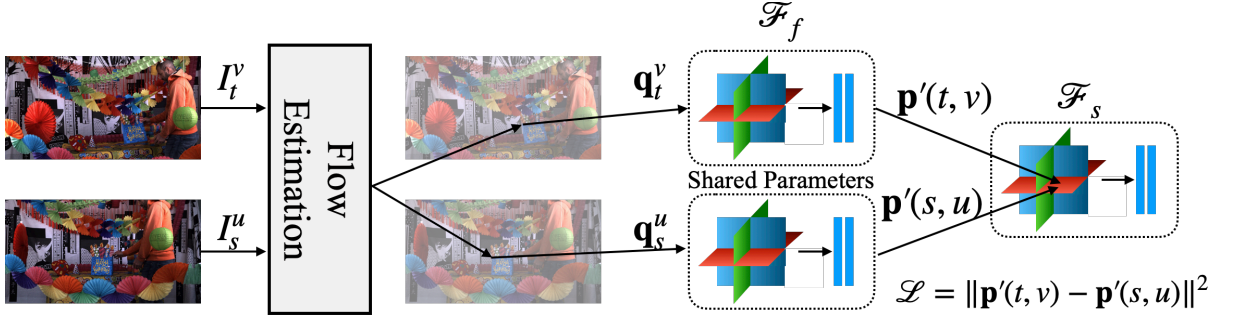

**Figure 5.3: Flow regularization:** Since the motion field $\mathcal{F}_f$ gives only the unidirectional flow from time $t$ to $t'$, we impose the flow prior by minimizing the distance between the 3D points in the canonical volume corresponding to the matched pixels $(\mathbf{q}_t^v, \mathbf{q}_s^u)$ in the input frames $(I_t^v, I_s^u)$.

matching pixels to map to the same 3D point in the canonical scene field as shown in Fig. 5.3.

Mathematically, let the flow corresponding to a pixel $\mathbf{q}_t^v$ to time $s$ and view $u$ be given by $\mathbf{f}_{t \to s}^{v \to u}$. Then the matching pixel at time $s$ and view $u$ is given by $\mathbf{q}_s^u = \mathbf{q}_t^v + \mathbf{f}_{t \to s}^{v \to u}$. Let the 3D points sampled along the rays corresponding to $\mathbf{q}_t^v$ and $\mathbf{q}_s^u$ be given by $\mathbf{p}_i(t,v)$ and $\mathbf{p}_i(s,u)$ respectively. Then, we impose the sparse flow constraint as

$$\mathcal{L}_{\mathrm{sf}} = \left\| \sum_{i=0}^{N_p-1} w_i(t,v)\mathbf{p}_i'(t,v) - \sum_{i=0}^{N_p-1} w_i(s,u)\mathbf{p}_i'(s,u) \right\|^2, \quad (5.7)$$

where $\mathbf{p}'(t,v)$ and $\mathbf{p}'(s,u)$ are computed from $\mathbf{p}(t,v)$ and $\mathbf{p}(s,u)$ using $\mathcal{F}_f$ as in Eq. (5.2), and $w_i(t,v)$ and $w_i(s,u)$ are computed using Eq. (5.5). The two terms in Eq. (5.7) represent $\mathbf{q}_t^v$ and $\mathbf{q}_s^u$ in the canonical volume respectively. Thus, $\mathcal{L}_{\mathrm{sf}}$ guides $\mathcal{F}_s$ on which two points in two different time instants belong to the same object.

In Eq. (5.7), we first find the canonical field points $\mathbf{p}_i'$ using $\mathcal{F}_f$ and then average the location of the 3D points using weights $w_i$ and not the other way for the following reason. The former approach regularizes the motion model $\mathcal{F}_f$ for every $\mathbf{p}_i$ giving a rich supervision to $\mathcal{F}_s$, whereas the latter approach regularizes $\mathcal{F}_f$ only for the expected 3D point $\mathbf{p} = \sum w_i \mathbf{p}_i$. Further, since we do not impose stop-gradient on $w_i$ in $\mathcal{L}_{\mathrm{sf}}$, the flow priors also help remove incorrect masses such as floaters.

### 5.2.3   Overall Model

We train our model by minimizing the combination of photometric loss $\mathcal{L}_{\mathrm{ph}}$ and sparse flow loss $\mathcal{L}_{\mathrm{sf}}$ as

$$\mathcal{L} = \mathcal{L}_{\mathrm{ph}} + \lambda_{\mathrm{sf}}\mathcal{L}_{\mathrm{sf}}, \tag{5.8}$$

where $\lambda_{\mathrm{sf}}$ is a hyper-parameter.

## 5.3   Experiments

### 5.3.1   Evaluation Setup

**Datasets:**  We evaluate our model on two popular multi-view dynamic scene datasets, namely N3DV [98] and InterDigital [143] with three input views. Following prior work [53], we downsample the videos spatially by a factor of two for all the experiments. We use the video at the center of the camera rig for testing and uniformly sample train videos from the remaining videos. *N3DV dataset* contains six real-world scenes with 17–21 static cameras per scene and 300 frames per viewpoint. The videos have a spatial resolution of $1352 \times 1014$ and a frame rate of 30fps. *InterDigital dataset* contains multiple real-world scenes with 16 static cameras and varying number of frames per scene. We undistort the video frames using the radial distortion parameters provided with the dataset and use the undistorted videos for our experiments. We select five scenes that contain at least 300 frames and choose the first 300 frames. The videos have a spatial resolution of $1024 \times 544$ and a frame rate of 30fps spanning 10 seconds in all our experiments.

**Evaluation measures:**  We evaluate the rendered frames of all the methods using PSNR, SSIM [191] and LPIPS [215]. We also evaluate the models on their ability to reconstruct the 3D scene by computing MAE on the rendered depth maps. Due to the unavailability of true depth maps on both datasets, we use the depth provided by K-Planes trained with dense input views as pseudo ground truth depth.

Table 5.2: **Quantitative results:** We compare our model with K-Planes on N3DV [98] and InterDigital [143] datasets with three input views. We also show the performance of our base DeRF model for reference. We report the PSNR, SSIM, and LPIPS scores for the rendered images and the depth MAE for the rendered depth maps. The best scores in each category are shown in bold.

| Model | N3DV | | | | InterDigital | | | |
|---|---|---|---|---|---|---|---|---|
| | PSNR ↑ | SSIM ↑ | LPIPS ↓ | Depth MAE ↓ | PSNR ↑ | SSIM ↑ | LPIPS ↓ | Depth MAE ↓ |
| HexPlane | 15.53 | 0.49 | 0.50 | 1.97 | 13.85 | 0.27 | 0.54 | 1.43 |
| K-Planes | 23.65 | 0.83 | 0.25 | 0.34 | 18.75 | 0.66 | 0.30 | 0.22 |
| DeRF | 22.61 | 0.81 | 0.27 | 0.39 | 18.72 | 0.68 | 0.30 | 0.22 |
| SF-DeRF | **24.79** | **0.87** | **0.22** | **0.20** | **20.06** | **0.73** | **0.26** | **0.13** |

## 5.3.2 Comparisons and Implementation Details

We mainly compare the performance of our model against K-Planes and HexPlane. We use the official code released by the authors of K-Planes and modify it to implement our model. We refer to our base model without any priors as DeRF model. To test the superiority of our sparse flow prior, we also compare our model against DeRF with dense flow priors. Since we use two factorized models instead of one used in K-Planes, we reduce the feature dimension in both of our models by half to keep the total number of parameters comparable to K-Planes. We train all the models for 30k iterations on a single NVIDIA RTX 2080 GPU. We randomly pick $s \in \{t - 10, t + 10\}$ to impose the flow prior losses and set $\lambda_{sf} = 1$. We use the same values as suggested by K-Planes for all the remaining hyperparameters.

K-Planes            DeRF            SF-DeRF        Ground Truth

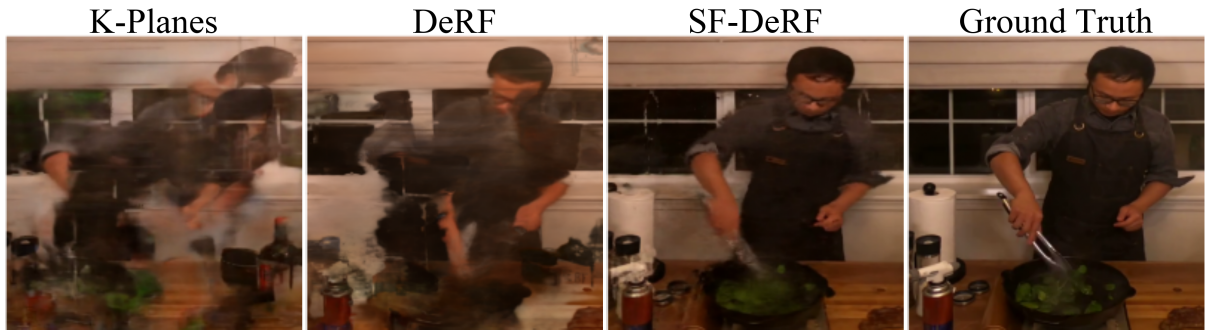

Figure 5.4: **Qualitative examples on N3DV dataset with three input views:** We can observe that K-Planes finds it hard to learn the moving person leading to significant distortions. Our DeRF model (without any priors) corrects a few errors by virtue of the common canonical volume. Imposing our priors leads to much better reconstruction.

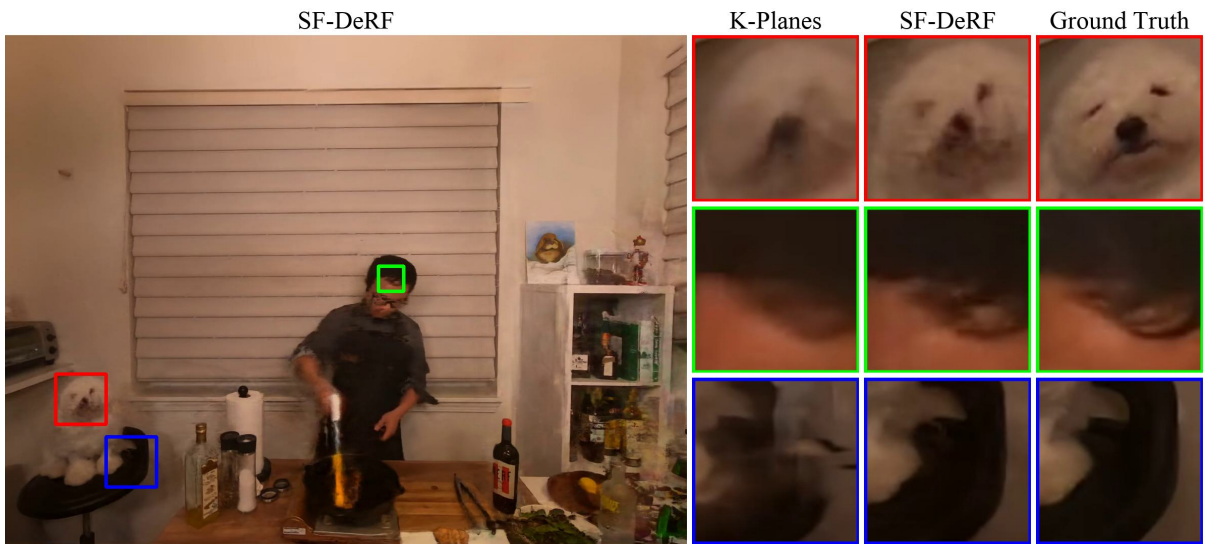

Figure 5.5: **Qualitative examples on N3DV dataset with three input views:** We observe that K-Planes blurs moving objects such as the face of the dog and the hairs on the face of the person in the first two examples respectively. We also find that K-Planes creates distortions in the static regions when an object moves in its vicinity as seen in the third row. However, our SF-DeRF model produces sharper and more accurate results.

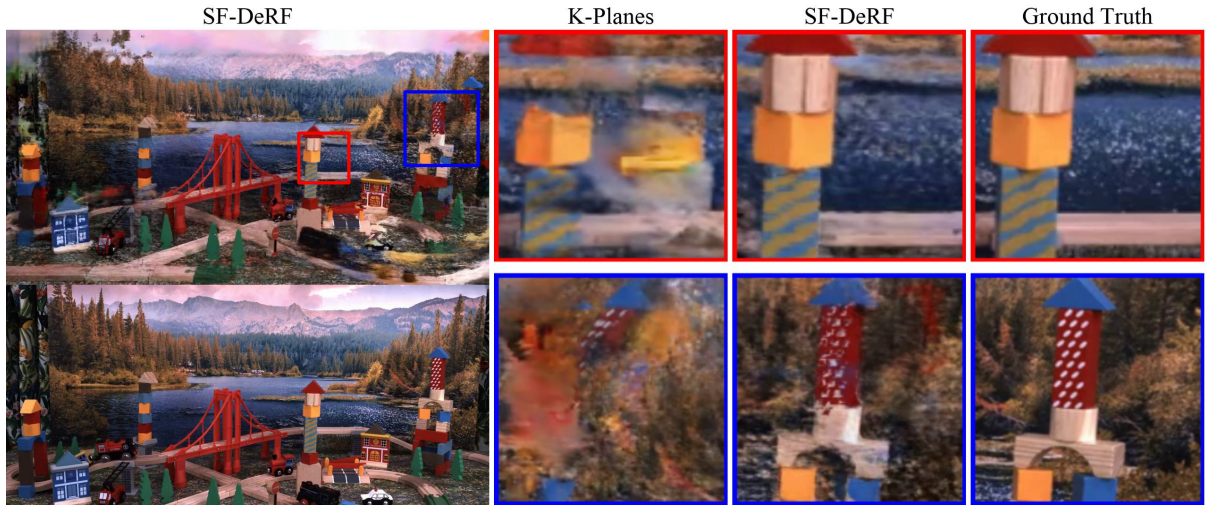

Figure 5.6: **Qualitative examples on InterDigital dataset:** In the first row, K-Planes creates a duplication of the tower, while the tower is significantly distorted in the second row. Our model correctly reconstructs both the towers.

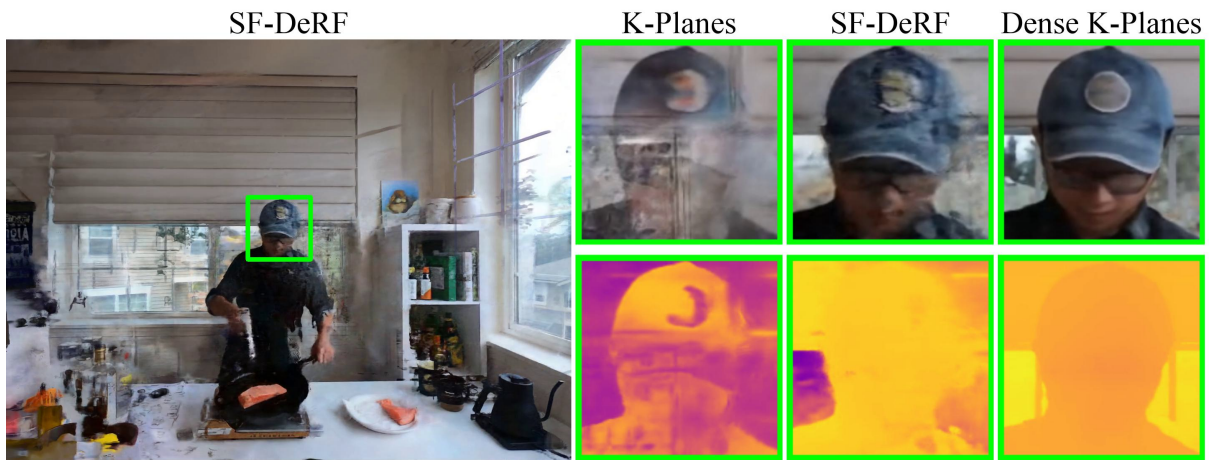

Figure 5.7: **Visualization of rendered depth on N3DV dataset:** Observe the difference in color of the depth map rendered by K-Planes which shows the errors in the estimated depth.

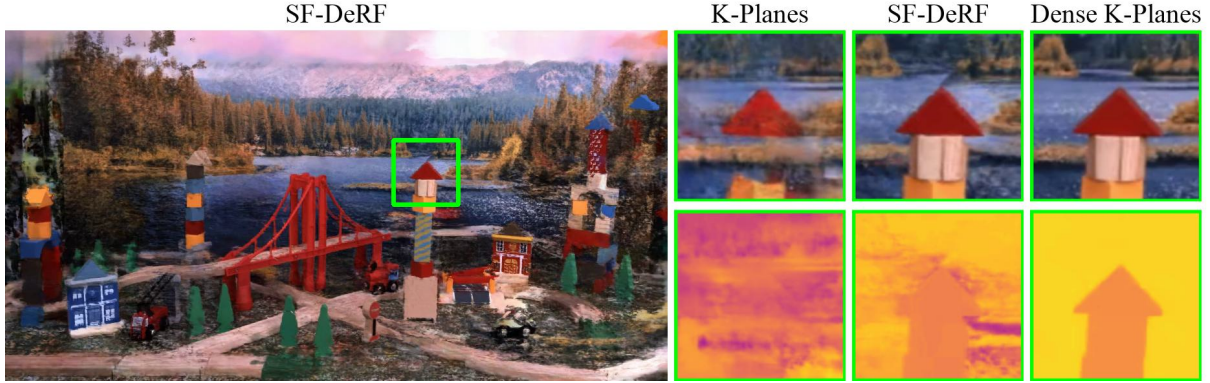

Figure 5.8: **Visualization of rendered depth on InterDigital dataset:** While our model learns better depth in the scene leading to better reconstruction of the tower, K-Planes is unable to learn the geometry correctly causing distortions in the tower.

### 5.3.3   Results

We show the quantitative performance of our model, K-Planes and HexPlane in Tab. 5.2, where we observe that our SF-DeRF model outperforms both K-Planes and HexPlane across all the settings on both the datasets. We observe that the performance of all the models is relatively higher on the N3DV dataset as compared to the InterDigital dataset. This is perhaps due to the InterDigital dataset having larger motion and highly textured regions. We also note that the performance of HexPlane is substantially lower than both K-planes and our model. This could be a result of optimizing HexPlane initially with a low-resolution grid, which causes the model to overfit the input-views and not utilize the high-resolution grid in the later stages of training. K-Planes does not suffer from this drawback by using a multi-resolution grid throughout the optimization. From Figs. 5.5 and 5.6, we observe that our model is able to correct errors over K-Planes. The improvements are more starkly visible in the supplementary videos on our project webpage[1].

We compare the rendered depth from both the models in Figs. 5.7 and 5.8 and observe that our model is able to reconstruct the depth more accurately than K-Planes. Tab. 5.2 also shows that our DeRF model (without any priors) performs better than K-Planes in the scenes with larger motion, such as the scenes in the InterDigital dataset, perhaps due to the temporal consistency enforced by the canonical volume. This is also visible in the example shown in Fig. 5.4. In our experiments, SF-DeRF roughly took 1.5 hours and 5GB GPU memory to train on a single scene, and 6 seconds to render a single frame. The size of the saved model parameters is approximately 280MB. Please refer to Fridovich-Keil et al. [53] for the training time for various models including K-Planes and DyNeRF.

---

[1]https://nagabhushansn95.github.io/publications/2024/RF-DeRF.html

Table 5.3: We test the performance of our DeRF model with dense flow priors and our reliable sparse flow priors. The performance of the DeRF model (without any priors) is also shown for reference.

| Model | N3DV | | InterDigital | |
|---|---|---|---|---|
| | LPIPS ↓ | Depth MAE ↓ | LPIPS ↓ | Depth MAE ↓ |
| K-Planes | 0.25 | 0.34 | 0.30 | 0.22 |
| DeRF w/o any priors | 0.27 | 0.39 | 0.30 | 0.22 |
| DeRF w/ dense flow priors | 0.30 | 0.49 | 0.36 | 0.29 |
| DeRF w/ our sparse flow priors | **0.22** | **0.20** | **0.26** | **0.13** |

**Ablations:** We analyze the significance of our sparse flow prior by disabling it in Tab. 5.3. We observe that the sparse flow prior gives a large boost in performance when included and a large drop in performance when excluded. The effect of sparse flow prior is more pronounced in terms of depth MAE, which measures the accuracy of the reconstructed 3D scene. This demonstrates the importance of reliable motion priors, even if sparse.

**Dense flow priors vs our sparse flow priors:** To validate our hypothesis that sparse flow priors are more effective than dense flow priors, we impose dense flow priors on our base DeRF model and evaluate the performance on both datasets. From Tab. 5.3 and Fig. 5.10, we observe that imposing noisy dense flow priors leads to a large drop in performance, whereas our priors help improve the performance significantly.

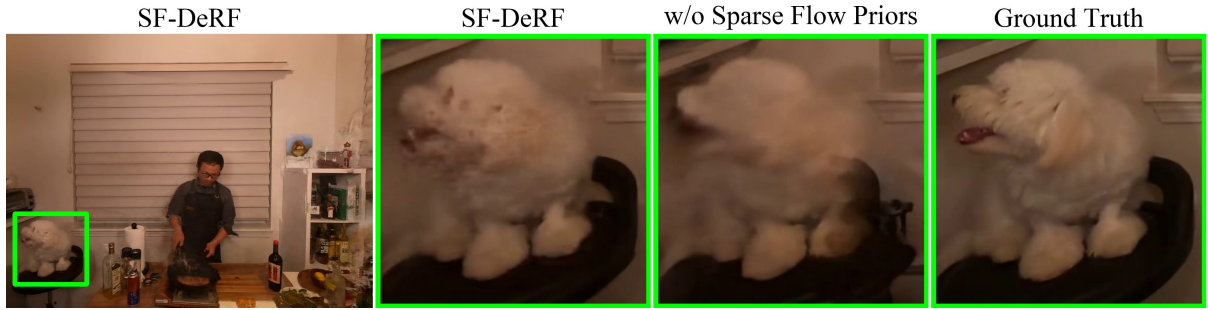

Figure 5.9: **Qualitative examples of ablated models on N3DV dataset:** Sparse flow prior is effective in regularizing moving regions in the scene. Without sparse flow priors, we observe that the face of the dog suffers from motion blur creating a fuzzy white mass.

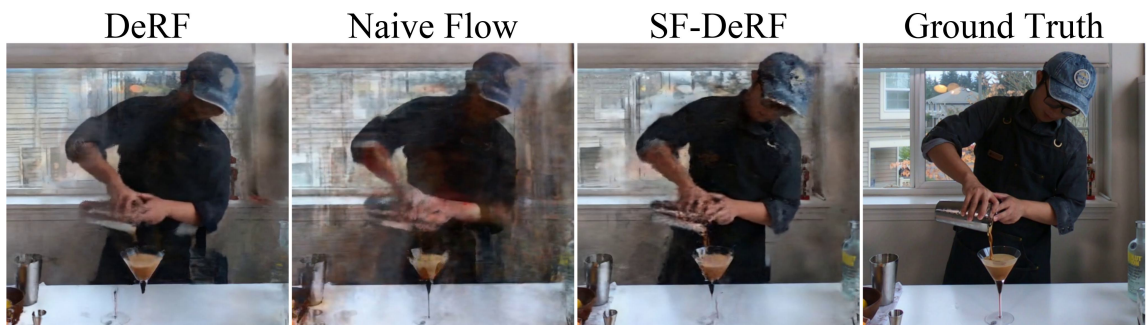

Figure 5.10: **Qualitative examples to show the effect of dense flow priors:** We observe that the hand of the person in the second column is distorted, perhaps due to incorrect priors provided by the dense flow. Further, we observe that the dense flow priors deteriorate the performance as compared to our base DeRF model. Our priors do not cause such distortions, while improving the overall reconstruction quality (observe the color of the shirt).

Table 5.4: We test the performance of our DeRF model against K-Planes with dense input views.

| Model | N3DV | | | InterDigital | | |
|-------|--------|--------|---------|--------|--------|---------|
|       | PSNR ↑ | SSIM ↑ | LPIPS ↓ | PSNR ↑ | SSIM ↑ | LPIPS ↓ |
| K-Planes | 30.55 | 0.96 | 0.12 | 29.00 | 0.96 | 0.07 |
| DeRF | 29.95 | 0.95 | 0.12 | 28.14 | 0.95 | 0.09 |

**Performance with dense input views:** We now test the ability of our model in reconstructing the dynamic scene when provided with dense multi-view inputs. Since motion priors may not be needed when dense views are available, we analyze the performance of our base DeRF model (without any priors). From Tab. 5.4, we observe that our DeRF model is competitive with K-Planes. Thus, one could employ our model in both dense and sparse input scenarios by additionally employing reliable flow priors in the latter case.

## 5.3.4  Limitations

Our approach of mapping the scene at every time instant to the canonical volume implies that only the objects present in the canonical volume can be rendered. In other words, our model may not handle objects entering or leaving the scene, which could happen over longer durations. This could be resolved by learning multiple models, each trained on self-contained shorter duration videos. We employ Colmap to generate sparse flow priors in our framework, which is a time-consuming process. In our experiments, we found that the generation of sparse flow priors takes about 45 minutes per scene. There is a need to explore faster approaches to determine reliable sparse correspondences.

## 5.4   Summary

We consider the setting of fast dynamic view synthesis when only a few videos of the scene as observed from different static cameras are available. By exploiting the spatio-temporal correlation of the motion field, we design an explicit motion model using factorized representations that is compact, fast, and allows effective regularization with the flow priors. We observe that the use of existing dense flow priors has a negative effect on the performance, while the use of reliable sparse flow priors provides a significant boost in performance. We demonstrate the effectiveness of our approach on two popular datasets and show that our approach outperforms the state-of-the-art fast and compact dynamic radiance fields by a large margin when only a few viewpoints are available.

# Chapter 6

# Temporal View Synthesis of Dynamic Scenes through 3D Object Motion Estimation with Multi-Plane Images

## 6.1 Introduction

Recall that Temporal View Synthesis of Dynamic Scenes (TVS-DS) involves causal frame-rate upsampling of graphically rendered videos as shown in Fig. 6.1 through temporal view synthesis. Concretely, given the past frames and their camera poses along with the camera pose of a future frame, TVS-DS aims at synthesizing the future frame. The main difference between TVS-DS and novel view synthesis of dynamic scenes [40, 114, 193, 217] is the object motion prediction between source and target views. For example, in the works by Gao et al. [56], Lin et al. [101] and Yoon et al. [208], the target frame is at the same time instant as one of the source frames. Thus, they do not address the question of moving objects at a future time instant. Dynamic NeRF based models such as

---

This chapter is based on the work published at ISMAR 2022 [158].

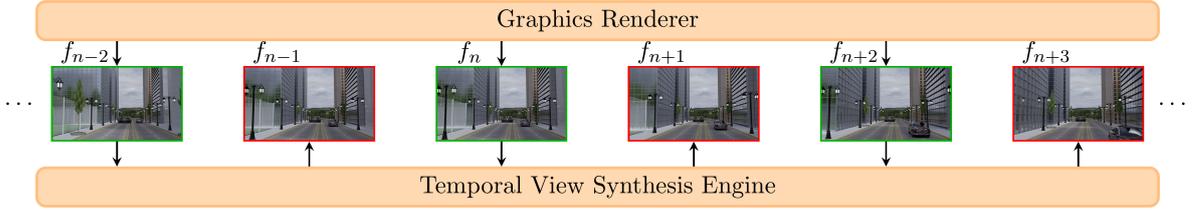

Figure 6.1: **Frame-rate upsampling of graphically rendered dynamic videos using Temporal View Synthesis.** This illustration shows upsampling by a factor of two. The graphics renderer renders alternate frames $\{f_{n-2}, f_n, f_{n+2}, \ldots\}$ and the intermediate frames $\{f_{n-1}, f_{n+1}, f_{n+3}, \ldots\}$ are predicted using temporal view synthesis. For better visualization of motion, we show frames which are 10 time instants apart instead of consecutive frames.

HyperNeRF [132] and NSFF [99] can interpolate object motion between the frames, but require hours of training for every scene. On the other hand, our problem formulation is close to that of video prediction [117, 162], but differs in the use and availability of camera motion. The explicit use of camera motion can help TVS-DS methods perform much better than generic video prediction. Thus, TVS-DS lies at the intersection of video prediction and view synthesis.

In this work, we propose a novel framework for TVS-DS by first isolating the object motion in past frames, estimate the object motion as flow in 3D using multiplane image (MPIs) representation, and then extrapolate it to predict the future object motion. We then incorporate the future camera motion by warping the MPI, infill the disocclusions in the MPI representation and render the MPI to obtain the final frame.

## 6.2   Problem Statement

We formulate the problem of temporal view synthesis of dynamic scenes for causal frame rate upsampling of synthetic videos. Consider the scenario of upsampling by $k$ times, where we predict $k-1$ future frames before the next rendered frame. Given previous frames $\{f_n, f_{n-k}, \ldots, f_{n-lk}\}$, their depth maps $\{d_n, d_{n-k}, \ldots, d_{n-lk}\}$, camera poses (extrinsics) $\{T_n, T_{n-k}, \ldots, T_{n-lk}\}$, camera intrinsics $K$ and the camera poses of the next frames $\{T_{n+1}, T_{n+2}, \ldots, T_{n+k-1}\}$, we seek to predict the next frames $\{f_{n+1}, f_{n+2}, \ldots, f_{n+k-1}\}$. We assume that the motion in the video is caused by both camera and object motion. We refer to the motion due to user or camera movement as global motion and that of objects as local motion.

Although the camera motion is available and large parts of the frame to be predicted can be generated by warping the previous frame to the desired view, the movement of objects creates additional challenges. An off-the-shelf application of video prediction algorithms can be inefficient since these algorithms do not effectively use the camera motion and the scene depth. Thus, the key challenge in predicting the next frame is to design a framework that can predict the motion of individual objects and utilize the available camera motion. We assume that the ground truth depth maps are available for the rendered frames since we focus on graphical rendering applications in this work. We also assume that illumination changes are minimal due to the high frame rates of the videos.

## 6.3 Method

### 6.3.1 Multi-Plane Images (MPI)

Before delving into the details of our model, we briefly discuss the MPI representation and its generation. The MPI representation introduced by Zhou et al. [216] expands a 2D RGB frame into a set of RGBA image planes, located at different depths. The alpha channel ($\alpha \in [0,1]$) in each plane denotes occupancy of the scene at the corresponding depth. Utilizing the knowledge of depth, we create the MPI directly from the RGB-D image instead of estimating the MPI as is common in literature [96, 172, 216]. For the given RGB-D image, we first sample $Z$ planes uniformly in inverse depth between the minimum and maximum depth of the scene. For every location $\mathbf{x}$, we set $\alpha = 1$ at the plane nearest to the true depth of $\mathbf{x}$ and set $\alpha = 0$ for the rest of the planes. Thus at each location $\mathbf{x}$, the $\alpha$ values across all the planes form a one-hot vector. We modify the MPI representation to contain true depth values in an additional channel along with RGBA. We denote the MPI representation of $f_n$ as $m_n = \{c_n, d_n, \alpha_n\}$, where $c_n$, $d_n$ and $\alpha_n$ are the RGB, depth and alpha channels respectively. To warp an MPI to a different camera view, we employ reprojection and bilinear splatting [79, 173] instead of inverse homography employed by Zhou et al. [216]. Finally, to render a 2D frame from an MPI, we use alpha compositing in back to front order using the standard over operation [216].

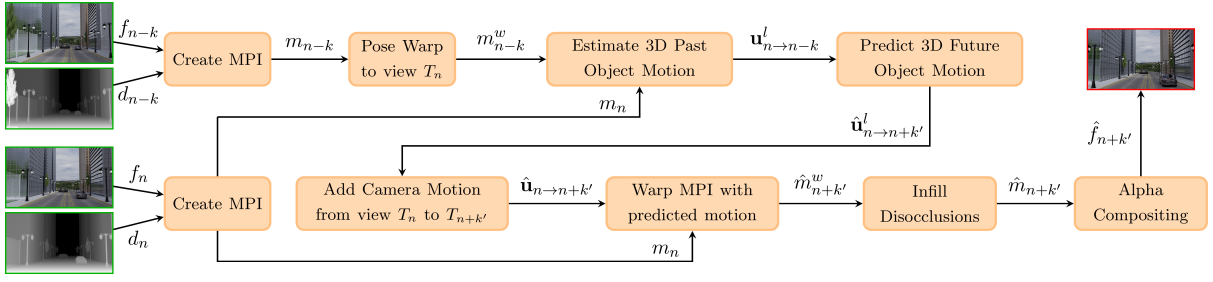

Figure 6.2: Overall architecture of DeCOMPnet. The given past frames are first converted to MPI and warped to the same camera view. 3D object motion is estimated between the warped MPIs and extrapolated to predict the future object motion. Future camera motion is incorporated to predict the total future motion, which is used to warp the MPI of $f_n$. The warped MPI is then infilled and alpha composited to obtain the predicted future frame. For better visualization, inverse depth maps are shown.

## 6.3.2   Overview of the Proposed Approach

We present our approach for predicting future frames of dynamic scenes using camera motion knowledge as follows. For a scene with moving objects captured by a moving camera, to explicitly use the available camera motion in predicting the next frame, we adopt the following two-step approach. We first hold the camera still and account for the object motion. We then keep objects still and account for the camera motion alone. We use MPIs to represent the 3D scenes. Let $k' \in \{1, 2, \ldots, k-1\}$ denote the prediction timestep and $\hat{\mathbf{u}}_{n \to n+k'}^l(\mathbf{x}, z)$ be the local optical flow in plane $z \in \{1, 2, \ldots, Z\}$ at location $\mathbf{x}$ that describes the motion of the pixel from time instant $n$ to $n + k'$ in view $T_n$. Let $P_{n \to n+k'}$ be the pose-warping operator from view $T_n$ to $T_{n+k'}$ that includes both the local and global motion in the warping. Corresponding to the location $(\mathbf{x}, z)$ in $m_n$ such that $\alpha_n(\mathbf{x}, z) = 1$, we obtain the MPI $\hat{m}_{n+k'}^w$ of a future frame $f_{n+k'}$ as

$$\hat{m}_{n+k'}^w(P_{n \to n+k'}(\mathbf{x}, \hat{\mathbf{u}}_{n \to n+k'}^l(\mathbf{x}, z), d_n(\mathbf{x}, z))) = m_n(\mathbf{x}, z), \tag{6.1}$$

where the pose-warping operator $P_{n \to n+k'}$ is defined as

$$P_{n \to n+k'}(\mathbf{x}, \mathbf{u}, d) = K T_{n+k'} T_n^{-1} (d + \mathbf{u}_z) K^{-1} (\mathbf{x} + \mathbf{u}_{xy}), \tag{6.2}$$

where $\mathbf{u}_{xy}$ and $\mathbf{u}_z$ denote the components of flow in the x-y plane and in the depth dimension respectively. Since Eq. (6.2) represents forward warping, to obtain the intensities at integer locations of $\hat{m}_{n+k'}^w$, we use splatting similar to [79, 173]. Along the depth dimension, we simply select the nearest plane. We omit the conversion between non-homogeneous and homogeneous coordinates for notation simplicity.

In Eq. (6.1), while the camera motion $P_{n \to n+k'}$ is known, the object motion $\hat{\mathbf{u}}_{n \to n+k'}^l$ is unknown and needs to be predicted. While warping $m_n$ to get $\hat{m}_{n+k'}^w$ using Eq. (6.1), multiple locations from $m_n$ can map to a same location but different depth planes in $\hat{m}_{n+k'}^w$. Thus, for a few locations in $\hat{m}_{n+k'}^w$ across all the planes, there may be no matching points in $m_n$. Rendering such an MPI using alpha-compositing creates disocclusions or holes. Hence, we infill the warped MPI $\hat{m}_{n+k'}^w$ to get $\hat{m}_{n+k'}$ before rendering the frame $\hat{f}_{n+k'}$. We summarize our approach in Fig. 6.2. In the following subsections, we present the main challenges and our contributions in local motion prediction and briefly discuss our disocclusion infilling module.

### 6.3.3   Local 3D Object Motion Prediction

We predict the 3D object motion $\hat{\mathbf{u}}_{n \to n+k'}^l$ in view $T_n$ by estimating the local motion between $m_n$ and $m_{n-k}$ corresponding to $f_n$ and $f_{n-k}$, and extrapolating it. We only use the past ground truth frames to avoid the accumulation of errors. Since the motion between $f_n$ and $f_{n-k}$ is a mixture of both global and local motion, the local motion alone needs to be extracted from the overall motion. To achieve this, we first nullify the global motion between the past frames by warping $m_{n-k}$ from view $T_{n-k}$ to $T_n$ to get $m_{n-k}^w$, using Eq. (6.2) by setting $\mathbf{u} = 0$. Thus, the residual motion between $m_n$ and $m_{n-k}^w$ corresponds to the object motion between time instants $n$ and $n-k$. We estimate the 3D optical flow between $m_n$ and $m_{n-k}^w$ to compute this local motion and use it to predict $\hat{\mathbf{u}}_{n \to n+k'}^l$.

**Past 3D flow estimation:** Given the success of deep convolutional neural networks for optical flow estimation, we explore such an approach to estimate the flow between the MPI representations. We note that the flow estimation using deep neural networks is more reliable here since the frames are captured from the same camera and contain only small differences between the two. Nonetheless, we encounter two challenges while estimating object motion between two MPIs. The first is that MPI representations are inherently sparse, i.e. a significant number of pixels in MPIs have $\alpha = 0$. We handle the sparsity of MPIs by introducing 3D partial convolution layers, which convolve the input only in the regions where $\alpha = 1$. 2D partial convolutions were introduced by Liu et al. [103] to infill holes in image inpainting applications. However, we apply partial convolution in a completely different domain of estimating 3D optical flow with MPIs. In this regard, we modify the partial convolution layer to not dilate the alpha mask at every layer, since our work aims to estimate optical flow where $\alpha = 1$. Estimating optical flow typically requires computing a cost volume using a correlation layer [166]. We design masked correlation layers to handle the sparsity of MPI while computing the 3D cost volume. For input features $\mathbf{h}_1, \mathbf{h}_2$ along with corresponding alpha masks $\alpha_{h_1}, \alpha_{h_2}$, we compute the cost volume and the corresponding alpha mask as

$$\mathrm{cv}((\mathbf{x}_1, z_1), (\mathbf{x}_2, z_2)) = (\mathbf{h}_1(\mathbf{x}_1, z_1)\alpha_{h_1}(\mathbf{x}_1, z_1))^T \cdot (\mathbf{h}_2(\mathbf{x}_2, z_2)\alpha_{h_2}(\mathbf{x}_2, z_2)), \qquad (6.3)$$

$$\alpha_{\mathrm{cv}}((\mathbf{x}_1, z_1), (\mathbf{x}_2, z_2)) = \alpha_{h_1}(\mathbf{x}_1, z_1) \cdot \alpha_{h_2}(\mathbf{x}_2, z_2). \qquad (6.4)$$

The above cost volume and mask are then fed to subsequent partial convolution layers to estimate the optical flow.

The second challenge is in representing the 3D flow due to the discrete nature of depth planes in the MPI representation. We use real-valued displacements $\mathbf{a} \in \mathbb{R}^2$ in the x-y dimensions. In the depth dimension, we model the flow at location $(\mathbf{x}, z)$ as a difference in the index of the planes in the MPI representation, from $m_n$ to $m_{n-k}^w$. We refer to this difference as $z'$, where $z' \in \{-s_z, -s_z + 1, \ldots, 0, \ldots, s_z - 1, s_z\}$ and $2s_z + 1$ is the size of the window around the plane $z$ in the depth dimension. The network outputs a probability distribution $b_{z'}$ on the differences $z'$. Thus,

$$b_{z'}(\mathbf{x}, z) \in [0, 1] : \sum_{z'=-s_z}^{s_z} b_{z'}(\mathbf{x}, z) = 1 \quad \forall (\mathbf{x}, z). \tag{6.5}$$

Implementing Eq. (6.1) requires a real-valued 3D flow vector, $\mathbf{u}_{n \to n-k}^l$, which we compute as

$$\mathbf{u}_{n \to n-k}^l (\mathbf{x}, z) = \left( \mathbf{a}(\mathbf{x}, z), \left( \sum_{z'=-s_z}^{s_z} b_{z'}(\mathbf{x}, z) d(z + z') \right) - d(z) \right) \in \mathbb{R}^3, \tag{6.6}$$

where $d(z)$ is the depth corresponding to the $z^{\text{th}}$ plane in the MPI.

We incorporate the above and design a multi-scale 3D flow estimation network using PWC-Net [166] as the backbone architecture. PWC-Net consists of an encoder-decoder style architecture, where optical flow is estimated in a coarse-to-fine manner. Specifically, we first obtain multi-scale 3D features of $m_n$ and $m_{n-k}^w$ using encoders at each scale with 3D partial convolutions and downsampling layers consisting of strided convolutions. Since the number of MPI planes is much smaller than the resolution of the other two spatial dimensions, we do not downsample/upsample the features along the depth dimension. At the decoder in each scale except the lowest one, we upsample the flow estimated by the previous scale. Using this flow, we warp the features of $m_{n-k}^w$ and feed it to the masked correlation layers, along with the features of $m_n$. The masked correlation layers, as described in Eq. (6.3) and Eq. (6.4), output a cost volume which is then processed by subsequent partial convolution layers to estimate the residual flow at that scale. We estimate the final flow at two scales lower than the original resolution and upsample it by four times, as is popular in deep flow estimation models [166]. We train the optical flow network $\mathcal{F}_\Theta$, with trainable parameters $\Theta$ to estimate the flow from $m_n$ to $m_{n-k}^w$ as

$$\mathbf{u}_{n \to n-k}^l = \mathcal{F}_\Theta(m_n, m_{n-k}^w). \qquad (6.7)$$

**Loss functions:** We train the network $\mathcal{F}_\Theta$ in an unsupervised fashion with a linear combination of photometric loss $\mathcal{L}_{\mathrm{ph}}$ and a smoothness loss $\mathcal{L}_{\mathrm{smooth}}$. Specifically, we warp $m_{n-k}^w$ using $\mathbf{u}_{n \to n-k}^l$ to reconstruct $\hat{m}_n$. Photometric loss is a combination of mean absolute error (MAE) and structural similarity (SSIM) [190] as

$$\mathcal{L}_{\mathrm{ph}} = \beta \|(m_n - \hat{m}_n) \odot o_n\|_1 + (1-\beta)\frac{1 - \mathrm{SSIM}(m_n \odot o_n, \hat{m}_n \odot o_n)}{2}, \qquad (6.8)$$

where, $\beta$ is a scaling constant, $o_n$ is the occlusion mask and $\odot$ represents the element-wise product. The MAE and SSIM losses are computed in each of the $Z$ planes and averaged.

Unsupervised optical flow algorithms [120] compute photometric loss in the non-occluded regions only using an occlusion mask $o_n$ as in Eq. (6.8). The occlusion mask is typically computed using forward-backward consistency of the optical flow. We instead utilize the 3D representation of the scene and determine the occluded pixels as those which are hidden after warping $m_n$ with $\mathbf{u}_{n \to n-k}^l$. Mathematically, we forward-warp $m_n$ using $\mathbf{u}_{n \to n-k}^l$ to get $\hat{m}_{n-k}$. We compute a visibility mask for $\hat{m}_{n-k}$ as

$$\hat{v}_{n-k}(\mathbf{x}, z) = \prod_{y=1}^{z-1}(1 - \hat{\alpha}_{n-k}(\mathbf{x}, y)). \qquad (6.9)$$

We then backward-warp $\hat{v}_{n-k}$ using $\mathbf{u}_{n \to n-k}^l$ to get $\hat{v}_n$. Finally, we compute the occlusion mask as

$$o_n = \mathbb{1}_{\{\hat{v}_n > 0.5\}}. \qquad (6.10)$$

A value of 0 in $o_n$ indicates that the point is occluded. For the edge-aware smoothness loss, along with gradients of RGB, we also use gradients of alpha channel to weigh the smoothness term as

$$\mathcal{L}_{\text{smooth}} = (1 - \nabla\alpha_n) \cdot \exp(-a \cdot \nabla c_n) \cdot \nabla\mathbf{u}^l_{n \to n-k}, \qquad (6.11)$$

where $a$ is a scaling constant. Thus, our overall loss function is

$$\mathcal{L}_{\text{of}} = \mathcal{L}_{\text{ph}} + \lambda\mathcal{L}_{\text{smooth}}. \qquad (6.12)$$

**Future flow prediction:** We employ a linear motion model [12, 13] to predict the future flow as

$$\hat{\mathbf{u}}^l_{n \to n+k'}(\mathbf{x}, z) = -\frac{k'}{k} \, \mathbf{u}^l_{n \to n-k}(\mathbf{x}, z). \qquad (6.13)$$

Thus, to predict the future local motion, we first isolate the local motion between the past frames by nullifying the global motion between them and then estimate the local motion as 3D optical flow between the MPIs of the past frames. We then extrapolate the past motion to predict the future motion.

### 6.3.4 Disocclusion Infilling

As argued earlier, implementing Eq. (6.1) creates disocclusions. Hence we infill the disoccluded regions in $\hat{m}^w_{n+k'}$ using an approach similar to the one used by Srinivasan et al. [161]. We feed $\hat{m}^w_{n+k'}$ to a 3D U-Net and predict 2D infilling vectors in the disoccluded regions that point to known regions in the same plane of MPI. We then infill the disoccluded regions by copying the intensities and alpha from the locations pointed by the predicted infilling vectors to obtain $\hat{m}_{n+k'}$. Alpha-compositing $\hat{m}_{n+k'}$ generates the predicted frame $\hat{f}_{n+k'}$. We train the disocclusion infilling network with mean squared error loss between the predicted frame $\hat{f}_{n+k'}$ and the true frame $f_{n+k'}$. We find that the network fails to completely infill large disoccluded regions, leaving partially unfilled disoccluded regions. Hence, during inference, we iteratively infill the disoccluded regions $g$ times by recursively feeding the infilled MPI to the network.

## 6.4   Experiments

### 6.4.1   Datasets

**Our Dataset**: We develop a new dataset of videos with both camera and object motion due to the lack of any large scale datasets suitable for evaluating temporal view synthesis of dynamic scenes. We render the videos of our dataset with Blender using blend files from blendswap [ble] and turbosquid [tur] and add camera and object motion to the scenes. We add motion to the pre-existing scene objects or add new objects to the scene and animate them. Our dataset contains 200 diverse scenes of indoor environments such as hospital, kitchen, restaurant, and supermarket and outdoor environments like village, poolside, street, lake and so on. The scenes contain various moving objects such as books, chairs, tables, cars, airplanes, etc. For every scene, we generate four different camera trajectories covering different parts of the scenes and different kinds of object motion. Each sequence has 12 frames rendered at full HD resolution ($1920 \times 1080$) and 30fps. Thus, our dataset consists of 800 videos with 9600 frames in total. For every frame in our dataset, we store the corresponding ground truth depth, camera pose, and camera intrinsics. We use 135 scenes for training and 65 for testing.

**MPI-Sintel**: The MPI-Sintel dataset [26], which is widely used for evaluating optical flow estimation algorithms, contains both camera and object movement and also provides the ground truth depth and camera poses. Thus, it can be used to evaluate temporal view synthesis models. Since the required ground truth is provided for the train set only, we further divide the train set into train and test sets. The videos have a resolution of $1024 \times 436$ at 24 frames per second. We use 13 scenes for training and 10 scenes for testing.

We experiment on synthetic datasets only and not on real world datasets since our problem formulation is motivated by use-cases in increasing the frame rate for graphical rendering. Thus, we assume that the depth is available.

## 6.4.2 Comparisons

We compare our model against a combination of video prediction and view synthesis models. We use MCnet [180], a popular video prediction model, PreCNet [164], a recent model based on predictive coding and DPG [58], a model based on flow prediction and disocclusion infilling. For all the models, we use four past frames. Therefore, the prediction of first few frames uses the true past frames and the subsequent predictions use the previously predicted frames.

Since the above methods do not make use of camera motion, we combine these video prediction models with a recent view synthesis model, SynSin [193]. We first incorporate the camera motion by warping the past frames $f_n$, $f_{n-1}$, $f_{n-2}$ and $f_{n-3}$ to the view of $f_{n+1}$ using SynSin. We use the ablation model of SynSin, which uses true depth of the past frames. We then use video prediction models such as MCnet, DPG, or PreCNet on these warped frames to account for local motion and predict the desired frame. In order to guage the performance capability of this approach, we feed the warped $f_{n-1}$ and $f_{n-3}$ to the video prediction model, although these are not available during frame rate upsampling.

We implement DPG ourselves and train the model on $256 \times 256$ patches on both datasets. For MCnet, PreCNet, and SynSin, we use the code and pretrained models provided by the authors and finetune them on both datasets. We test both the pretrained and the finetuned models and report the best performance.

**Implementation details:** We train the optical flow estimation network $\mathcal{F}_\Theta$ and the disocclusion infilling network separately due to GPU memory constraints. We initialize our flow estimation network using pretrained weights provided by ARFlow [105] and finetune it on the respective datasets. We modify the pretrained weights appropriately to work for 3D convolutions. We train both the networks for 10000 iterations with patches of size $256 \times 256$ and a batch size of 4. We set the hyper-parameters as $s_z =$

Table 6.1: Quantitative comparison of different models on ours and MPI Sintel datasets for single frame prediction. Models indicated with * are a combination of view synthesis and video prediction models, that we design.

| Model | Our Dataset | | | | MPI Sintel | | | |
|---|---|---|---|---|---|---|---|---|
| | PSNR ↑ | SSIM ↑ | LPIPS ↓ | ST-RRED ↓ | PSNR ↑ | SSIM ↑ | LPIPS ↓ | ST-RRED ↓ |
| MCnet [180] | 24.66 | 0.7813 | 0.2406 | 207 | 24.00 | 0.7511 | 0.2230 | 530 |
| DPG [58] | 28.24 | 0.8634 | 0.1091 | 71 | 20.00 | 0.6385 | 0.3056 | 1129 |
| PreCNet [164] | 24.86 | 0.8191 | 0.2409 | 244 | 25.60 | 0.7952 | 0.2463 | 571 |
| SynSin [193] + MCnet* | 26.87 | 0.8254 | 0.1567 | 92 | 25.67 | 0.8031 | 0.1639 | 315 |
| SynSin + DPG* | 27.30 | 0.8461 | 0.1268 | 74 | 23.77 | 0.7795 | 0.2520 | 600 |
| SynSin + PreCNet* | 26.81 | 0.8432 | 0.1508 | 100 | 25.92 | 0.8205 | 0.1581 | 330 |
| **DeCOMPnet** | **30.60** | **0.9314** | **0.0634** | **28** | **29.64** | **0.8975** | **0.1032** | **259** |

$1, Z = 4, \beta = 0.15, a = 10, \lambda = 10, g = 3$.

**Evaluation Measures:** We evaluate the predicted frames using various image quality measures such as peak signal-to-noise ratio (PSNR), structural similarity index (SSIM) [190] and LPIPS [215]. Further, since image quality measures do not evaluate temporal quality, we also employ a video quality assessment measure, ST-RRED [160] that measures both the spatial and temporal quality of the predicted frames. Since the focus of this work is not on predicting new regions entering the scene, we crop out 40 pixels on the top and bottom of the frames and 60 pixels on the left and right sides of the frames before evaluating the predictions.

### 6.4.3   Single Frame Prediction

In single frame prediction, the goal is to predict every alternate frame and this can be studied by setting $k = 2$ in our problem definition. Specifically, to predict $\hat{f}_{n+1}$, we use $f_n$ and $f_{n-2}$.

We first present examples of a few future frame predictions by DeCOMPnet and visualizations of outputs of various stages in our framework in Fig. 6.3. In particular, we show the outputs after predicting the object motion alone, $\tilde{f}_{n+1}^w$, and after incorporating the global motion. Since such outputs are in the MPI representation space, we use

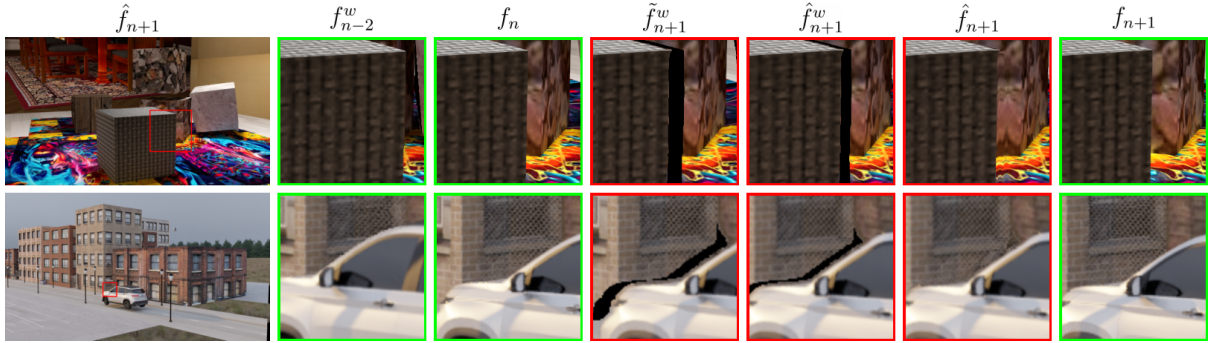

Figure 6.3: Visualization of outputs of various stages in our framework: Each row shows a different sample. The first column shows the full resolution frame and the subsequent columns show an enlarged region of a cropped region. The second and third columns show past frames after camera motion compensation. The fourth and fifth columns show the frame after predicting local and global motion respectively, which contain disocclusions (shown in black). The sixth column shows the result after infilling and the last column shows the true frame.

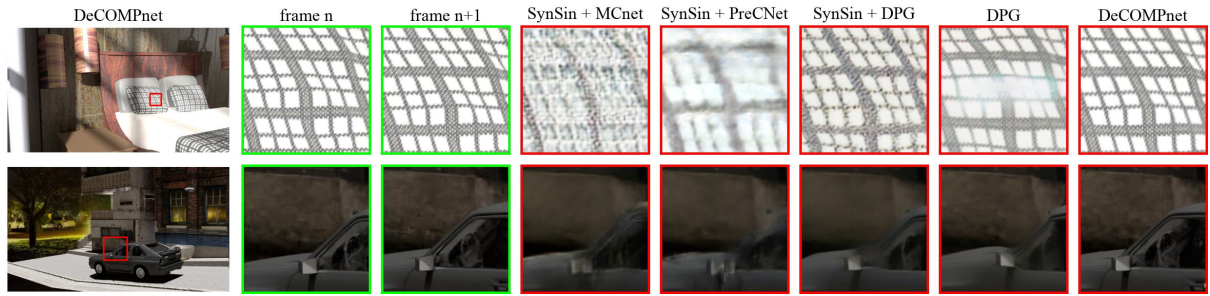

Figure 6.4: Qualitative comparisons on our dataset for single frame prediction. The first column shows a predicted frame by our model, DeCOMPnet, and the subsequent columns show enlarged versions of a cropped region for different models. The frames with green border are graphically rendered, and those with red border are predicted by different models. In the scene in the first row, the pillows along with the bed are moving towards the camera. The car is moving left in the second scene. All scenes have camera motion in addition to object motion. We observe that other models fail to produce sharp predictions or retain the object shape, whereas our model has retained the shape and textures.

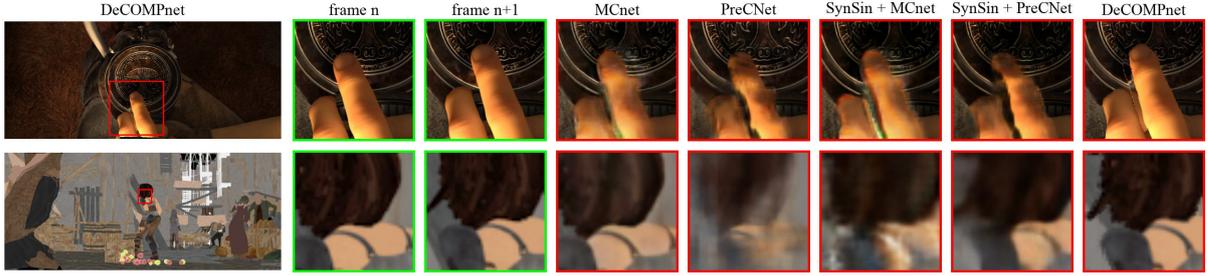

Figure 6.5: Qualitative comparisons on the MPI Sintel dataset for single frame prediction. The fingers are moving up in the first scene, and the girl is moving to the right in the second scene. We observe that our model has retained the shape of the objects, which the other models fail to.

alpha-compositing to obtain the corresponding images.

We compare the quantitative results of DeCOMPnet against the competing methods in Tab. 6.1. Our model outperforms all the competing methods in terms of all the quality measures. The relatively lower ST-RRED scores for DeCOMPnet indicate that the predictions by our model are superior in temporal quality. We observe that most models perform better on our dataset than on the MPI-Sintel dataset in general. This may be because the MPI-Sintel dataset has complex motion to make it challenging for optical flow estimation, making it even more challenging for prediction. We also observe that combining SynSin with video prediction models improves their performances, except for DPG on our dataset. Since DPG is performing reasonably well, when combined with SynSin, the artifacts introduced by SynSin may lead to a decrease in performance. However, on the MPI-Sintel dataset, since the performance of DPG is lower, it benefits from using SynSin. Further, we note that even though the combination of view synthesis and video prediction models use the knowledge of the true frames $f_{n-1}$ and $f_{n-3}$ which are not available at test time, our model still shows superior performance.

We show the qualitative results of our model and the benchmarked models in Fig. 6.4 and Fig. 6.5. While other models introduce artifacts such as blur or distortions in the shape and texture of objects, DeCOMPnet predicts the future frame reasonably well. To notice the temporal superiority of our model, view the supplementary videos on our project webpage[1].

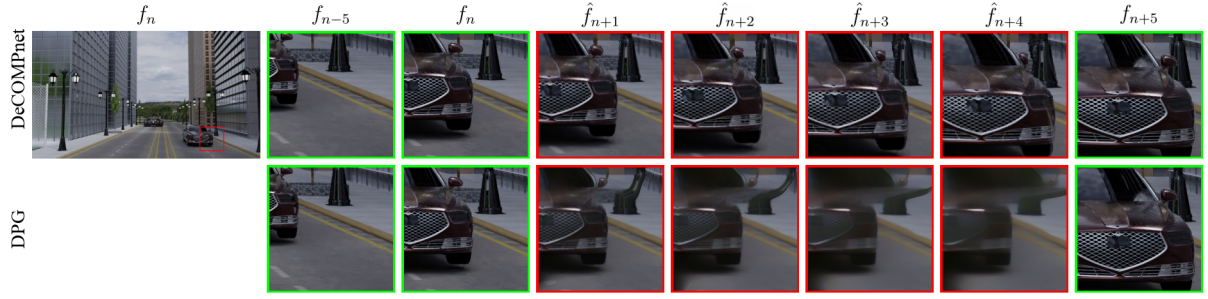

Figure 6.6: Multi frame predictions by DeCOMPnet. The first column shows $f_n$ at full resolution and the subsequent columns focus on a cropped region of $f_{n-5}, f_n$ and the four predicted frames. The last column shows $f_{n+5}$ for reference.

### 6.4.4 Multi Frame Prediction

We now analyze the ability of our model to predict multiple frames into the future. In particular, we study frame-rate upsampling by a factor of five times by setting $k = 5$. In our framework, we estimate the object motion $\mathbf{u}_{n \to n-5}^l$ only once, and compute the predicted motion for each of the future time steps using Eq. (6.13). We then use Eq. (6.1) to warp $m_n$ to $\hat{m}_{n+1}^w, \hat{m}_{n+2}^w, \hat{m}_{n+3}^w$ and $\hat{m}_{n+4}^w$, which are then infilled and alpha composited to predict the future frames. For the benchmark comparison models, $f_n, f_{n-1}, f_{n-2}$ and $f_{n-3}$ are used to predict $\hat{f}_{n+1}$. Thus, compared to our model, the benchmarked models have the additional knowledge of $f_{n-1}, f_{n-2}$ and $f_{n-3}$. Although, these frames are not available in practice, the goal of this experiment is to analyze the performance of this approach.

Fig. 6.6 compares example multi-frame predictions by DeCOMPnet with DPG and Fig. 6.7 shows average PSNR and SSIM for different models. We observe that DeCOMPnet outperforms all the competing models in terms of SSIM. In terms of PSNR, we are competitive with DPG in the prediction of $\hat{f}_{n+1}$ on our dataset, despite DPG additionally using $f_{n-1}, f_{n-2}$ and $f_{n-3}$. Further, DPG predictions are often blurry, which is not captured well by PSNR.

---

[1]https://nagabhushansn95.github.io/publications/2022/DeCOMPnet.html

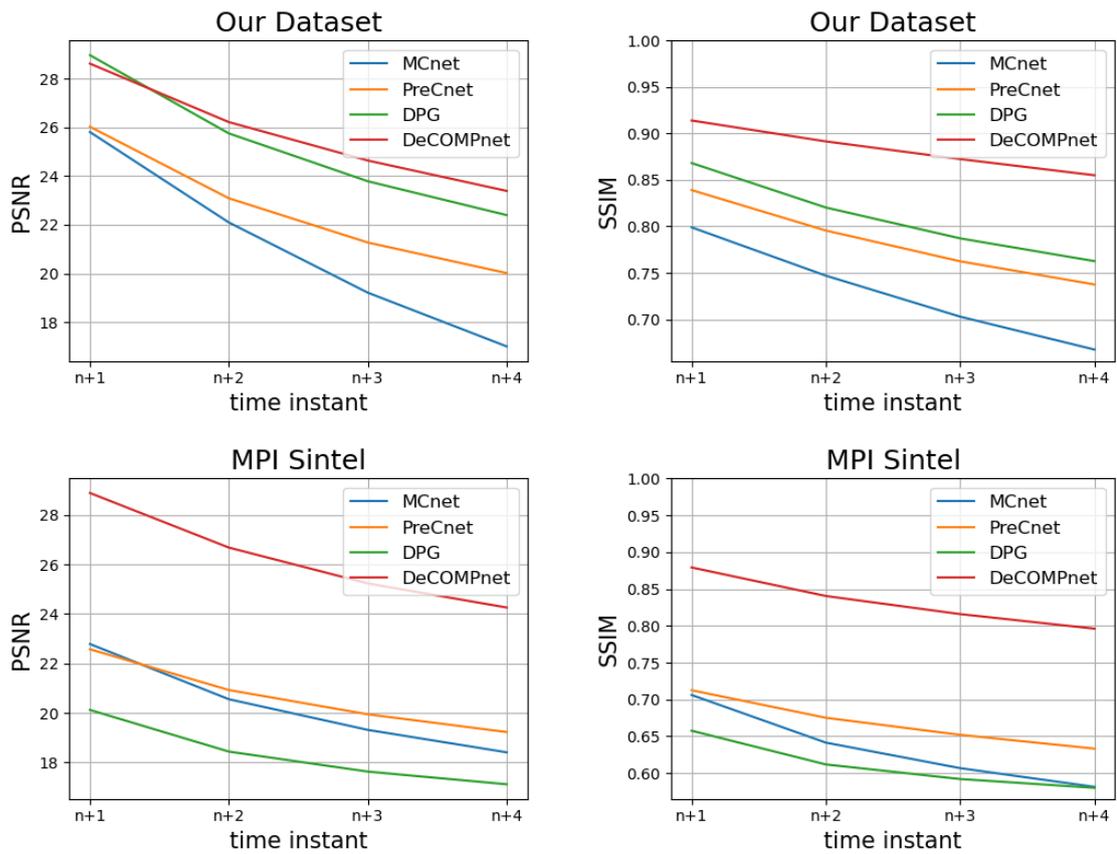

Figure 6.7: Quantitative comparison of the proposed DeCOMPnet against competing methods for multi frame prediction. The plots show average quality score for the predicted frames $\hat{f}_{n+1}$, $\hat{f}_{n+2}$, $\hat{f}_{n+3}$ and $\hat{f}_{n+4}$.

Table 6.2: Comparison of average endpoint error for the flows predicted by different ablated models, for single frame prediction.

| Model | Endpoint Error |
|---|---|
| 2D Flow | 2.8 |
| 3D Flow w/o p-conv and mask-corr | 2.0 |
| **3D Flow** | **1.7** |

## 6.4.5   Ablations

**2D vs 3D Flow Prediction:** We compare our 3D flow prediction model against a 2D flow model by predicting 2D flow between the frames $f_n$ and $f_{n-2}^w$. We use a model similar to the one described in Sec. 6.3.3 on frames with 2D convolutions and cost volumes. Note that this model still uses partial convolutions and masked correlation layers to handle holes in $f_{n-2}^w$. We also feed depth as input to the flow estimation network for a fair comparison. Owing to different ranges of depth across multiple scenes, we first normalize depth to the range $[0, 1]$, and then feed it to the flow estimation network. While the 2D model uses depth naively by concatenating depth with the input in an additional channel, the 3D model uses a more structured MPI representation. This comparison allows us to analyse the importance of using MPIs for flow estimation.

We evaluate the flows predicted by 2D and 3D models using average endpoint error (AEPE) [105] for single frame prediction. For the test scenes in our dataset, we additionally render the optical flow corresponding to object motion alone and use it to compute the endpoint errors. Even though our model predicts 3D flow, we use the x-y components only to compute the AEPE. As argued earlier, estimating the object motion in 3D allows better matching of points, leading to a more accurate estimation of flow, even in x-y dimensions.

From Tab. 6.2, we observe that estimating the flow in 3D using MPI reduces AEPE

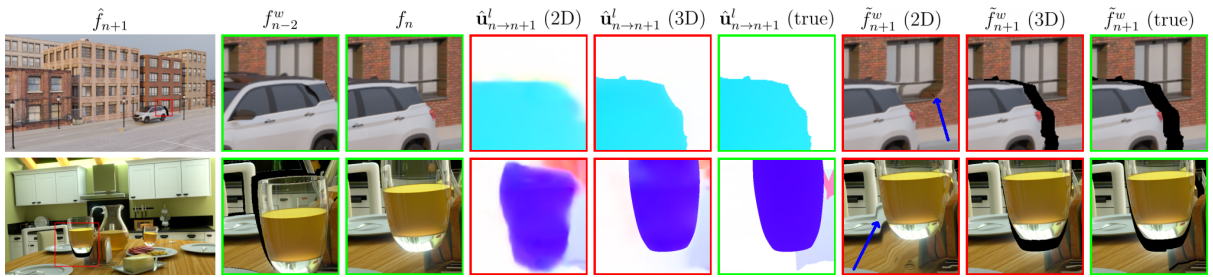

Figure 6.8: Qualitative comparison of 2D and 3D flow estimations for single frame prediction. The second and third columns show the input frames $f_n$, $f_{n-2}^w$ to the flow estimation networks. Fourth, fifth and sixth columns visualize the x-y component of flows predicted by the models and the ground truth flow. The next three columns show the corresponding frames $\tilde{f}_{n+1}^w$ reconstructed by applying local flow $\hat{\mathbf{u}}_{n\to n+1}^l$ on $f_n$. Notice the sharpness of 3D flow and the distortions in the background of the frame reconstructed with 2D flow as pointed by the blue arrow. Disoccluded regions are shown in black. Here we only visualize object motion prediction and do not show the final predicted frame. Global motion and infilling need to be applied on top of $\tilde{f}_{n+1}^w$ as shown in Fig. 6.3 to obtain $\hat{f}_{n+1}$. Optical flow visualization is similar to Baker et al. [10].

Table 6.3: Performance bound analysis of different components of our model for single frame prediction. LMP: Local motion prediction; DI: Disocclusion infilling. Pred indicates flow prediction or infilling done by the network. GT indicates ground truth flow or infilling.

| LMP | DI | Our Dataset | | MPI Sintel | |
|:---:|:---:|:---:|:---:|:---:|:---:|
| | | PSNR ↑ | SSIM ↑ | PSNR ↑ | SSIM ↑ |
| pred | pred | 30.60 | 0.9314 | 29.64 | 0.8975 |
| GT | pred | 30.67 | 0.9354 | 31.90 | 0.9426 |
| pred | GT | 32.00 | 0.9377 | 30.35 | 0.9097 |
| GT | GT | 33.53 | 0.9453 | 34.02 | 0.9613 |

by 38%. Further, we observe in Fig. 6.8 that the flow predicted by our 3D model is sharper leading to undistorted reconstructions at the edges, in contrast to 2D flow.

**Impact of partial convolutions and masked correlations:** We study the impact of the partial convolution and masked correlation layers in DeCOMPnet by replacing them with standard 3D convolution and correlation layers. We evaluate the performance of object motion prediction using AEPE in Tab. 6.2. We observe that the proposed masked correlations and the use of partial convolutions to handle the sparsity in MPI representation lead to a significant improvement in the performance of object motion prediction.

### 6.4.6   Analysis of Performance Bounds

We now analyze the upper bound on the performance of our model components for single frame prediction. We establish an upper bound on the performance that can be achieved by improving the object motion prediction by replacing the predicted total motion with the true optical flow provided by the graphics renderer. We warp $f_n$ with the ground truth optical flow to get $\hat{f}_{n+1}$ and then create $\hat{m}_{n+1}^w$ as explained in Sec. 6.3.1, which is then fed to the disocclusion infilling module. To upper bound the performance that can be achieved by improving the disocclusion infilling, we apply alpha compositing on $\hat{m}_{n+1}^w$ and replace the disoccluded regions with true intensities from $f_{n+1}$. We also obtain a joint bound using both true optical flow and infilling with true intensities.

We see from Tab. 6.3 that the performance of our model is close to the upper bound on our dataset. The larger gap in the MPI-Sintel database could be attributed to the challenging motion trajectories. The non-perfect reconstruction performance of the bound in the last row of Tab. 6.3 may be due to splatting approximations in warping.

### 6.4.7   Timing Analysis

Our model takes 4.5s to predict a single full HD frame on an Intel Core i7-9700F CPU with 32GB RAM and NVIDIA RTX 2080 Ti GPU, whereas Blender typically takes about 5m-1h to render a single frame depending on the scene. On further analysis, we find that the convolutional layers in optical flow estimation and disocclusion infilling take about 70ms and 3ms, respectively. Thus a significant amount of time in our implementation is consumed by warping operations. However, it is possible to optimize warping as shown by Barnes et al. [15] and Waveren et al. [177], which use less than 10ms. Further, due to the sparsity of the MPI representation, at any given location, alpha will be 0 on $Z - 1$ planes. Although we ignore convolution layer outputs at locations where $\alpha = 0$, inference time can be further reduced by $Z$ times by not convolving such points. With the above optimizations, the inference time of our model could reduce to less than 33ms, making it feasible for real-time use.

## 6.5   Summary

In this chapter, we propose a novel framework for temporal view synthesis of dynamic scenes in the context of causal frame-rate upsampling of videos. We account for camera and object motion sequentially, which allows our framework to exploit the availability of camera motion effectively. Further, we estimate and predict object motion in the 3D MPI representation using masked correlations and partial convolutions. Finally, we infill disocclusions in the warped MPIs and use alpha-compositing to render the predicted frames. To evaluate our model, we develop a new dataset that brings out the challenges in temporal view synthesis.

# Chapter 7

# Conclusion

In this thesis, we present several sparse input novel view synthesis algorithms for both static and dynamic scenes. To train NeRFs effectively for sparse input view synthesis, we first propose the visibility prior that is related to the relative depth of the objects in the scene. We compute the visibility prior using plane sweep volumes without the need to train a neural network on large datasets. We show that the visibility prior is dense and more reliable than existing priors on absolute depth, thereby providing better prior to the sparse input NeRF. For effective imposition of the visibility prior, we reformulate the NeRF representation and show that our model outperforms existing approaches on popular datasets.

To further improve the priors, we explore the use of augmented models to obtain dense depth supervision that does not suffer from generalization issues. We design the augmented models by reducing the capability of the NeRF. This reduced capability NeRF predicts only simpler solutions and provides better depth supervision in certain regions of the scene. We ensure that our dense depth priors do not suffer from generalization issues by training these augmented models on the given scene alone and in tandem with the main NeRF. Further, we show that our framework of learning simpler solutions is not only applicable to implicit models such as NeRF but also to newer and faster radiance field models like TensoRF and ZipNeRF. Through extensive experiments on four different datasets, we show that our Simple-RF family of models significantly improve the performance of their respective base models, as well as prior works, in sparse input novel view synthesis of static scenes.

For novel view synthesis of dynamic scenes, we design a compact motion model based on factorized volumes that optimizes quickly and is amenable to be regularized with motion priors in the sparse input setting. We introduce reliable sparse flow priors based on robust SIFT feature matching to constrain the motion field, since we find that the popularly employed dense optical flow priors are unreliable. We show the benefits of our approach in sparse input settings, where our motion representation, along with our priors, outperform prior works significantly.

Finally, we study the application of view synthesis for frame rate upsampling in video gaming applications. We utilize the available scene depth in such synthetic rendering settings to isolate the object motion from the camera motion in the given past frames. Since the object motion in 2D is noisy and unreliable due to disocclusions, we present an approach to estimate the object motion in 3D. Specifically, we employ the multi-plane images to represent the scene in 3D and design a deep flow estimation network to estimate the object motion in 3D. In particular, we show that the use of MPIs for motion estimation significantly improves the view synthesis performance.

In all the above work, we consider the applications where camera parameters are known exactly. In the future, it might be interesting to consider the case where the camera parameters are not known exactly and are optimized jointly with the scene representation. While such frameworks are studied in the dense input setting [131], achieving the same in sparse input setting is non-trivial and requires a careful investigation. Specific to the multi-view dynamic scene novel view synthesis, we assume that the cameras are temporally synchronized. Handling multi-view videos with small temporal shifts could enable wider application of the proposed methods. In this thesis, we consider datasets that have uniform lighting across all the views. However, in some applications [116], the lighting conditions might vary across the views. It would be interesting to design regularizations to train the radiance fields in such extreme conditions. Finally, we operate in the setting where the input views are fixed, and we try to maximize the quality of the synthesized novel views. In many practical applications, it might be more useful to determine the number of input views required and the optimal position of cameras to achieve a certain target quality of novel views.

# Chapter 8